\documentclass[10pt,twocolumn,letterpaper]{article}
\pdfoutput=1

\usepackage{iccv}     
\usepackage{graphicx}
\usepackage{amsmath}
\usepackage{amssymb}
\usepackage{booktabs}
\usepackage{times}

\usepackage{lipsum,graphicx,subcaption}
\usepackage{duckuments}
\usepackage{bbding}
\usepackage{pifont}
\usepackage{verbatim}
\usepackage{tikz}
\usepackage{xcolor}
\usepackage[percent]{overpic}
\usepackage{pict2e}
\usepackage{booktabs}
\usepackage{enumitem}  
\usepackage{tablefootnote}
\usepackage{bm}
\usepackage{docmute}

\usepackage[pagebackref,breaklinks,colorlinks]{hyperref}

\def\vec#1{\mathbf{#1}}

\usepackage[capitalize]{cleveref}
\crefname{section}{Sec.}{Secs.}
\crefname{section}{Section}{Sections}
\crefname{table}{Table}{Tables}
\crefname{table}{Tab.}{Tabs.}

\iccvfinalcopy

\begin{document}
\title{BlobGAN-3D: A Spatially-Disentangled 3D-Aware \\
Generative Model for Indoor Scenes}

\author{Qian Wang\\
KAUST\\
{\tt\small qian.wang@kaust.edu.sa}
\and
Yiqun Wang\\
KAUST\\
{\tt\small csyqwang@hotmail.com}
\and
Michael Birsak\\
KAUST\\
{\tt\small michael.birsak@kaust.edu.sa}
\and
Peter Wonka\\
KAUST\\
{\tt\small peter.wonka@kaust.edu.sa}
}

\def\iccvPaperID{10155} 
\def\confName{ICCV}
\def\confYear{2023}

\maketitle
\begin{abstract}
3D-aware image synthesis has attracted increasing interest as it models the 3D nature of our real world. However, performing realistic object-level editing of the generated images in the multi-object scenario still remains a challenge. Recently, a 2D GAN termed BlobGAN has demonstrated great multi-object editing capabilities on real-world indoor scene datasets. In this work, we propose \textbf{BlobGAN-3D}, which is a 3D-aware improvement of the original 2D BlobGAN. We enable explicit camera pose control while maintaining the disentanglement for individual objects in the scene by extending the 2D blobs into 3D blobs. We keep the object-level editing capabilities of BlobGAN and in addition allow flexible control over the 3D location of the objects in the scene. We test our method on real-world indoor datasets and show that our method can achieve comparable image quality compared to the 2D BlobGAN and other 3D-aware GAN baselines while being able to enable camera pose control and object-level editing in the challenging multi-object real-world scenarios.
\end{abstract}

\section{Introduction}
Generative Adversarial Networks (GANs) \cite{goodfellow2014gan} have seen great success in generating highly-realistic images in various tasks~\cite{karras2017progan,zhu2017cyclegan,brock2018biggan,choi2019starganv2}, and achieved state-of-the-art image quality \cite{anokhin2020cips,kumari2021visionaidedgan,sauer2021projectedgan,sauer2021styleganxl,karras2021stylegan3}. Recently, a lot of exciting progress was made in 3D-aware image generation~ \cite{park2017transgrounded,zhu2018von,nguyen2019hologan,schwarz2020graf,chan2021pigan,chan2022eg3d}. 3D-aware image generation is aiming at generating images that allow for explicit camera control. Different from approaches that require multiview images \cite{park2017transgrounded} or additional geometry supervision \cite{zhu2018von} as input, recent works focus on only utilizing single-view images to model the 3D nature of real-world objects~\cite{nguyen2019hologan,schwarz2020graf,chan2021pigan,chan2022eg3d}. Nevertheless, current 3D-aware image generation mainly targets the synthesis of individual objects using a NeRF (Neural Radiance Field)~\cite{mildenhall2020nerf,zhang2020nerfpp} representation. This yields good control for the properties of a single object, \eg shape and texture, but it is not a suitable representation to learn the composition of a scene consisting of many individual objects. 

\begin{figure}
  \centering
  \begin{minipage}[t]{0.04\linewidth}
  \centering
  \rotatebox{90}{\ \ \ \ \ \ \ Viewpoint}
  \end{minipage}
  \begin{subfigure}[t]{0.31\linewidth}
    \includegraphics[width=\linewidth]{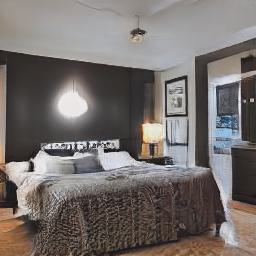}
\end{subfigure}
  \begin{subfigure}[t]{0.31\linewidth}
    \includegraphics[width=\linewidth]{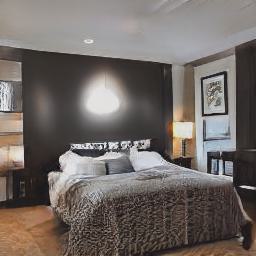}
\end{subfigure}
  \begin{subfigure}[t]{0.31\linewidth}
    \includegraphics[width=\linewidth]{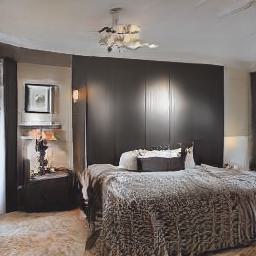}
\end{subfigure}
  \begin{minipage}[t]{0.04\linewidth}
  \centering
  \rotatebox{90}{\ \ \ \ \ \ \ \ \ Moving}
  \end{minipage}
  \begin{subfigure}[t]{0.31\linewidth}
    \includegraphics[width=\linewidth]{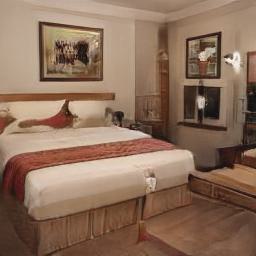}
\end{subfigure}
  \begin{subfigure}[t]{0.31\linewidth}
    \includegraphics[width=\linewidth]{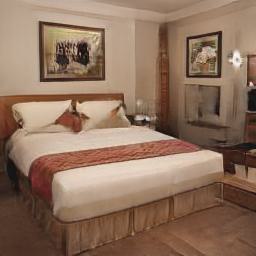}
\end{subfigure}
  \begin{subfigure}[t]{0.31\linewidth}
    \includegraphics[width=\linewidth]{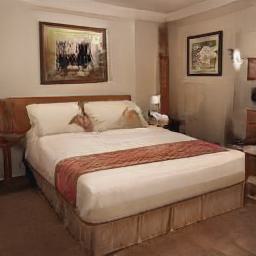}
\end{subfigure}
  \begin{minipage}[t]{0.04\linewidth}
  \rotatebox{90}{\ \ \ \ \ \ \ Restyling}
  \end{minipage}
  \begin{subfigure}[t]{0.31\linewidth}
{\includegraphics[width=\linewidth, height=\linewidth]{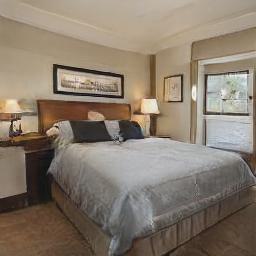}}
\end{subfigure}
  \begin{subfigure}[t]{0.31\linewidth}
{\includegraphics[width=\linewidth, height=\linewidth]{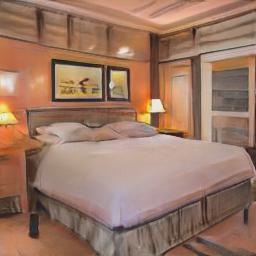}}
\end{subfigure}
  \begin{subfigure}[t]{0.31\linewidth}
{\includegraphics[width=\linewidth, height=\linewidth]{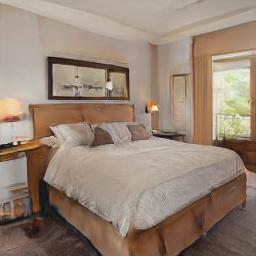}}
\end{subfigure}
  \caption{\textbf{Images generated by BlobGAN-3D.} Our method can control the viewpoint while enabling object-level editing like moving and restyling on a real-world indoor dataset.}
\end{figure}

For editing capabilities, multi-object scene editing is more challenging than single-object editing, as it requires disentanglement of individual objects. Ideally, a network should learn to distinguish individual objects without explicit supervision. Initial approaches towards that goal \cite{eslami2016attend_infer,kosiorek2018sequential_attend,engelcke2019genesis,burgess2019monet,anciukevicius2020object_centric,yang2020manipulate_individual,steenkiste2020invest_object_comp} have studied to manipulate the location or the pose of objects in 2D space in relatively simple synthetic datasets \cite{justin2016clevr,sun2019multimnist,burgess2019monet}. BlockGAN \cite{nguyen2020blockgan} adopts a 3D framework to a GAN backbone to model individual objects in a synthetic dataset. GIRAFFE \cite{niemeyer2021giraffe} is a 3D-aware GAN that can manipulate single objects in real-world datasets and can scale to multi-object scenes thanks to its compositional modeling of scenes. However, for multi-object scenes, GIRAFFE can only manipulate synthetic datasets but fails to perform well on more complex real-world scenes. \par
Recently, BlobGAN \cite{epstein2022blobgan} has achieved significant progress in object-level disentanglement on complex real-world datasets \cite{fisher2015lsun}. As a result, a user can edit the shape, appearance, size, and location of individual objects in a scene by manipulating the corresponding blob representation of objects while maintaining the realism of generated images. However, as a 2D GAN, BlobGAN has its limitations, \eg it is not aware of the 3D location of objects. Therefore, we extend BlobGAN into a 3D-aware GAN while keeping its disentanglement for individual objects in the scene. Specifically, we extend the blob parameterization from 2D to 3D and use a mapping network to predict the 3D parameters along with the feature and style vectors for each blob. Given a camera pose, we use the Mahalanobis distance to compute the density at each query point along a camera ray. We use volume rendering to compute a feature map per blob and solve mutual occlusions by taking into account the depth of their centroid w.r.t. the camera, resulting in a feature grid. This feature grid is then fed into a synthesis network to generate an RBG image. In addition, we use a pre-trained depth estimator to improve the multiview consistency of images. \par
As indoor scenes have a multi-object nature, we choose to focus on indoor datasets. We test our model on real-world indoor datasets and find that our model can achieve an image quality comparable to 2D BlobGAN and other 3D-aware GAN baselines. Moreover, to the best of our knowledge, our method is the \textbf{first} to enable camera pose control and object-level editing in this challenging multi-object scenario without supervised disentanglement of the individual objects. We also support new editing capabilities by explicitly controlling the 3D location of the blobs in the scene, which allows us to achieve a foreshortening effect. We summarize our contributions as the following:
\begin{enumerate}[label=(\roman*)]
\item We extend 2D BlobGAN to be 3D-aware, which enables us to generate images by explicitly controlling the camera. Meanwhile, we keep the disentanglement of objects and allow for realistic object-level editing.
\item We allow more editing capabilities by explicitly controlling the 3D  location of blobs, while at the same time enabling more freedom to determine the occlusions between objects.
\item We achieve a foreshortening effect when moving blobs along the depth dimension, which does not exist in BlobGAN due to its 2D nature.
\end{enumerate}

\section{Related Work}
\subsection{Unconditional 2D GANs} 
GANs \cite{goodfellow2014gan} have achieved great success in unconditional image synthesis tasks \cite{karras2017progan,zhang2018sagan, brock2018biggan, shaham2019singan}. StyleGAN in all its variants \cite{karras2018stylegan,karras2019stylegan2,karras2021stylegan3} has produced high-quality images while enabling explicit control over the disentangled latent space. StyleGAN has also been adopted to be the backbone of other image synthesis works~ \cite{zhang2021styleswin,Karras2020ada,sauer2021styleganxl,epstein2022blobgan}. However, 2D GANs do not encode the 3D nature of the depicted image content, which is essential to novel view synthesis and certain types of edits.
\vspace{-1mm}
\begin{figure*}[htbp]
  \centering   \includegraphics[width=\linewidth,trim=0 120 0 100,clip]{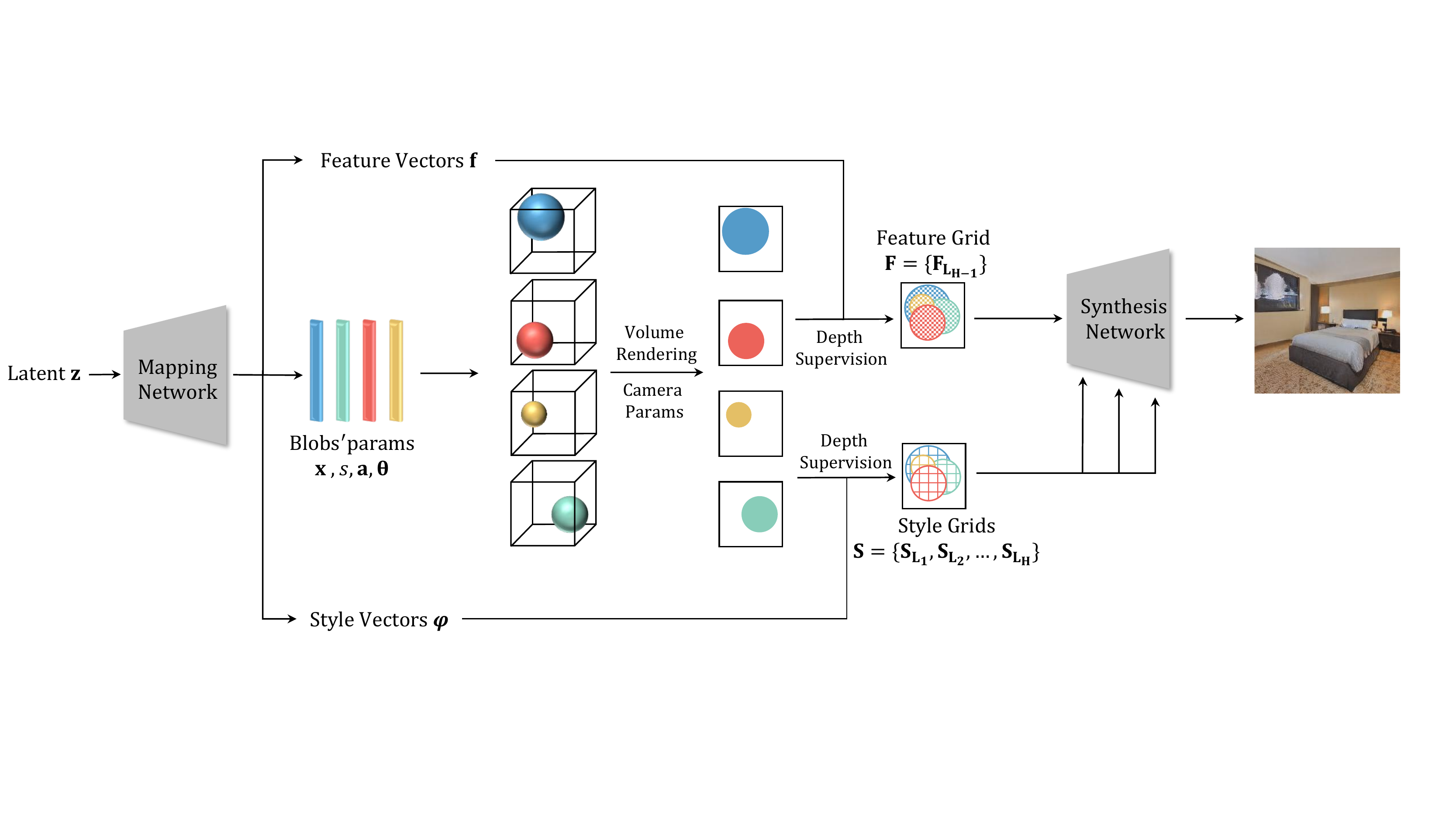}
   \caption{\textbf{The architecture of the generator in BlobGAN-3D.} A latent vector is fed into a mapping network to obtain the parameters of all the blobs. Given a camera pose, the blobs' parameters are then used to compute the density for all the query points along the camera rays. We then use volume rendering to go from 3D density volumes to 2D density maps. The feature vectors and style vectors are then splatted on the density maps to get the corresponding grids, which serve as the input and condition for the synthesis network, respectively. The final RGB image is rendered by the synthesis network. }
   \label{fig:generator}
\vspace{-4mm}
\end{figure*}
\subsection{3D-aware GANs}
Instead of directly synthesizing 2D images, 3D-aware GANs find an object's or even whole scene's 3D representation and derive 2D images from it. Given only single-view 2D images as input, 
voxel representations \cite{henzler2018platonicgan,nguyen2019hologan} are used but they introduce large memory constraints. In contrast, neural fields are continuous and can synthesize images at arbitrary resolution. Neural fields are popular for 3D geometry processing tasks~\cite{chen2018implicit_decoder,mescheder2019occupancy,park2019deepsdf,genova2019shapetemplates,chabra2020localshapes,peng2020convoccupancy,vicini2022sdf} and also enable 3D-aware image synthesis \cite{or2022stylesdf,schwarz2022voxgraf,pan2021shading,xu2021generative}. 
NeRF~\cite{mildenhall2020nerf,zhang2020nerfpp} combines the neural implicit network with differentiable volume rendering to enable explicit camera control for novel view image generation. Based on NeRF, GRAF \cite{schwarz2020graf}, pi-GAN \cite{chan2021pigan} and GRAM\cite{deng2022gram} propose to learn an implicit network to model the scene. GIRAFFE and GIRAFFE-HD \cite{niemeyer2021giraffe,xue2022giraffe_hd} learn a compositional neural field and enable object-level control.
StyleNeRF \cite{gu2022stylenerf} learns a low-dimensional feature map using NeRF and then progressively upsamples it to obtain the final RGB image. VolumeGAN \cite{xu2021volumegan} learns an extra convolutional structural representation as the coordinate descriptor for the mapping network. EG-3D \cite{chan2022eg3d} proposes an explicit-implicit tri-plane representation to combine efficiency and expressiveness. CIPS-3D \cite{zhou2021cips3d} synthesizes images directly in a pixel-wise manner and enables convenient transfer learning tasks. 
MVCGAN \cite{zhang2022mvcgan} uses explicit multiview constraints to enhance multiview consistency over images. \cite{shi2022indoor_depth} builds a dual generator to take both RGB images and depth maps as input to help generate 3D indoor scene images. 
EpiGRAF \cite{skorokhodov2022epigraf} proposes a new patch-wise training scheme to efficiently obtain high-resolution images. 
However, most of these works only focus on the quality of the generated image and multiview consistency but are not aware of local multi-object editing in images. Some of the works \cite{zhu2018von,nguyen2019hologan,xu2021volumegan,sun2022ide3d,sun2022fenerf} can enable local editing, but are only limited to the single object context. 
\subsection{Editability in GANs}
For object-level editing, some works \cite{zhu2018von,nguyen2019hologan,xu2021volumegan} can learn disentangled shape and appearance latent codes and enable independent control over the structure and texture. IDE-3D \cite{sun2022ide3d} and FENeRF \cite{sun2022fenerf} use semantic masks as an extra source of input to enable more local editing on human faces. However, some works \cite{sun2022ide3d,sun2022fenerf} only focus on face data, and require paired semantic mask labels as input. BlockGAN \cite{nguyen2020blockgan} computes a voxel feature grid for individual objects and successfully allows object-level control in a synthetic multiple-object scene \cite{justin2016clevr}. Instead of using a voxel representation, GIRAFFE \cite{niemeyer2021giraffe} learns multiple implicit fields for objects and also achieves good editing results on synthetic datasets. Liao \etal \cite{liao2020towards} propose 3D controllable image synthesis. Compared to our work, they use a differentiable projection layer rather than volume rendering, and use a 2D generator to render each object individually, rather than rendering a composite image. For BlockGAN, GIRAFFE and \cite{liao2020towards}, when dealing with more complex real-world datasets like LSUN Bedroom \cite{fisher2015lsun}, they struggle to generate high-quality images and synthesize poor multiview images. BlobGAN \cite{epstein2022blobgan} uses a mid-level blob representation to model the scene and enables various local editing operations for individual objects in the scene on the challenging LSUN indoor datasets. It can move, remove, duplicate, swap, and re-stylize objects in the scene by manipulating the corresponding blobs. However, as a 2D GAN, BlobGAN ignores the 3D structure of the real-world scene, which limits its capability of editing. In this work, we extend BlobGAN to be 3D-aware, therefore allowing more control over editing while synthesizing novel-view images.

\section{Method}
Our method adopts the model architecture and design from BlobGAN \cite{epstein2022blobgan}. Similarly, a random noise vector is fed into an MLP mapping network to obtain a representation of each blob. These representations are then mutually processed to get a joint feature grid, which is fed into a StyleGAN2-based synthesis network to generate images. Originally, the blobs are parameterized as ellipses, \ie in 2D space. We extend the blobs to be modeled as 3-dimensional ellipsoids and colocate them in 3D scene space. We use volume rendering to go from 3D scene space to a 2D feature map and use a StyleGAN2 synthesis network to obtain the final RGB image. The architecture of the generator is shown in \cref{fig:generator}.

\subsection{Extending from 2D space to 3D space}

\subsubsection{Blob parameterization}
We model the whole scene as a 3-dimensional normalized coordinate system and use $M$ 3D blobs to represent its layout. We effectively double the number compared to BlobGAN and use 10 geometric parameters to define each blob $\{ b_i \}_{i=1}^M$ by its center coordinate $\vec x_i \in \mathbb{R}^3$, scale factor $s_i \in \mathbb{R}$, normalized aspect ratio $\vec a_i \in \mathbb{R}^3$ and Euler rotation angles $\boldsymbol{\theta}_i \in \mathbb{R}^3$. We also learn a feature vector $\vec f_i \in \mathbb{R}^{d_{s}}$ and a style vector $\boldsymbol{\phi}_i \in \mathbb{R}^{d_{t}}$ to encode the structure and texture information per blob, where $d_s$ and $d_t$ are the dimensions of the feature vector and style vector, respectively.

\subsubsection{Volume rendering}
For a single pass, we assume a fixed camera pose to render all visible blobs. To obtain the density value at a query point $\vec x_q$ along a ray through the camera, we first calculate the squared Mahalanobis distance $d(\vec x_q, \vec x_i)$ between $\vec x_q$ and the blob center $\vec x_i$:
\begin{equation}
\setlength{\abovedisplayskip}{4pt}
\setlength{\belowdisplayskip}{4pt}
d(\vec x_q, \vec x_i) = (\vec x_q - \vec x_i)^T (R_i \Sigma R_i^T)^{-1} (\vec x_q - \vec x_i),
\end{equation}
where $R_i$ is a $3 \times 3$ rotation matrix computed by angles $\boldsymbol{\theta}_i$,
\begin{equation}
    \Sigma = c \begin{pmatrix}
      a_1 & 0   & 0\\
        0 & a_2 & 0\\
        0 & 0   & a_3
    \end{pmatrix}
\end{equation}
is a diagonal matrix with aspect ratio $\vec a_i$ along its diagonal  and $c$ is a small scalar to control the sharpness of blob edges. We calculate the final density value at query point $\vec x_q$ w.r.t. blob center $\vec x_i$ by
\begin{equation}
\setlength{\abovedisplayskip}{4pt}
\setlength{\belowdisplayskip}{4pt}
\sigma(\vec x_q, \vec x_i) = \text{sigmoid}(s_i - d(\vec x_q, \vec x_i)),
\end{equation}
where the sigmoid operation ensures a density value between 0 and 1, and the scale factor $s_i$ controls the size of the blob by adjusting the magnitude of density. 
For simplicity, we assume a clear ordering of all $M$ blobs along the depth axis, which is specified by their center position $\vec x_i$. This assumption implies that difficult blob intersections will not greatly influence the results and we therefore evaluate M density volumes independently.

While the density value changes depending on $\vec x_q$, we predefine the structure feature vector $\vec f_i$ representing the feature value of blob $b_i$ to stay constant, allowing us to derive $M$ spatially-invariant feature volumes in total. This is different than other 3D modeling designs, which obtain feature values depending on the query points. We justify this design choice by referring to the high memory consumption a spatially-variant feature volume computation would imply for our multi-blob layout. Moreover, we counteract a possible degradation of output quality by choosing $\vec f_i$ to consist of 768 features values, while other works doing spatially-variant computation usually only have 32. 
\label{sec:vr}
We follow the method in \cite{mildenhall2020nerf,xu2021volumegan} to render the density volume of blob $b_i$ in a front-to-back manner and choose $N$ sample points along each ray to compute the weighted feature vector $\vec f_i$ at a single pixel position $\vec x_p$ in feature map $\vec M_i$. Having a constant feature vector $\vec f_i$ per blob $b_i$ allows us to compute a scalar weight per pixel first and only performing a single multiplication per pixel afterwards by applying
\begin{equation}
\vec M_i(\vec x_p) = O_i(\vec x_p) \vec f_i
\end{equation}
with
\begin{equation}
O_i(\vec x_p) = \sum^{N}_{k=1}T_k (1 - \exp(-\sigma_k \delta_k))
\end{equation}
and
\begin{equation}
\setlength{\abovedisplayskip}{4pt}
\setlength{\belowdisplayskip}{4pt}
T_k = \exp(-\sum^{k - 1}_{j = 1} \sigma_j \delta_j),
\end{equation}
where $O_i$ is the opacity map of $b_i$, $\delta_k = ||\vec x_{k+1} - \vec x_{k}||_2$ is the distance between adjacent sample points, $\sigma_k = \sigma(\vec x_k, \vec x_i)$ denotes the density at sample point $\vec x_k$ w.r.t. blob center $\vec x_i$ and $T_k$ is a measure for the accumulated transmittance along the ray between $\vec x_1$ and $\vec x_k$. 


\subsubsection{Occlusions and foreshortening effect}
\label{sec:occlusions}
BlobGAN \cite{epstein2022blobgan} resolves occlusions between objects using alpha compositing~\cite{Porter1984alphacomp} but only used the assigned index of each blob to identify the order. After training, changing this order is only possible by a permutation of indices. Since we also find a clear back-to-front ordering of blobs to do alpha compositing but explicitly model the depth, BlobGAN can be seen as a special case of our proposed method. \par

Specifically, given the camera intrinsics and extrinsics, we identify the direction of a ray traced through each pixel and calculate the depth of each blob in the camera coordinate system. 
In our approach, the depth order can naturally change in case the camera or the blobs are freely moved in space.
After identifying the depth ordering of blobs along a ray, the vector-value at each pixel $x_p$ on the final feature grid $\vec F$ is given by:
\begin{equation}
\setlength{\abovedisplayskip}{4pt}
\setlength{\belowdisplayskip}{4pt}
\vec F(\vec x_p) = \sum_{i=1}^{M} O_i(\vec x_p) \left( \prod^{M}_{j = i + 1} \left(1 - O_j(\vec x_p)\right) \right) \vec f_i.
\end{equation}
We argue that the benefits of our 3D design are not only to give the users more freedom to determine occlusions, but we explicitly control the 3D location of objects during inference. More importantly, we apply a perspective projection from 3D to 2D, while recent work \cite{epstein2022blobgan} only applies a parallel projection. In the latter case, there is no foreshortening effect in the generated images. Even when changing the implicit depth by permuting indices in BlobGAN, the projected object size will not vary because of the parallel projection, while in our method the size of the projection is inherently connected to the depth.

\subsection{Image Generation}
In BlobGAN, the resolution $L_H$ of the computed opacity map is chosen to be equal to the resolution of the output RGB image. Then $2 \times$ downsampling is applied consecutively to get a sequence of opacity maps at lower resolutions $L_1, L_2, ..., L_H$. BlobGAN replaces the constant input in the StyleGAN2 generator with the feature grid $\vec F$ at $L_1$ by splatting the feature vectors onto the opacity maps and then summing over these opacity maps. Also, the affine-transformed style vector which modulates the convolution weights in the StyleGAN2 generator block is replaced with a collection of hierarchical spatial-variant style grids $\vec S = \{\vec S_{L_1}, \vec S_{L_2}, ..., \vec S_{L_H}\}$. Specifically, they splat the opacity maps of hierarchical spatial resolutions with style vectors $\boldsymbol{\phi}_i$ to get the style grids of hierarchical spatial resolutions. They do a per-pixel affine transformation on the normalized style grid and then multiply it with the input feature grid before applying the normalized convolutional operator. As in StyleGAN2, the low spatial resolution input feature grid is transformed into a high-resolution RGB image by a sequence of convolutional upsampling blocks. For each feature grid at a certain resolution, a style grid of the same spatial resolution controls the modulation of the corresponding convolutional block. In BlobGAN, to output a $L_H \times L_H$ RGB image, they compute the opacity map at a resolution of $L_H \times L_H$. However, we find that instead of computing each opacity map at $L_H \times L_H$, we advocate directly computing it at a lower resolution. We choose $L_{H - 1} \times L_{H - 1}$ to find a balance between detail preservation and computational efficiency. For the remaining lower-resolution opacity maps, we keep using downsampling; for the higher-resolution opacity maps, \eg $L_H \times L_H$, we design an upsampling block to obtain them. \par
Given input $X \in R^{H \times W \times C}$, to obtain the output $X_{\text{up}} \in R^{2H \times 2W \times C}$, we have:
\begin{equation}
\setlength{\abovedisplayskip}{2pt}
\setlength{\belowdisplayskip}{2pt}
X_{\text{up}} = \text{Bilinear}(X) + \text{PixelShuffle}(\text{ModConv}(X)),
\end{equation}
where ModConv is a $1\times1$ convolution modulated by the feature vector of each blob, PixelShuffle is an upsampling operator proposed in \cite{shi2016pixelshuffle}. We initialize the weight of the ModConv to $0$, such that the upsampling block reduces to a bilinear operator at the beginning. In fact, in 3D image synthesis tasks, it could be very expensive to directly compute the opacity maps $O = \{O_1, O_2, ..., O_M\}$ at $L_H \times L_H$ as in BlobGAN, which means we have to do volume rendering at $L_H \times L_H$ to obtain the feature maps $\vec M = \{\vec M_1, \vec M_2, ..., \vec M_M\}$. As the third dimension is modeled, we have to decrease the spatial resolution of the volume rendering step to save memory. The difference between the computation of the hierarchical opacity maps in BlobGAN and BlobGAN-3D and the design of the upsampler module are shown in the supplementary materials.

\subsection{Multiview Control}
\subsubsection{Viewpoint disentanglement in StyleGAN2}
As demonstrated in \cite{harkonen2020ganspace,leimk2021freestylegan,abdal2021styleflow,tewari2020stylerig}, 
StyleGAN2 can disentangle viewpoints in the early layers. Also, as is the design in BlobGAN, the input feature grid of the generator is at a low resolution of $L_1 \times L_1$, which could lead to a loss of spatial details when doing the upsampling to obtain the final $L_H \times L_H$ RGB image. In a 3D-aware setting, this loss may also cause multiview inconsistency. Therefore, to get a better multiview consistency effect, we add a skip connection of the input to the output of the first two convolutional blocks in the synthesis network. We find this can help multiview consistency to some extent, but there is still room for improvement. We subsequently use depth supervision to enhance the multiview consistency. 

\subsubsection{Depth supervision}
Unlike other 3D-aware image synthesis networks, where the scene is modeled as a whole, blobs decompose the scene into different parts. This blob design brings extra challenges in multiview rendering. In BlobGAN-3D, when moving the camera, all the blobs rendered in the image will move at the same time, compared to the blob-level editing where only one or at most several blobs are moving at the same time. It is much more challenging to render a realistic scene when all the blobs can appear in different locations on the image plane, as once the depths of some blobs in 3D space are wrong, they may unexpectedly occlude other blobs when moving, resulting in multiview inconsistency and other artifacts. Thus, it is critical to correctly model the depths of the blobs to have good multiview consistency. We find that it could be hard for the network to optimize the depths for all the blobs in an unsupervised way. So, we include an off-the-shelf pre-trained depth estimation network to help the network correct the depth it learns for each blob. We apply an extra MSE loss $l_{\text{depth}}$ between the learned depth of the centroid of each blob and the predicted depth of that centroid projected onto the image using the depth estimator. In detail, for the blob centroid $(x_s, y_s, z_s)$ in the world coordinate, which is projected to the location $(u, v)$ on the image, we have:
\begin{equation}
l_{\text{depth}} = ||z_s - D(u, v)||^2,
\label{eq:depth_loss}
\end{equation}
where $D$ is the depth estimator. We also find that it is not helpful to start training with the multiview rendering unless the learned depths of the blobs are correct. Thus, we employ a two-stage training strategy:
\begin{enumerate}[label=\roman*).]
\item Fixing the camera pose as front-view, training the network until it reaches a good texture quality.\
\item Applying the depth estimator to correct the depths of the blob centroids while rendering the image from a randomly sampled camera pose.\
\end{enumerate}

\section{Experiments}

\begin{table*}[t]
\centering
\caption{\textbf{Comparison between BlobGAN and our method.} Our method not only enables object-evel editing in BlobGAN, but also enables additional 3D effects which do not exist in BlobGAN.}
\begin{tabular}{lccccccc}
\toprule
           & Moving & Resizing & Reshaping & Restyling & Multiview & Depth control & Foreshortening \\ \hline
BlobGAN    & \ding{51}     & \ding{51}         & \ding{51}          & \ding{51}          & \ding{55}          & \ding{55}              & \ding{55}               \\
\textbf{Ours} & \ding{51}       & \ding{51}         & \ding{51}          & \ding{51}           & \ding{51}          & \ding{51}              & \ding{51}               \\ 
\bottomrule
\end{tabular}
\label{table:comparison-blobgan}
\vspace{-1.8mm}
\end{table*}
\begin{table}[h]
\caption{\textbf{Comparison of disentanglement properties.} Our method is 3D aware and also object-level disentangled. \textit{Fg/Bg} means the method can disentangle foreground and background, but cannot further disentangle individual objects. \textit{Objects} means the method can disentangle individual objects.}
\begin{tabular}{lccc}
\toprule
          & 3D-aware & Disentang. & \begin{tabular}[c]{@{}c@{}}Disentang. \\ level\end{tabular}  \\ \hline
GIRAFFE   & \ding{51}       & \ding{51}               & Objects     \\
VolumeGAN  & \ding{51}        & \ding{55}               & --                      \\
StyleNeRF  & \ding{51}        & \ding{51}               & Fg/Bg \\
EG3D       & \ding{51}        & \ding{55}               & --                      \\
BlobGAN   & \ding{55}        & \ding{51}               & Objects     \\
\textbf{Ours}      & \ding{51}       & \ding{51}               & Objects                      \\ 
\bottomrule
\end{tabular}
\label{table:comparison}
\vspace{-4mm}
\end{table}
\subsection{Settings}
\vspace{-1mm}
\paragraph{Dataset.} We use two indoor multi-object scene datasets, \textit{Bedroom} and \textit{Conference room}, from LSUN \cite{fisher2015lsun} to evaluate our model. The datasets contain $3$M and $200$K images, respectively. We train at $256 \times 256$ resolution. \par 
\paragraph{Baselines.} We compare our method with BlobGAN\cite{epstein2022blobgan},  GIRAFFE\cite{niemeyer2021giraffe}, VolumeGAN\cite{xu2021volumegan}, StyleNeRF\cite{gu2022stylenerf} ,and EG3D\cite{chan2022eg3d}. We build our implementation on top of BlobGAN. The differences between BlobGAN and BlobGAN-3D are shown in \cref{table:comparison-blobgan}. GIRAFFE is a 3D-aware GAN with multi-object editing capabilities on synthetic datasets. VolumeGAN, StyleNeRF and EG3D are 3D-aware GANs without multi-object editability. A comparison of the properties of our method and the baselines is shown in \cref{table:comparison}. \par
\paragraph{Metric.} We use FID and KID \cite{binkowski2018kid}. We use $50$K real images and $50$K fake images to compute both FID and KID using the method proposed in \cite{parmar2021clean-fid}.\par
\paragraph{Implementation details.} For both datasets, we empirically identified a suitable number of blobs $M=10$. A smaller number results in the decrease of quality, while a larger number implicates a higher computational complexity. The output image resolution is $L_H \times L_H = 256 \times 256$. We perform volume rendering at $L_{H - 1} \times L_{H - 1} = 128 \times 128$. The resolution of the feature grid we feed into the synthesis network is $L_1 \times L_1 = 16 \times 16$. The off-the-shelf depth estimator we adopt is from \cite{ranftl2019midas}. We randomly sample $32$ points along each ray of the camera. More details can be found in Supplementary materials. \par

\begin{table}[htbp]
\caption{\textbf{Quantitative results of image quality on two datasets.} Our method achieves good image quality compared to the 3D GAN baselines and BlobGAN.}
\begin{tabular}{lcccc}
\toprule
\multicolumn{1}{c}{} & \multicolumn{2}{c}{Bedroom} & \multicolumn{2}{c}{Conference room} \\
\multicolumn{1}{c}{} & FID $\downarrow$            & KID $\downarrow$      & FID $\downarrow$              & KID $\downarrow$            \\ \hline
GIRAFFE             & 49.62        & 0.046        & 73.99                & 0.066              \\
VolumeGAN          & 17.3             & --        & 43.92                & 0.032              \\
StyleNeRF           & 13.50         & 0.011        & 23.09                & 0.018              \\
EG3D             & 10.25              & 0.0071        & 15.08                & 0.0099              \\
\textbf{Ours}                 & \textbf{4.16}         & \textbf{0.0023}        & \textbf{7.80}                & \textbf{0.0039}              \\ \hline
\hline
BlobGAN              & 3.76            & 0.0021        & 6.76              & 0.0033              \\
\begin{tabular}[c]{@{}l@{}}BlobGAN  \\ ($25000$K)\end{tabular}      & 11.5            & 0.0084        & 11.01                & 0.0064             \\
\textbf{Ours} ($25000$K)        & 11.7            & 0.0096        & 15.38                & 0.0100              \\
\bottomrule
\end{tabular}
\label{table:fid}
\end{table}
\begin{table}[h]
\centering
\caption{\textbf{FID after editing the scene.} \textit{\# blobs} means the number of the blobs being moved. We randomly choose which blobs to move and move them by 0.2 in normalized coordinate space in the horizontal direction.}
\begin{tabular}{lcccccc}
\toprule
        & \multicolumn{3}{c}{Bedroom} & \multicolumn{3}{c}{Conf. room} \\
\# blobs   & 1       & 3       & 5       & 1          & 3          & 5         \\ \hline
BlobGAN  & 4.04         & 4.24        & 4.68        & 6.94           & 7.22           & 7.61          \\
\textbf{Ours}    & 4.11        & 4.13        & 4.23        & 7.80          & 8.09           & 8.24          \\ 
\bottomrule
\end{tabular}
\label{table:FID-editing}
\vspace{-1mm}
\end{table}

\begin{figure}[t]
  \centering
  \begin{subfigure}[t]{0.24\linewidth}
    \includegraphics[width=\linewidth]{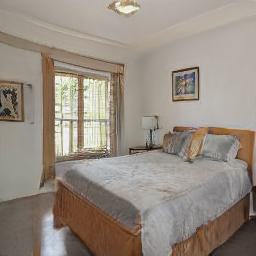}
\end{subfigure}
  \begin{subfigure}[t]{0.24\linewidth}
    \includegraphics[width=\linewidth]{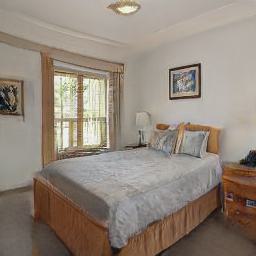}
\end{subfigure}
  \begin{subfigure}[t]{0.24\linewidth}
    \includegraphics[width=\linewidth]{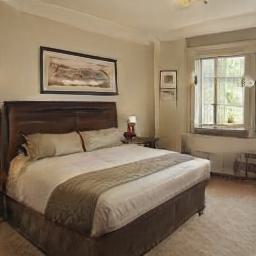}
\end{subfigure}
  \begin{subfigure}[t]{0.24\linewidth}
    \includegraphics[width=\linewidth]{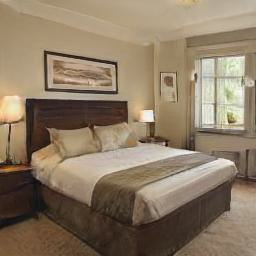}
\end{subfigure}
\quad
\centering
  \begin{subfigure}[t]{0.24\linewidth}
    \includegraphics[width=\linewidth]{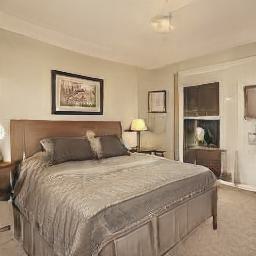}
\end{subfigure}
  \begin{subfigure}[t]{0.24\linewidth}
    \includegraphics[width=\linewidth]{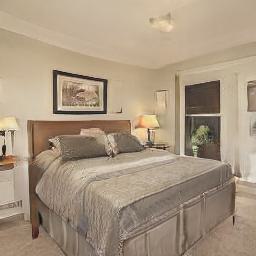}
\end{subfigure}
  \begin{subfigure}[t]{0.24\linewidth}
    \includegraphics[width=\linewidth]{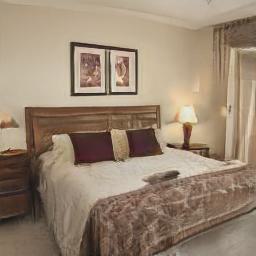}
\end{subfigure}
  \begin{subfigure}[t]{0.24\linewidth}
    \includegraphics[width=\linewidth]{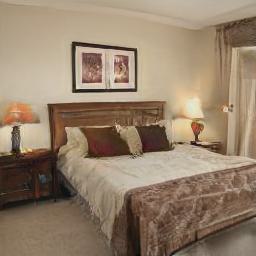}
\end{subfigure}
\caption{\textbf{Foreshortening effect.} By adjusting the horizontal position and the depth of the blob at the same time, we achieve a foreshortening effect: when the bed is moving away from the camera, its size in the image becomes smaller.}
\label{fig:foreshortening}
\vspace{-2mm}
\end{figure}

\begin{figure*}[htbp]
  \centering
  \begin{subfigure}[t]{0.12\linewidth}
    \includegraphics[width=\linewidth]{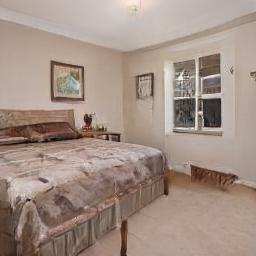}
\end{subfigure}
\begin{subfigure}[t]{0.12\linewidth}
    \includegraphics[width=\linewidth]{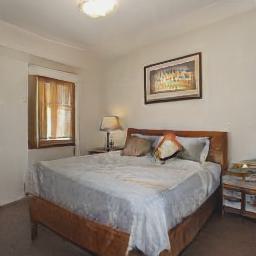}
\end{subfigure}
  \begin{subfigure}[t]{0.12\linewidth}
    \includegraphics[width=\linewidth]{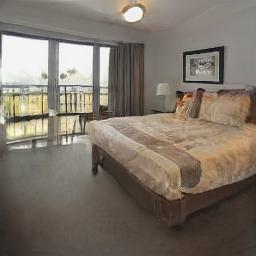}
\end{subfigure}
  \begin{subfigure}[t]{0.12\linewidth}
    \includegraphics[width=\linewidth,height=\linewidth]{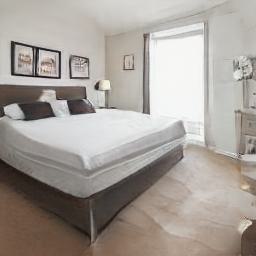}
\end{subfigure}
  \begin{subfigure}[t]{0.12\linewidth}
    \includegraphics[width=\linewidth,height=\linewidth]{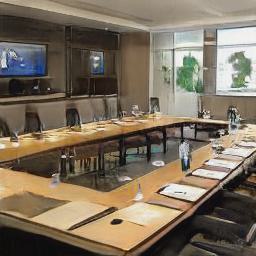}
\end{subfigure}
  \begin{subfigure}[t]{0.12\linewidth}
    \includegraphics[width=\linewidth,height=\linewidth]{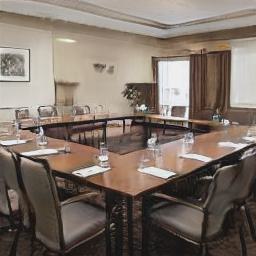}
\end{subfigure}
  \begin{subfigure}[t]{0.12\linewidth}
    \includegraphics[width=\linewidth]{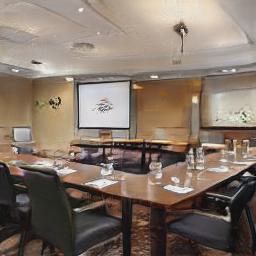}
\end{subfigure}
  \begin{subfigure}[t]{0.12\linewidth}
    \includegraphics[width=\linewidth]{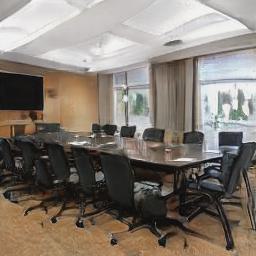}
\end{subfigure}
\quad
  \begin{subfigure}[t]{0.12\linewidth}
       \begin{overpic}[width=\textwidth]{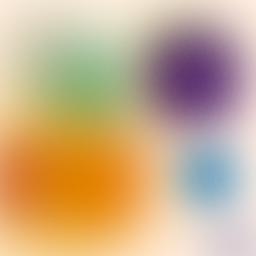}
\linethickness{1.5pt}
\put(35,30){\color[RGB]{255,255,255}\vector(1,0){40}}
\end{overpic}
\end{subfigure}
  \begin{subfigure}[t]{0.12\linewidth}
   \begin{overpic}[width=\textwidth]{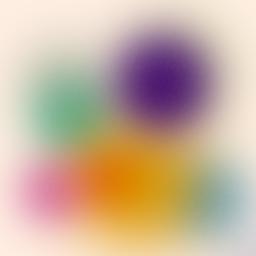}
\linethickness{1.5pt}
\put(55,30){\color[RGB]{255,255,255}\vector(1,0){40}}
\end{overpic}
\end{subfigure}
  \begin{subfigure}[t]{0.12\linewidth}
  \begin{overpic}[width=\textwidth]{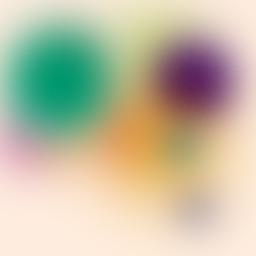}
\linethickness{1.5pt}
\put(77,68){\color[RGB]{255,255,255}\vector(1,0){25}}
\end{overpic}  
\end{subfigure}
  \begin{subfigure}[t]{0.12\linewidth}
   \begin{overpic}[width=\textwidth]{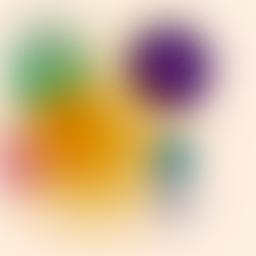}
\linethickness{1.5pt}
\put(67,72){\color[RGB]{255,255,255}\vector(1,0){30}}
\end{overpic}  
\end{subfigure}
  \begin{subfigure}[t]{0.12\linewidth}
  \begin{overpic}[width=\textwidth]{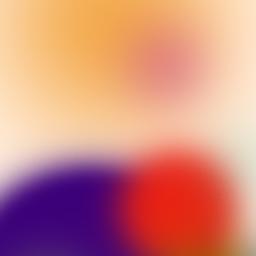}
\linethickness{1.5pt}
\put(71,21){\color[RGB]{255,255,255}\vector(1,0){30}}
\end{overpic}  
\end{subfigure}
  \begin{subfigure}[t]{0.12\linewidth}
  \begin{overpic}[width=\textwidth]{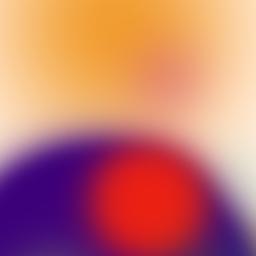}
\linethickness{1.5pt}
\put(58,21){\color[RGB]{255,255,255}\vector(1,0){30}}
\end{overpic}  
\end{subfigure}
  \begin{subfigure}[t]{0.12\linewidth}
  \begin{overpic}[width=\textwidth]{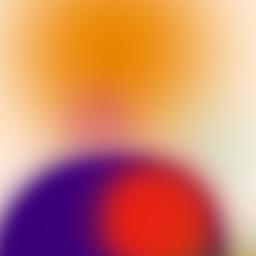}
\linethickness{1.5pt}
\put(40,58){\color[RGB]{255,255,255}\vector(1,0){30}}
\end{overpic}  
\end{subfigure}
  \begin{subfigure}[t]{0.12\linewidth}
  \begin{overpic}[width=\textwidth]{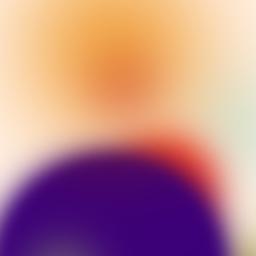}
\linethickness{1.5pt}
\put(48,70){\color[RGB]{255,255,255}\vector(1,0){30}}
\end{overpic}  
\end{subfigure}
\quad
  \begin{subfigure}[t]{0.12\linewidth}
    \includegraphics[width=\linewidth]{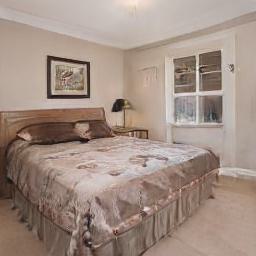}
\end{subfigure}
  \begin{subfigure}[t]{0.12\linewidth}
    \includegraphics[width=\linewidth]{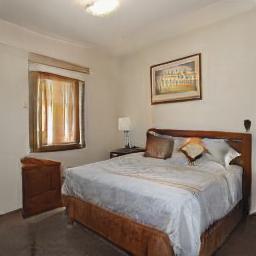}
\end{subfigure}
  \begin{subfigure}[t]{0.12\linewidth}
    \includegraphics[width=\linewidth]{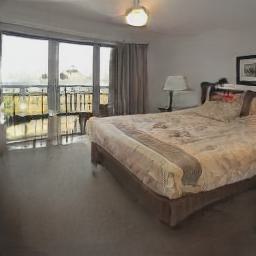}
\end{subfigure}
  \begin{subfigure}[t]{0.12\linewidth}    \includegraphics[width=\linewidth,height=\linewidth]{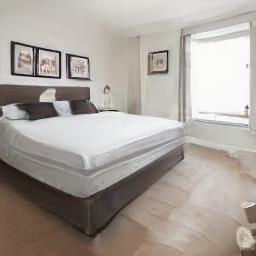}
\end{subfigure}
  \begin{subfigure}[t]{0.12\linewidth}
    \includegraphics[width=\linewidth,height=\linewidth]{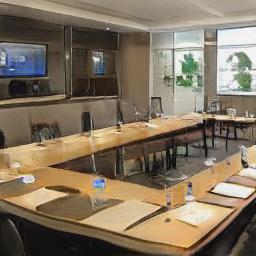}
\end{subfigure}
  \begin{subfigure}[t]{0.12\linewidth}
    \includegraphics[width=\linewidth,height=\linewidth]{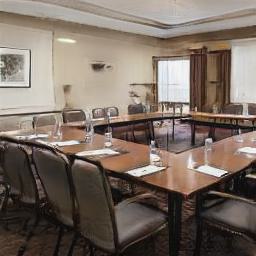}
\end{subfigure}
  \begin{subfigure}[t]{0.12\linewidth}
    \includegraphics[width=\linewidth]{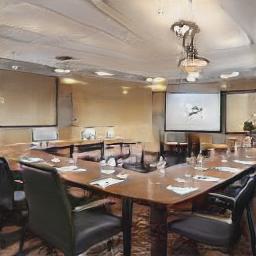}
\end{subfigure}
  \begin{subfigure}[t]{0.12\linewidth}
    \includegraphics[width=\linewidth]{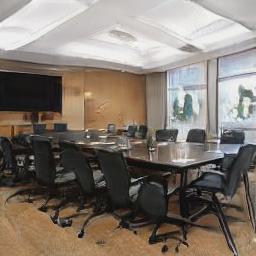}
\end{subfigure}
\caption{\textbf{Moving an object in the scene.} The first row shows the original generated images; the second row shows the corresponding blob layout maps with marks demonstrating the moving direction. The last row shows the synthesized images after moving one blob.}
\label{fig:moving}
\end{figure*}

\begin{figure*}[t]
  \centering
  \begin{subfigure}[t]{0.12\linewidth}
    \includegraphics[width=\linewidth]{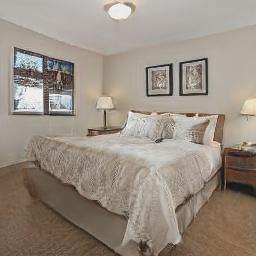}
\end{subfigure}
  \begin{subfigure}[t]{0.12\linewidth}
    \includegraphics[width=\linewidth]{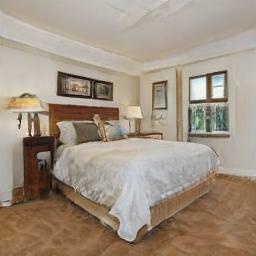}
\end{subfigure}
  \begin{subfigure}[t]{0.12\linewidth}
    \includegraphics[width=\linewidth]{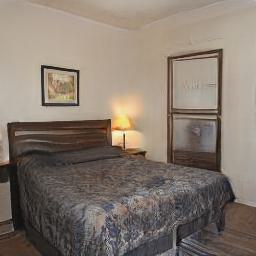}
\end{subfigure}
  \begin{subfigure}[t]{0.12\linewidth}
    \includegraphics[width=\linewidth,height=\linewidth]{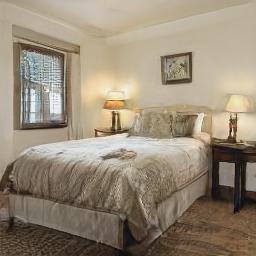}
\end{subfigure}
  \begin{subfigure}[t]{0.12\linewidth}
    \includegraphics[width=\linewidth,height=\linewidth]{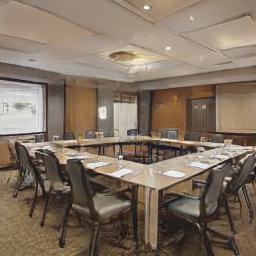}
\end{subfigure}
  \begin{subfigure}[t]{0.12\linewidth}
    \includegraphics[width=\linewidth,height=\linewidth]{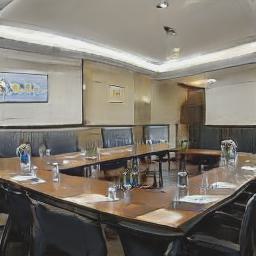}
\end{subfigure}
  \begin{subfigure}[t]{0.12\linewidth}
    \includegraphics[width=\linewidth]{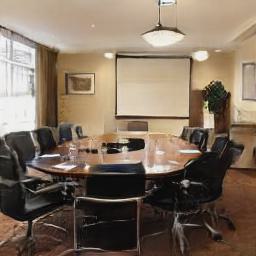}
\end{subfigure}
  \begin{subfigure}[t]{0.12\linewidth}
    \includegraphics[width=\linewidth]{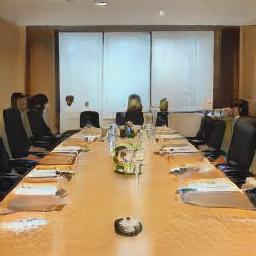}
\end{subfigure}
\quad
  \begin{subfigure}[t]{0.12\linewidth}
  \begin{overpic}[width=\textwidth]{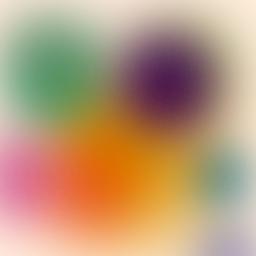}
\linethickness{1.5pt}
\put(38,40){\color[RGB]{255,255,255}\line(1,-1){20}}
\put(58,40){\color[RGB]{255,255,255}\line(-1,-1){20}}
\end{overpic} 
\end{subfigure}
  \begin{subfigure}[t]{0.12\linewidth}
  \begin{overpic}[width=\textwidth]{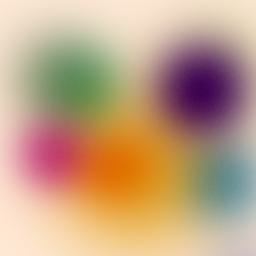}
\linethickness{1.5pt}
\put(43,40){\color[RGB]{255,255,255}\line(1,-1){20}}
\put(63,40){\color[RGB]{255,255,255}\line(-1,-1){20}}
\end{overpic} 
\end{subfigure}
  \begin{subfigure}[t]{0.12\linewidth}
  \begin{overpic}[width=\textwidth]{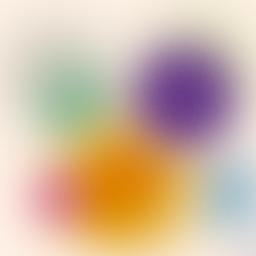}
\linethickness{1.5pt}
\put(66,70){\color[RGB]{255,255,255}\line(1,-1){15}}
\put(81,70){\color[RGB]{255,255,255}\line(-1,-1){15}}
\end{overpic} 
\end{subfigure}
  \begin{subfigure}[t]{0.12\linewidth}
  \begin{overpic}[width=\textwidth]{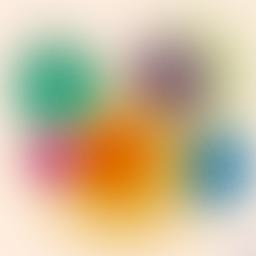}
\linethickness{1.5pt}
\put(62,77){\color[RGB]{255,255,255}\line(1,-1){15}}
\put(77,77){\color[RGB]{255,255,255}\line(-1,-1){15}}
\end{overpic} 
\end{subfigure}
  \begin{subfigure}[t]{0.12\linewidth}
  \begin{overpic}[width=\textwidth]{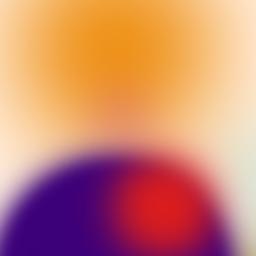}
\linethickness{1.5pt}
\put(35,40){\color[RGB]{255,255,255}\line(1,-1){30}}
\put(63,40){\color[RGB]{255,255,255}\line(-1,-1){30}}
\end{overpic} 
\end{subfigure}
  \begin{subfigure}[t]{0.12\linewidth}
  \begin{overpic}[width=\textwidth]{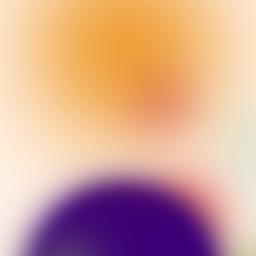}
\linethickness{1.5pt}
\put(38,32){\color[RGB]{255,255,255}\line(1,-1){28}}
\put(64,32){\color[RGB]{255,255,255}\line(-1,-1){28}}
\end{overpic} 
\end{subfigure}
  \begin{subfigure}[t]{0.12\linewidth}
  \begin{overpic}[width=\textwidth]{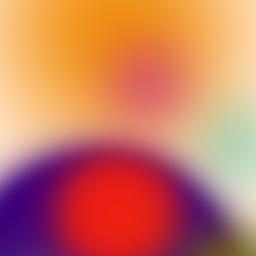}
\linethickness{1.5pt}
\put(48,78){\color[RGB]{255,255,255}\line(1,-1){20}}
\put(68,78){\color[RGB]{255,255,255}\line(-1,-1){20}}
\end{overpic} 
\end{subfigure}
  \begin{subfigure}[t]{0.12\linewidth}
  \begin{overpic}[width=\textwidth]{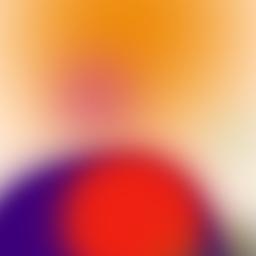}
\linethickness{1.5pt}
\put(28,72){\color[RGB]{255,255,255}\line(1,-1){20}}
\put(48,72){\color[RGB]{255,255,255}\line(-1,-1){20}}
\end{overpic} 
\end{subfigure}
\quad
  \begin{subfigure}[t]{0.12\linewidth}
    \includegraphics[width=\linewidth]{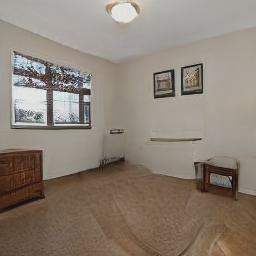}
\end{subfigure}
  \begin{subfigure}[t]{0.12\linewidth}
    \includegraphics[width=\linewidth]{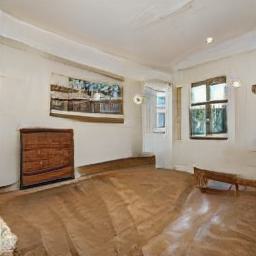}
\end{subfigure}
  \begin{subfigure}[t]{0.12\linewidth}
    \includegraphics[width=\linewidth]{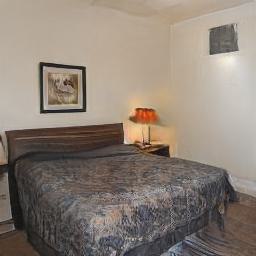}
\end{subfigure}
  \begin{subfigure}[t]{0.12\linewidth}    \includegraphics[width=\linewidth,height=\linewidth]{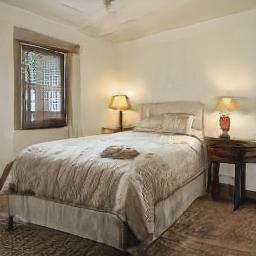}
\end{subfigure}
  \begin{subfigure}[t]{0.12\linewidth}
    \includegraphics[width=\linewidth,height=\linewidth]{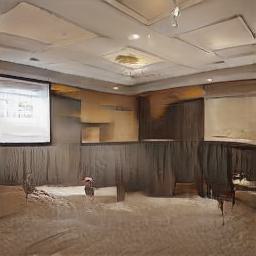}
\end{subfigure}
  \begin{subfigure}[t]{0.12\linewidth}
    \includegraphics[width=\linewidth,height=\linewidth]{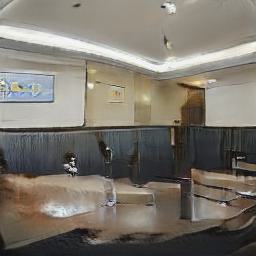}
\end{subfigure}
  \begin{subfigure}[t]{0.12\linewidth}
    \includegraphics[width=\linewidth]{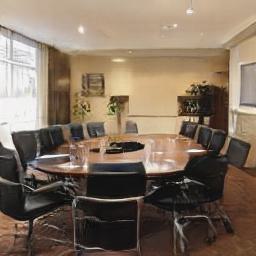}
\end{subfigure}
  \begin{subfigure}[t]{0.12\linewidth}
    \includegraphics[width=\linewidth]{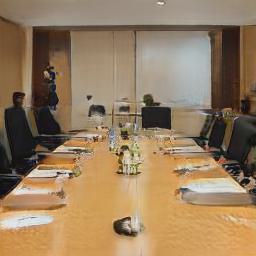}
\end{subfigure}
\caption{\textbf{Removing an object in the scene.} The first row shows the originally generated images; the second row shows the corresponding blob layout maps with marks demonstrating the removed blob. The last row shows the synthesized images after removing one blob.}
\label{fig:resizing}
\vspace{-3mm}
\end{figure*}

\begin{figure}[t]
\vspace{-1mm}
  \centering
  \begin{subfigure}[t]{0.24\linewidth}
    \includegraphics[width=\linewidth]{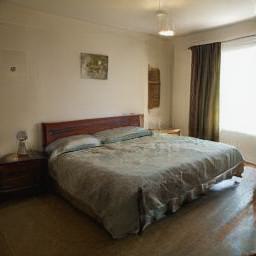}
\end{subfigure}
  \begin{subfigure}[t]{0.24\linewidth}
    \includegraphics[width=\linewidth]{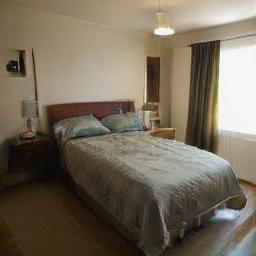}
\end{subfigure}
  \begin{subfigure}[t]{0.24\linewidth}
    \includegraphics[width=\linewidth]{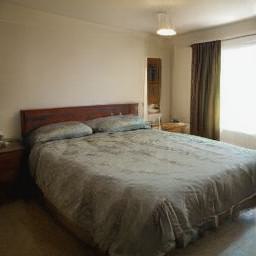}
\end{subfigure}
  \begin{subfigure}[t]{0.24\linewidth}
    \includegraphics[width=\linewidth]{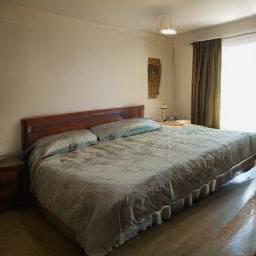}
\end{subfigure}
\quad
\centering
  \begin{subfigure}[t]{0.24\linewidth}
    \includegraphics[width=\linewidth]{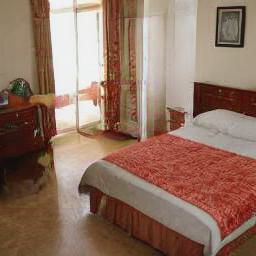}
\end{subfigure}
  \begin{subfigure}[t]{0.24\linewidth}
    \includegraphics[width=\linewidth]{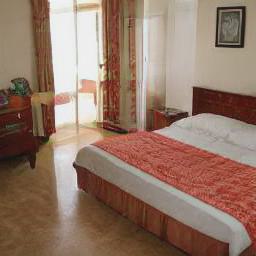}
\end{subfigure}
  \begin{subfigure}[t]{0.24\linewidth}
    \includegraphics[width=\linewidth]{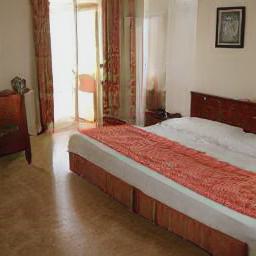}
\end{subfigure}
  \begin{subfigure}[t]{0.24\linewidth}
    \includegraphics[width=\linewidth]{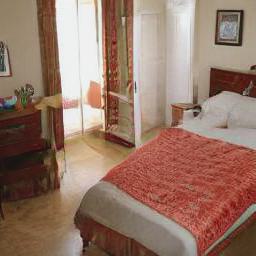}
\end{subfigure}
\caption{\textbf{Changing the aspect ratio of the blobs.} Different shapes of the blobs are obtained by changing the aspect ratio of the blob which represents the bedroom.}
\label{fig:aspect}
\vspace{-4mm}
\end{figure}

\begin{figure*}[!h]
  \centering
  \begin{minipage}[t]{0.05\linewidth}
  \centering
  \rotatebox{90}{\ \ \ \ \ \ \ GIRAFFE}
  \end{minipage}
  \begin{subfigure}[t]{0.14\linewidth}
    \includegraphics[width=\linewidth]{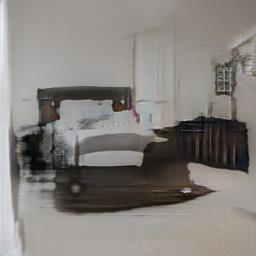}
\end{subfigure}
  \begin{subfigure}[t]{0.14\linewidth}
    \includegraphics[width=\linewidth]{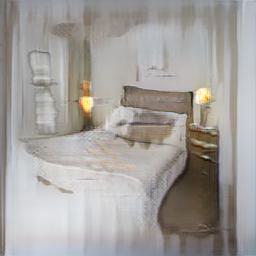}
\end{subfigure}
  \begin{subfigure}[t]{0.14\linewidth}
    \includegraphics[width=\linewidth]{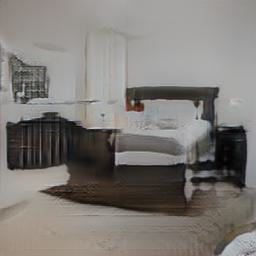}
\end{subfigure}\hspace{1mm}
  \begin{subfigure}[t]{0.14\linewidth}
\includegraphics[width=\linewidth,height=\linewidth]{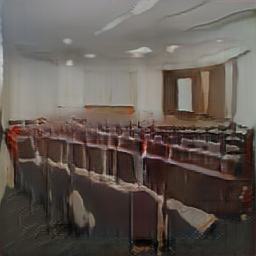}
\end{subfigure}
  \begin{subfigure}[t]{0.14\linewidth}
    \includegraphics[width=\linewidth]{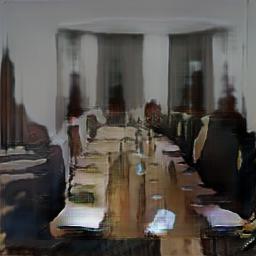}
\end{subfigure}
  \begin{subfigure}[t]{0.14\linewidth}
    \includegraphics[width=\linewidth]{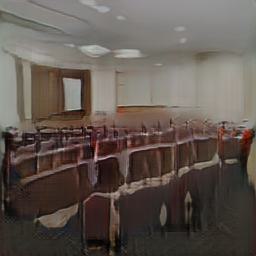}
\end{subfigure}
\quad
\\
  \begin{minipage}[t]{0.05\linewidth}
  \centering
  \rotatebox{90}{\ \ \ \ \ VolumeGAN }
  \end{minipage}
  \begin{subfigure}[t]{0.14\linewidth}
\includegraphics[width=\linewidth,height=\linewidth]{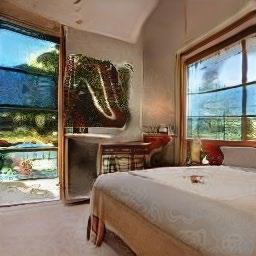}
\end{subfigure}
  \begin{subfigure}[t]{0.14\linewidth}
\includegraphics[width=\linewidth,height=\linewidth]{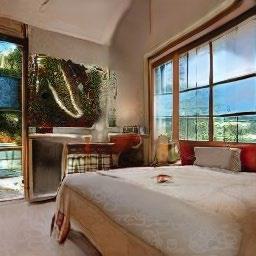}
\end{subfigure}
  \begin{subfigure}[t]{0.14\linewidth}
\includegraphics[width=\linewidth,height=\linewidth]{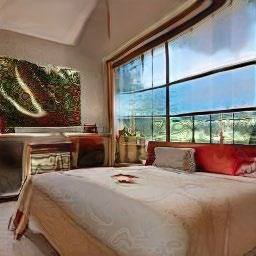}
\end{subfigure}\hspace{1mm}
  \begin{subfigure}[t]{0.14\linewidth}
\includegraphics[width=\linewidth,height=\linewidth]{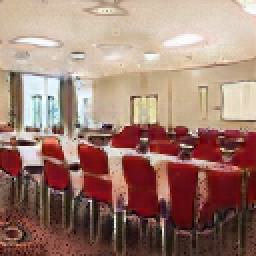}
\end{subfigure}
  \begin{subfigure}[t]{0.14\linewidth}
\includegraphics[width=\linewidth,height=\linewidth]{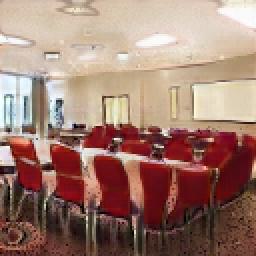}
\end{subfigure}
  \begin{subfigure}[t]{0.14\linewidth}
\includegraphics[width=\linewidth,height=\linewidth]{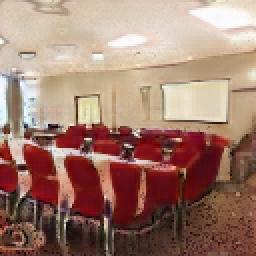}
\end{subfigure}
\quad
\\
  \begin{minipage}[t]{0.05\linewidth}
  \centering
  \rotatebox{90}{\ \ \ \ \ \ \ \ StyleNeRF}
  \end{minipage}
  \begin{subfigure}[t]{0.14\linewidth}
\includegraphics[width=\linewidth,height=\linewidth]{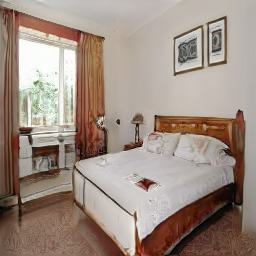}
\end{subfigure}
  \begin{subfigure}[t]{0.14\linewidth}
\includegraphics[width=\linewidth,height=\linewidth]{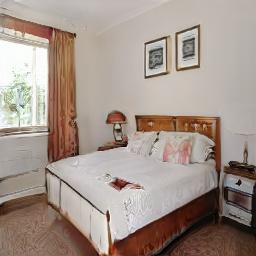}
\end{subfigure}
  \begin{subfigure}[t]{0.14\linewidth}
\includegraphics[width=\linewidth,height=\linewidth]{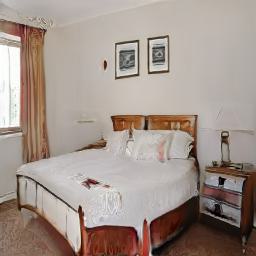}
\end{subfigure}\hspace{1mm}
  \begin{subfigure}[t]{0.14\linewidth}
\includegraphics[width=\linewidth,height=\linewidth]{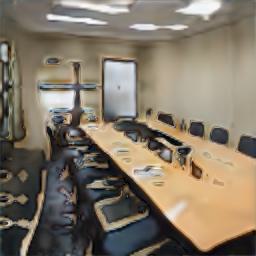}
\end{subfigure}
  \begin{subfigure}[t]{0.14\linewidth}
\includegraphics[width=\linewidth,height=\linewidth]{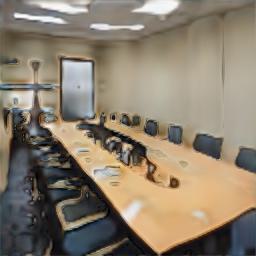}
\end{subfigure}
  \begin{subfigure}[t]{0.14\linewidth}
\includegraphics[width=\linewidth,height=\linewidth]{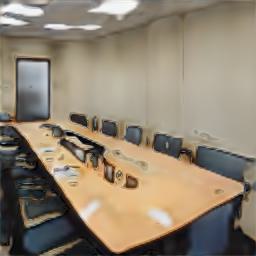}
\end{subfigure}
\quad
\\
  \begin{minipage}[t]{0.05\linewidth}
  \centering
  \rotatebox{90}{\ \ \ \ \ \ \ \ \ \ EG3D}
  \end{minipage}
  \begin{subfigure}[t]{0.14\linewidth}
\includegraphics[width=\linewidth,height=\linewidth]{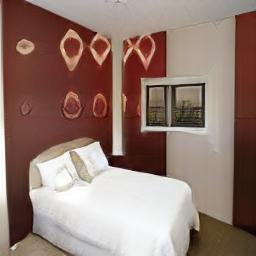}
\end{subfigure}
  \begin{subfigure}[t]{0.14\linewidth}
\includegraphics[width=\linewidth,height=\linewidth]{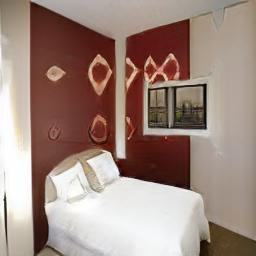}
\end{subfigure}
  \begin{subfigure}[t]{0.14\linewidth}
\includegraphics[width=\linewidth,height=\linewidth]{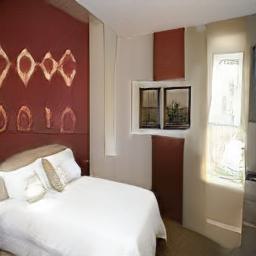}
\end{subfigure}\hspace{1mm}
  \begin{subfigure}[t]{0.14\linewidth}
\includegraphics[width=\linewidth,height=\linewidth]{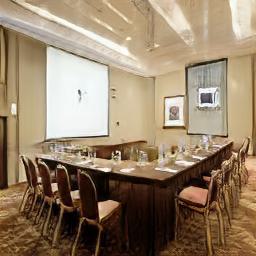}
\end{subfigure}
  \begin{subfigure}[t]{0.14\linewidth}
\includegraphics[width=\linewidth,height=\linewidth]{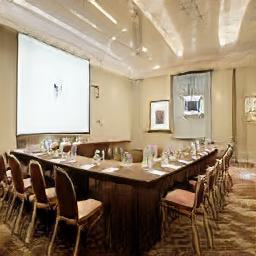}
\end{subfigure}
  \begin{subfigure}[t]{0.14\linewidth}
\includegraphics[width=\linewidth,height=\linewidth]{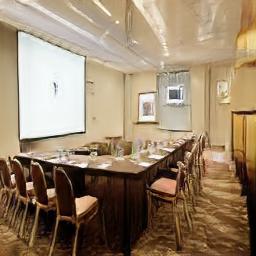}
\end{subfigure}
\quad
\\
  \begin{minipage}[t]{0.05\linewidth}
  \centering
  \rotatebox{90}{\ \ \ \ \ \ \ \ \ \ \textbf{Ours}}
  \end{minipage}
  \begin{subfigure}[t]{0.14\linewidth}
\includegraphics[width=\linewidth,height=\linewidth]{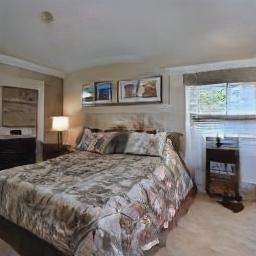}
\end{subfigure}
  \begin{subfigure}[t]{0.14\linewidth}
\includegraphics[width=\linewidth,height=\linewidth]{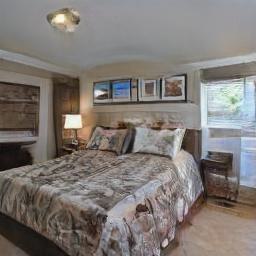}
\end{subfigure}
  \begin{subfigure}[t]{0.14\linewidth}
\includegraphics[width=\linewidth,height=\linewidth]{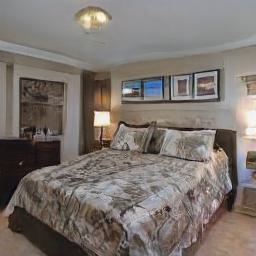}
\end{subfigure}\hspace{1mm}
  \begin{subfigure}[t]{0.14\linewidth}
\includegraphics[width=\linewidth,height=\linewidth]{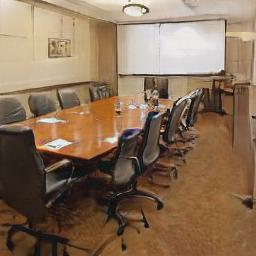}
\end{subfigure}
  \begin{subfigure}[t]{0.14\linewidth}
\includegraphics[width=\linewidth,height=\linewidth]{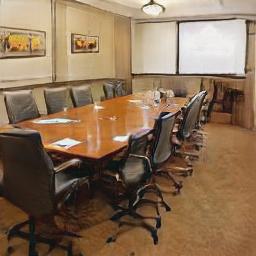}
\end{subfigure}
  \begin{subfigure}[t]{0.14\linewidth}
\includegraphics[width=\linewidth,height=\linewidth]{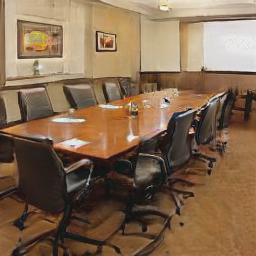}
\end{subfigure}
\quad
\\
  \begin{minipage}[t]{0.05\linewidth}
  \centering
  \rotatebox{90}{\ \ \ \ \ \ \ \ \ \ \textbf{Ours}}
  \end{minipage}
  \begin{subfigure}[t]{0.14\linewidth}
\includegraphics[width=\linewidth,height=\linewidth]{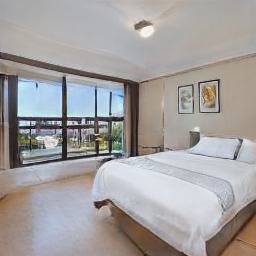}
\end{subfigure}
  \begin{subfigure}[t]{0.14\linewidth}
\includegraphics[width=\linewidth,height=\linewidth]{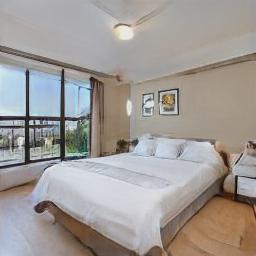}
\end{subfigure}
  \begin{subfigure}[t]{0.14\linewidth}
\includegraphics[width=\linewidth,height=\linewidth]{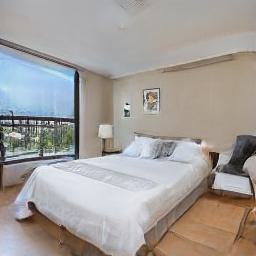}
\end{subfigure}\hspace{1mm}
  \begin{subfigure}[t]{0.14\linewidth}
\includegraphics[width=\linewidth,height=\linewidth]{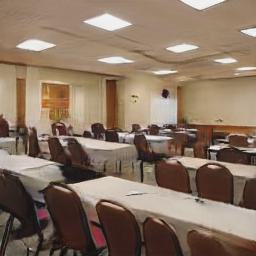}
\end{subfigure}
  \begin{subfigure}[t]{0.14\linewidth}
\includegraphics[width=\linewidth,height=\linewidth]{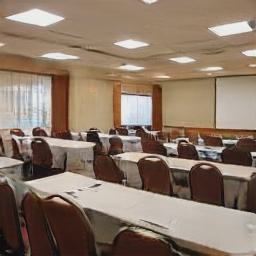}
\end{subfigure}
  \begin{subfigure}[t]{0.14\linewidth}
\includegraphics[width=\linewidth,height=\linewidth]{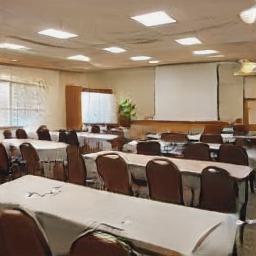}
\end{subfigure}
\quad
\\
\vspace{-2mm}
  \caption{\textbf{Multiview image synthesis results.} The results of VolumeGAN on the Bedroom dataset are borrowed from \cite{xu2021volumegan}. We disable the camera pose conditioning in EG3D as it is difficult to estimate camera poses for indoor datasets.}
  \label{fig:multiview}
\vspace{-3mm}
\end{figure*}

\begin{figure*}[htbp]
  \centering
  \begin{subfigure}[t]{0.12\linewidth}
    \includegraphics[width=\linewidth]{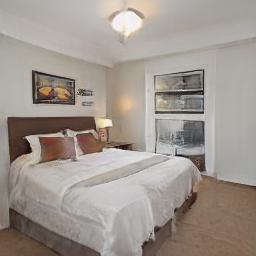}
\end{subfigure}
  \begin{subfigure}[t]{0.12\linewidth}
    \includegraphics[width=\linewidth]{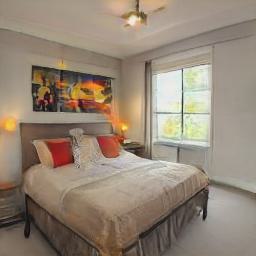}
\end{subfigure}
  \begin{subfigure}[t]{0.12\linewidth}
    \includegraphics[width=\linewidth]{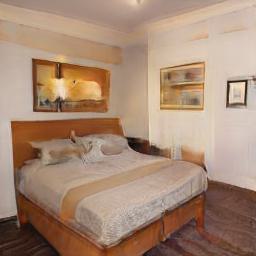}
\end{subfigure}
  \begin{subfigure}[t]{0.12\linewidth}
    \includegraphics[width=\linewidth,height=\linewidth]{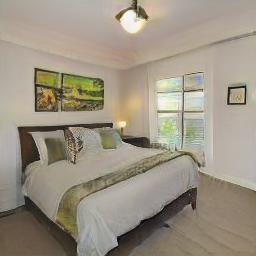}
\end{subfigure}\hspace{1mm}
  \begin{subfigure}[t]{0.12\linewidth}
    \includegraphics[width=\linewidth,height=\linewidth]{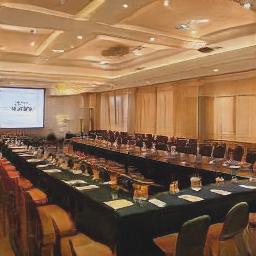}
\end{subfigure}
  \begin{subfigure}[t]{0.12\linewidth}
    \includegraphics[width=\linewidth,height=\linewidth]{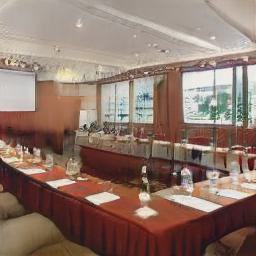}
\end{subfigure}
  \begin{subfigure}[t]{0.12\linewidth}
    \includegraphics[width=\linewidth]{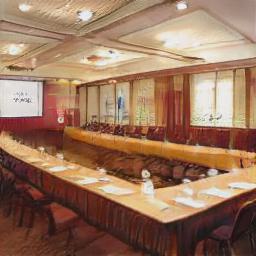}
\end{subfigure}
  \begin{subfigure}[t]{0.12\linewidth}
    \includegraphics[width=\linewidth]{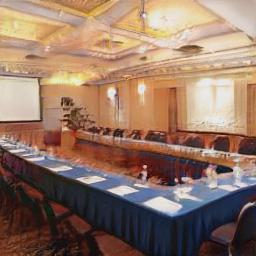}
\end{subfigure}
\quad
  \begin{subfigure}[t]{0.12\linewidth}
    \includegraphics[width=\linewidth]{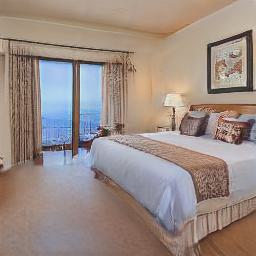}
\end{subfigure}
  \begin{subfigure}[t]{0.12\linewidth}
    \includegraphics[width=\linewidth]{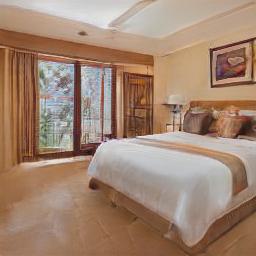}
\end{subfigure}
  \begin{subfigure}[t]{0.12\linewidth}
    \includegraphics[width=\linewidth]{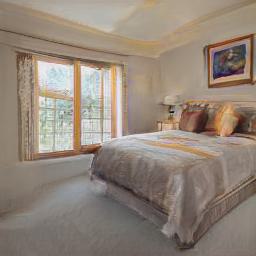}
\end{subfigure}
  \begin{subfigure}[t]{0.12\linewidth}
    \includegraphics[width=\linewidth,height=\linewidth]{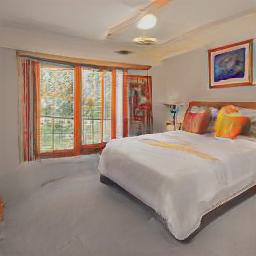}
\end{subfigure}\hspace{1mm}
  \begin{subfigure}[t]{0.12\linewidth}
    \includegraphics[width=\linewidth,height=\linewidth]{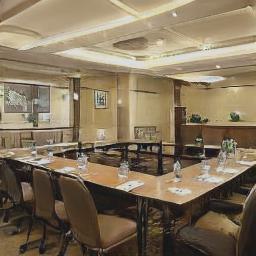}
\end{subfigure}
  \begin{subfigure}[t]{0.12\linewidth}
    \includegraphics[width=\linewidth,height=\linewidth]{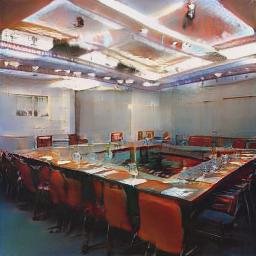}
\end{subfigure}
  \begin{subfigure}[t]{0.12\linewidth}
    \includegraphics[width=\linewidth]{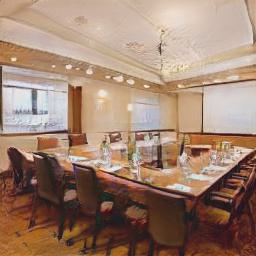}
\end{subfigure}
  \begin{subfigure}[t]{0.12\linewidth}
    \includegraphics[width=\linewidth]{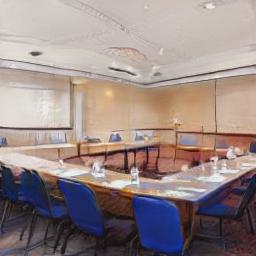}
\end{subfigure}
\caption{\textbf{Restyling of the scene.} In each row we show different decorating styles of the same bedroom or conference room by simply swapping different style vectors $\boldsymbol{\phi}_i$ for all the blobs.}
\label{fig:restyling}
\vspace{-2mm}
\end{figure*}

\subsection{Quantitative results}
We show the quantitative results of our method and baselines in \cref{table:fid}. Our method achieves comparable quality against 2D BlobGAN and 3D-aware baselines. We also provide the results of BlobGAN and our method when training for $25000$K images for a fair comparison. \par
For evaluating the quality of synthesized images after editing, we provide the FID evaluation of moving the objects in the scene in \cref{table:FID-editing}. It is interesting to see that BlobGAN-3D is relatively more stable to the editings than BlobGAN.

\subsection{Qualitative results}
\subsubsection{Multiview rendering}
\vspace{-2mm}
We show the qualitative comparison of multiview rendering between our method and baselines in \cref{fig:multiview}. Our method can achieve high image quality by synthesizing more realistic texture and fewer artifacts compared to the 3D-aware GAN baselines. 
\vspace{-3mm}
\subsubsection{Scene editing}
\vspace{-2mm}
We show the scene editing results of moving (\cref{fig:moving}), removing (\cref{fig:resizing}) and restyling (\cref{fig:restyling}). More editing results can be found in the supplementary materials. We show that our method can disentangle individual objects in the scene to apply precise edits. Despite some minor changes, most of the other objects appearing in the scene remain the same after editing while the scenes still look realistic. We additionally show the results of changing the three aspect ratio values of the blobs in 
\cref{fig:aspect} and the foreshortening effect in \cref{fig:foreshortening}.
\vspace{-3mm}

\section{Conclusions}
\vspace{-2mm}
In this work, we extend 2D GAN BlobGAN to be 3D-aware while keeping the disentanglement of the objects and can perform realistic object-level editing on complex real-world indoor datasets. We show that our model can achieve comparable image quality to the original BlobGAN and other 3D-aware GANs. Meanwhile, our method can achieve good multiview consistency while enabling object-level editing as done in BlobGAN. We also allow more editing capabilities by explicitly controlling the 3D location of the blobs in the scene and achieving a foreshortening effect. \par

{\small
\bibliographystyle{ieee_fullname}
\bibliography{PaperForReview}
}

\clearpage

\appendix

\section{More implementation details}
\subsection{Upsampler}
\begin{figure*}[t]
  \centering
   \includegraphics[width=\linewidth, trim=0 150 0 70, clip]{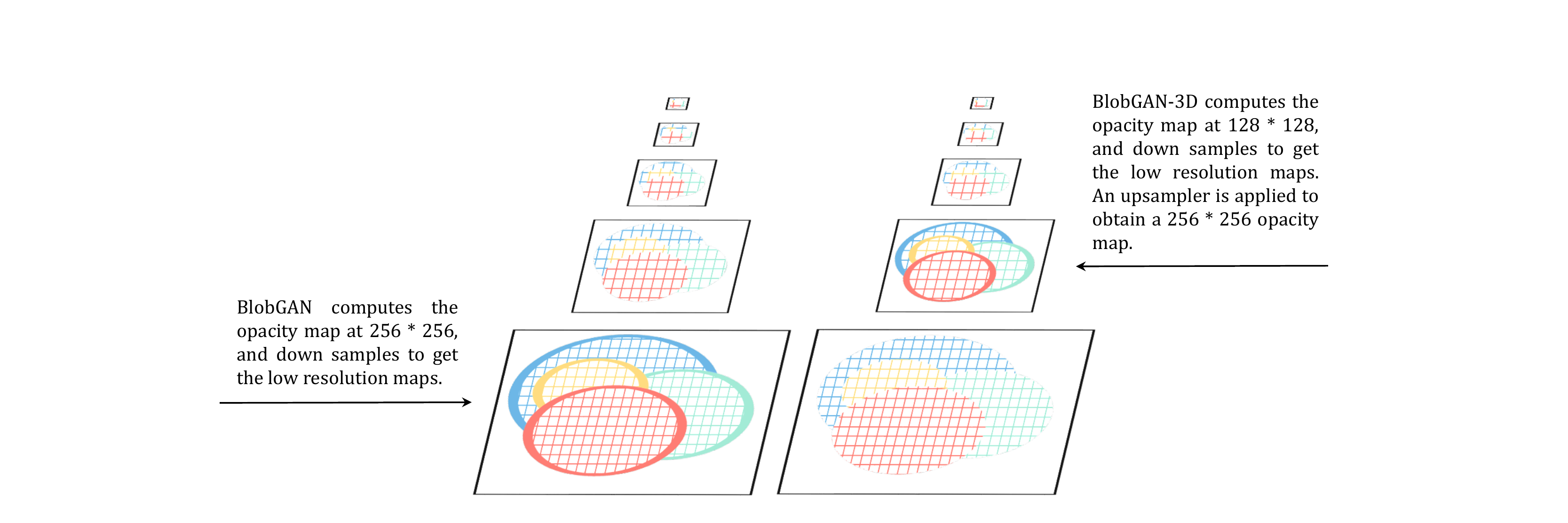}
   \caption{\textbf{The difference between BlobGAN and BlobGAN-3D on how the hierarchical opacity maps are computed (left) and the designed upsampler module (right).} In BlobGAN-3D, an additional upsampler module is applied to upsample the computed opacity map from $128 \times 128$ to $256 \times 256$. }
   \label{fig:condition}
\end{figure*}

We show the difference between BlobGAN and our BlobGAN-3D on how the hierarchical opacity maps are computed in \cref{fig:condition} (left). In \cref{fig:condition} (right), the architecture of the upsampler module is shown.  We find the balance between computational complexity and performance by computing the opacity map at a resolution of $128 \times 128$. 

\subsection{Additional training design}
During the third training stage, in each iteration we randomly drop one blob with a probability of 0.5 when rendering the images. By doing this we want the network to synthesize realistic images even when an object disappears in the scene, which in turn can improve the disentanglement. When determining the occlusions, we initially forced the network to learn the depth of all blobs in descending order. This however is not helpful because it constrains the learning ability of the MLP network. We later allow the network to learn an arbitrary normalized depth for each blob followed by a sorting operation.

\subsection{Training time and hyperparameters}
Compared to other models which directly model the scene as a whole, BlobGAN and BlobGAN-3D suffer from a low convergence rate because the image rendering takes place on the feature grid of the blob representation instead of the scene representation. Originally, BlobGAN trained their models for 1.5 million gradient steps on 8 NVIDIA A100 GPUs for 4 weeks. This is roughly equal to a $288000$K-image training run. Due to the limited resources, we trained our model for around $200000$K images on each dataset on 4 NVIDIA A100 GPUs, compared to $25000$K images for other baselines. We use a per-GPU batch size of $16$. 

For most of the settings, we follow BlobGAN. We use AdamW \cite{loshchilov2017adamw} optimizer with $\beta_0 = 0$ and $\beta_1 = 0.99$. We also use an exponentially moving average of the network weights \cite{yasin2018ema} with a decay of $0.998$ as it was used in BlobGAN. For the mapping network, we use an $8$-layer $1280$-channel MLP compared to an $8$-layer $1024$-channel MLP in BlobGAN.\par

\subsection{Volume rendering}
Compared to Nerf-like volume rendering, we do not represent the volume as implicit function and do not use an MLP to query the density value and feature value. We instead use a squared Mahalanobis distance to compute the density at each query point in 3D space. This is similar to BlobGAN where an opacity value is computed at each query point in 2D space. We reuse the feature vector of each blob as feature value. The Mahalanobis distance computes the distance from a point to a distribution. We model the density of each blob as a normal distribution and choose a suitable parameterization. The closer a query point is to the center of the blob, the higher is the density value. Therefore, the density is a continuous value and can be computed at any given point. As the density of each location is closely related to the distance to the center of the blob, when the blob is moving, the density at each location will change accordingly.

\section{Qualitative results}
\begin{figure*}[htbp]
  \centering   
  \includegraphics[width=\linewidth]{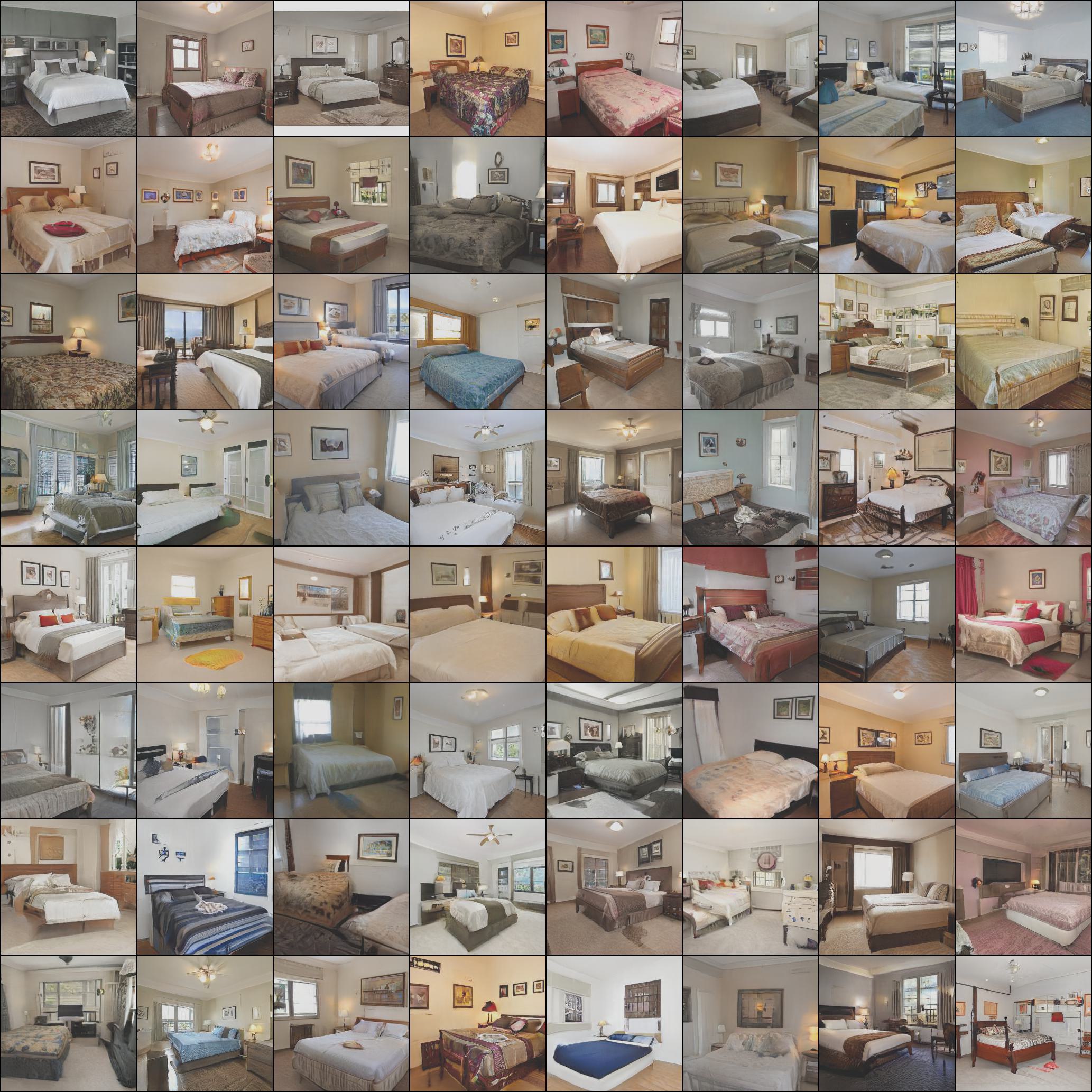}
   \caption{Generated images on bedroom dataset. \textbf{FID = 4.16}. Truncation weight $w = 0.3$.}
   \label{fig:fid_bedroom}
\end{figure*}

\begin{figure*}[htbp]
  \centering   
  \includegraphics[width=\linewidth]{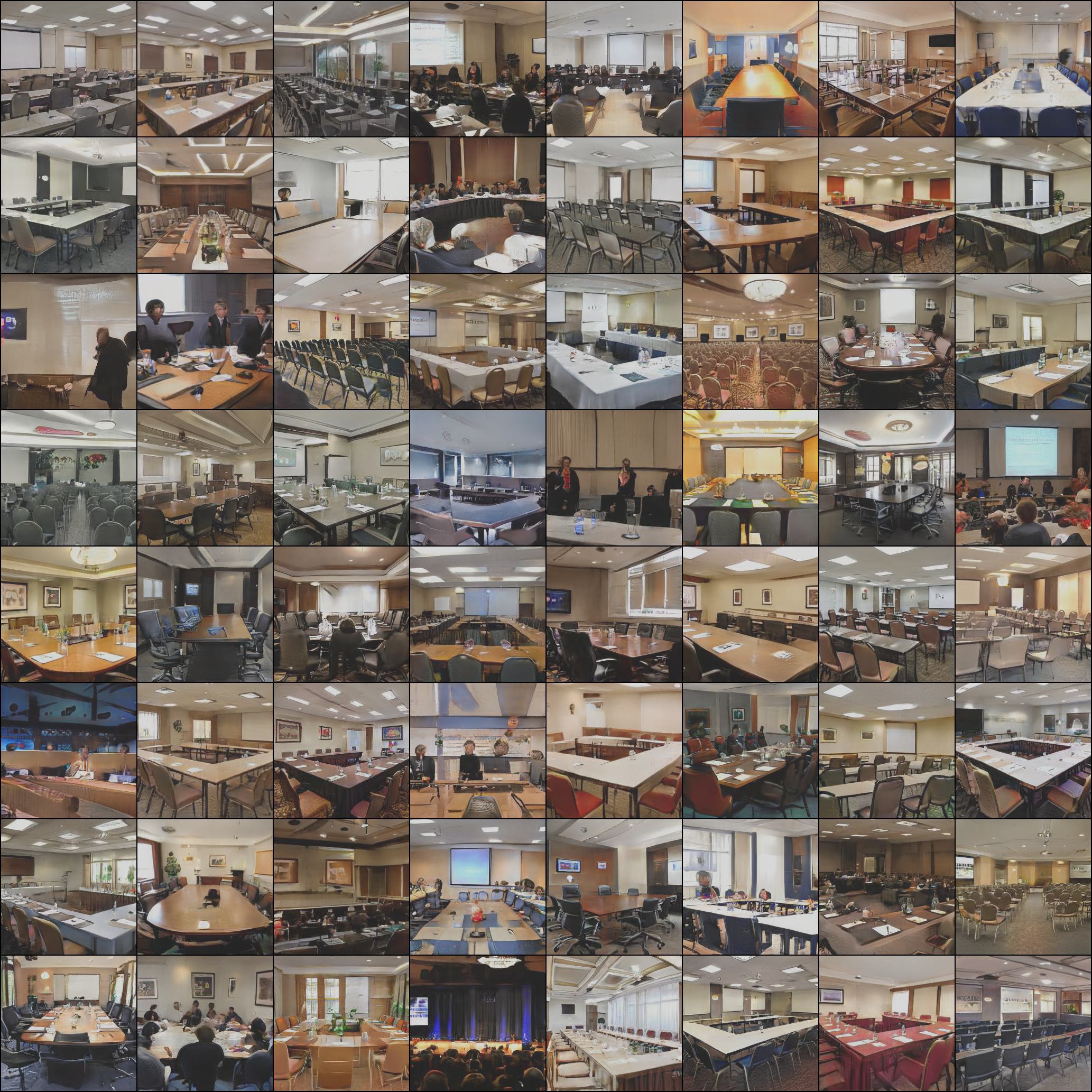}
   \caption{Generated images on conference room dataset. \textbf{FID = 7.80}. Truncation wegiht $w = 0.3$.}
   \label{fig:fid_conference}
\end{figure*}

We use the truncation technique similar to BlobGAN to improve the image quality. For all the figures in the paper, we use a truncation weight $w = 0.6$. The truncation weight $w$ ranges from $0$ to $1$. The higher the truncation weight, the higher the image quality but the lower the diversity. To show the diversity of our generated images, we provide samples in \cref{fig:fid_bedroom} and \cref{fig:fid_conference} using $w = 0.3$. \par
\begin{figure*}[t]
  \centering
  \begin{subfigure}[t]{0.11\linewidth}
    \includegraphics[width=\linewidth]{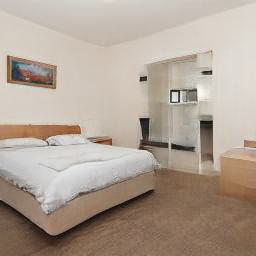}
\end{subfigure}
\begin{subfigure}[t]{0.11\linewidth}
    \includegraphics[width=\linewidth]{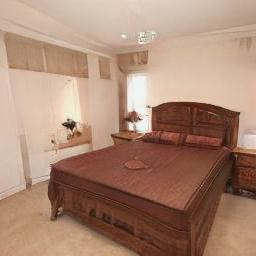}
\end{subfigure}
  \begin{subfigure}[t]{0.11\linewidth}
    \includegraphics[width=\linewidth]{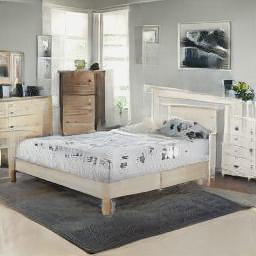}
\end{subfigure}
  \begin{subfigure}[t]{0.11\linewidth}
    \includegraphics[width=\linewidth,height=\linewidth]{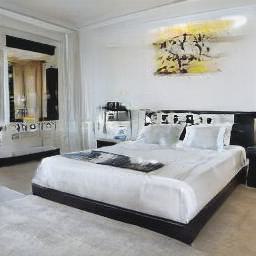}
\end{subfigure}
  \begin{subfigure}[t]{0.11\linewidth}
    \includegraphics[width=\linewidth,height=\linewidth]{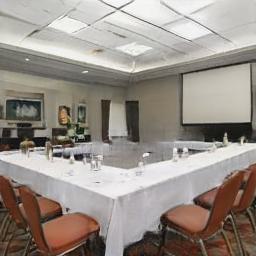}
\end{subfigure}
  \begin{subfigure}[t]{0.11\linewidth}
    \includegraphics[width=\linewidth,height=\linewidth]{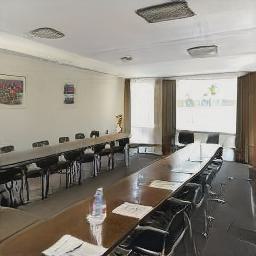}
\end{subfigure}
  \begin{subfigure}[t]{0.11\linewidth}
    \includegraphics[width=\linewidth]{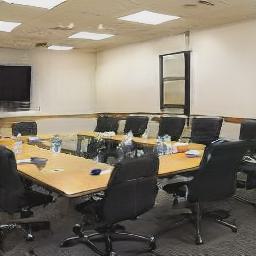}
\end{subfigure}
  \begin{subfigure}[t]{0.11\linewidth}
    \includegraphics[width=\linewidth]{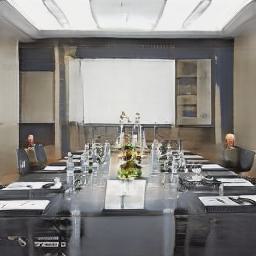}
\end{subfigure}
\quad
  \begin{subfigure}[t]{0.11\linewidth}
       \begin{overpic}[width=\textwidth]{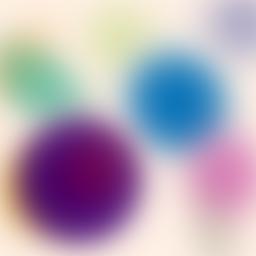}
\linethickness{1.5pt}
\put(32,30){\color[RGB]{255,255,255}\vector(1,0){40}}
\end{overpic}
\end{subfigure}
  \begin{subfigure}[t]{0.11\linewidth}
   \begin{overpic}[width=\textwidth]{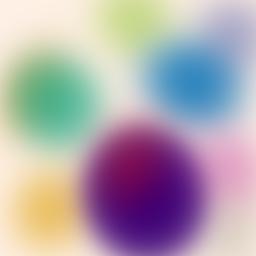}
\linethickness{1.5pt}
\put(57,27){\color[RGB]{255,255,255}\vector(1,0){40}}
\end{overpic}
\end{subfigure}
  \begin{subfigure}[t]{0.11\linewidth}
  \begin{overpic}[width=\textwidth]{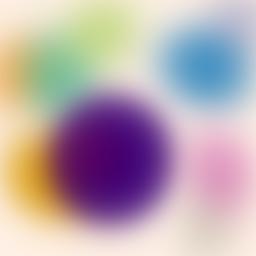}
\linethickness{1.5pt}
\put(80,75){\color[RGB]{255,255,255}\vector(1,0){25}}
\end{overpic}  
\end{subfigure}
  \begin{subfigure}[t]{0.11\linewidth}
   \begin{overpic}[width=\textwidth]{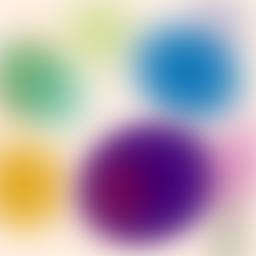}
\linethickness{1.5pt}
\put(74,72){\color[RGB]{255,255,255}\vector(1,0){30}}
\end{overpic}  
\end{subfigure}
  \begin{subfigure}[t]{0.11\linewidth}
  \begin{overpic}[width=\textwidth]{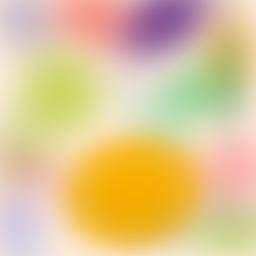}
\linethickness{1.5pt}
\put(51,24){\color[RGB]{255,255,255}\vector(1,0){30}}
\end{overpic}  
\end{subfigure}
  \begin{subfigure}[t]{0.11\linewidth}
  \begin{overpic}[width=\textwidth]{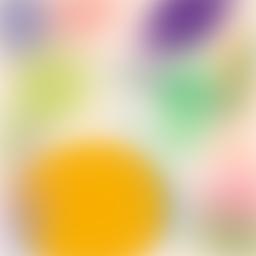}
\linethickness{1.5pt}
\put(38,21){\color[RGB]{255,255,255}\vector(1,0){30}}
\end{overpic}  
\end{subfigure}
  \begin{subfigure}[t]{0.11\linewidth}
  \begin{overpic}[width=\textwidth]{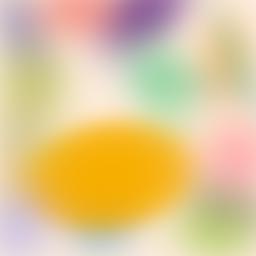}
\linethickness{1.5pt}
\put(65,68){\color[RGB]{255,255,255}\vector(1,0){30}}
\end{overpic}  
\end{subfigure}
  \begin{subfigure}[t]{0.11\linewidth}
  \begin{overpic}[width=\textwidth]{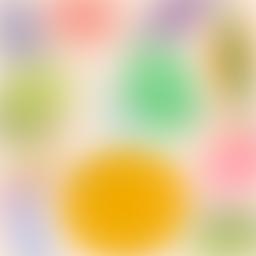}
\linethickness{1.5pt}
\put(60,65){\color[RGB]{255,255,255}\vector(1,0){30}}
\end{overpic}  
\end{subfigure}
\quad
  \begin{subfigure}[t]{0.11\linewidth}
    \includegraphics[width=\linewidth]{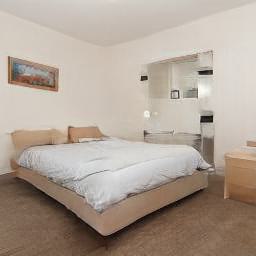}
\end{subfigure}
  \begin{subfigure}[t]{0.11\linewidth}
    \includegraphics[width=\linewidth]{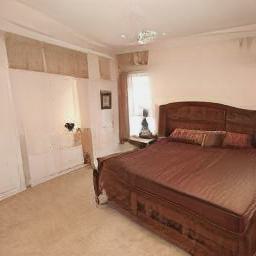}
\end{subfigure}
  \begin{subfigure}[t]{0.11\linewidth}
    \includegraphics[width=\linewidth]{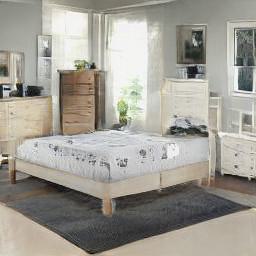}
\end{subfigure}
  \begin{subfigure}[t]{0.11\linewidth}    \includegraphics[width=\linewidth,height=\linewidth]{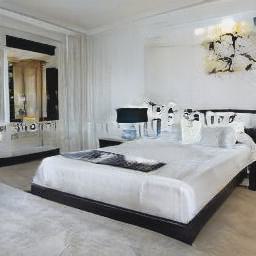}
\end{subfigure}
  \begin{subfigure}[t]{0.11\linewidth}
    \includegraphics[width=\linewidth,height=\linewidth]{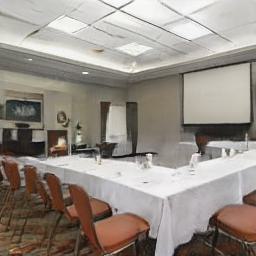}
\end{subfigure}
  \begin{subfigure}[t]{0.11\linewidth}
    \includegraphics[width=\linewidth,height=\linewidth]{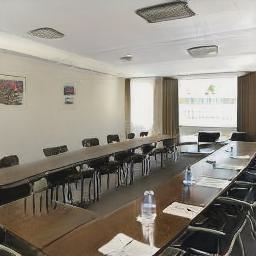}
\end{subfigure}
  \begin{subfigure}[t]{0.11\linewidth}
    \includegraphics[width=\linewidth]{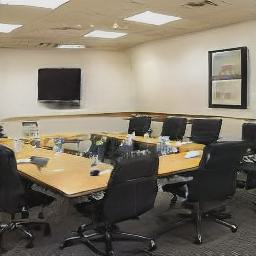}
\end{subfigure}
  \begin{subfigure}[t]{0.11\linewidth}
    \includegraphics[width=\linewidth]{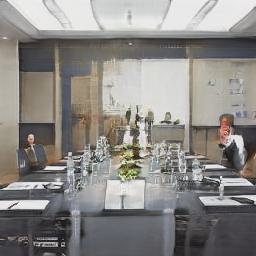}
\end{subfigure}
\caption{\textbf{Original BlobGAN: Moving an object in the scene.} The first row shows the original generated images; the second row shows the corresponding blob layout maps with marks demonstrating the moving direction. The last row shows the synthesized images after moving one blob. We show these as a comparison to BlobGAN-3D.}
\label{fig:moving_2d}
\end{figure*}
\begin{figure*}[t]
  \centering
  \begin{subfigure}[t]{0.11\linewidth}
    \includegraphics[width=\linewidth]{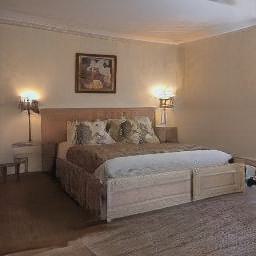}
\end{subfigure}
  \begin{subfigure}[t]{0.11\linewidth}
    \includegraphics[width=\linewidth]{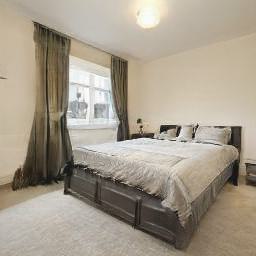}
\end{subfigure}
  \begin{subfigure}[t]{0.11\linewidth}
    \includegraphics[width=\linewidth]{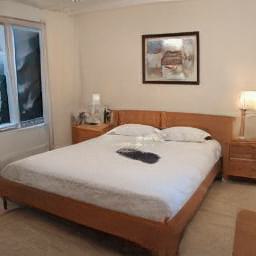}
\end{subfigure}
  \begin{subfigure}[t]{0.11\linewidth}
    \includegraphics[width=\linewidth,height=\linewidth]{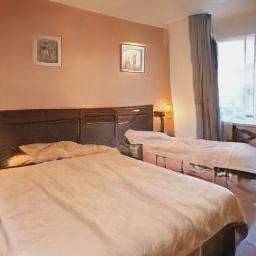}
\end{subfigure}
  \begin{subfigure}[t]{0.11\linewidth}
    \includegraphics[width=\linewidth,height=\linewidth]{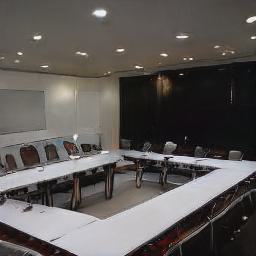}
\end{subfigure}
  \begin{subfigure}[t]{0.11\linewidth}
    \includegraphics[width=\linewidth,height=\linewidth]{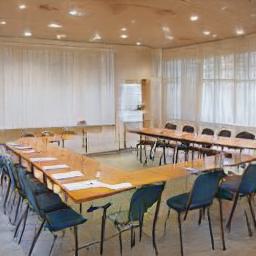}
\end{subfigure}
  \begin{subfigure}[t]{0.11\linewidth}
    \includegraphics[width=\linewidth]{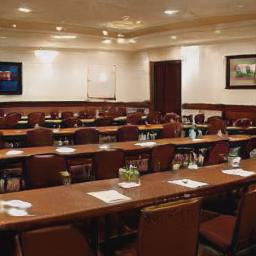}
\end{subfigure}
  \begin{subfigure}[t]{0.11\linewidth}
    \includegraphics[width=\linewidth]{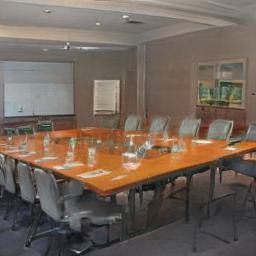}
\end{subfigure}
\quad
  \begin{subfigure}[t]{0.11\linewidth}
  \begin{overpic}[width=\textwidth]{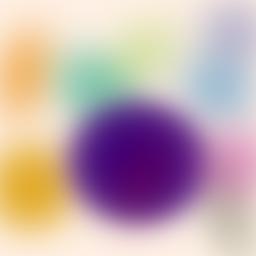}
\linethickness{1.5pt}
\put(45,45){\color[RGB]{255,255,255}\line(1,-1){20}}
\put(65,45){\color[RGB]{255,255,255}\line(-1,-1){20}}
\end{overpic} 
\end{subfigure}
  \begin{subfigure}[t]{0.11\linewidth}
  \begin{overpic}[width=\textwidth]{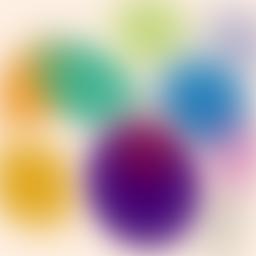}
\linethickness{1.5pt}
\put(48,38){\color[RGB]{255,255,255}\line(1,-1){20}}
\put(68,38){\color[RGB]{255,255,255}\line(-1,-1){20}}
\end{overpic} 
\end{subfigure}
  \begin{subfigure}[t]{0.11\linewidth}
  \begin{overpic}[width=\textwidth]{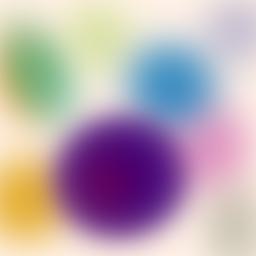}
\linethickness{1.5pt}
\put(60,74){\color[RGB]{255,255,255}\line(1,-1){15}}
\put(75,74){\color[RGB]{255,255,255}\line(-1,-1){15}}
\end{overpic} 
\end{subfigure}
  \begin{subfigure}[t]{0.11\linewidth}
  \begin{overpic}[width=\textwidth]{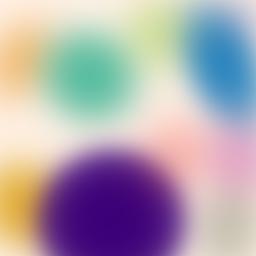}
\linethickness{1.5pt}
\put(80,79){\color[RGB]{255,255,255}\line(1,-1){15}}
\put(95,79){\color[RGB]{255,255,255}\line(-1,-1){15}}
\end{overpic} 
\end{subfigure}
  \begin{subfigure}[t]{0.11\linewidth}
  \begin{overpic}[width=\textwidth]{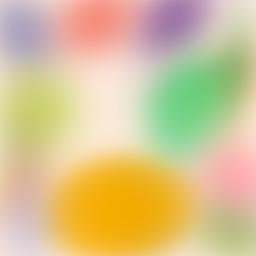}
\linethickness{1.5pt}
\put(35,35){\color[RGB]{255,255,255}\line(1,-1){30}}
\put(63,35){\color[RGB]{255,255,255}\line(-1,-1){30}}
\end{overpic} 
\end{subfigure}
  \begin{subfigure}[t]{0.11\linewidth}
  \begin{overpic}[width=\textwidth]{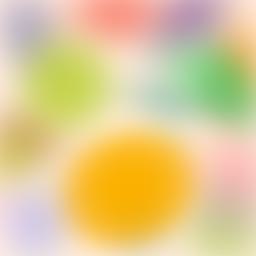}
\linethickness{1.5pt}
\put(38,38){\color[RGB]{255,255,255}\line(1,-1){28}}
\put(64,38){\color[RGB]{255,255,255}\line(-1,-1){28}}
\end{overpic} 
\end{subfigure}
  \begin{subfigure}[t]{0.11\linewidth}
  \begin{overpic}[width=\textwidth]{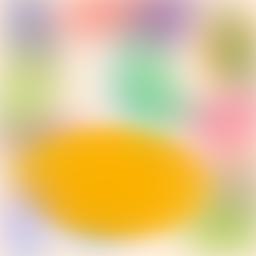}
\linethickness{1.5pt}
\put(48,78){\color[RGB]{255,255,255}\line(1,-1){20}}
\put(68,78){\color[RGB]{255,255,255}\line(-1,-1){20}}
\end{overpic} 
\end{subfigure}
  \begin{subfigure}[t]{0.11\linewidth}
  \begin{overpic}[width=\textwidth]{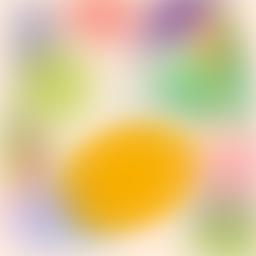}
\linethickness{1.5pt}
\put(68,76){\color[RGB]{255,255,255}\line(1,-1){20}}
\put(88,76){\color[RGB]{255,255,255}\line(-1,-1){20}}
\end{overpic} 
\end{subfigure}
\quad
  \begin{subfigure}[t]{0.11\linewidth}
    \includegraphics[width=\linewidth]{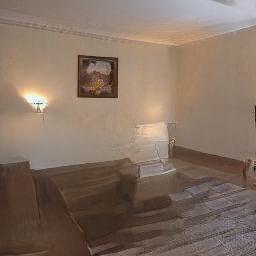}
\end{subfigure}
  \begin{subfigure}[t]{0.11\linewidth}
    \includegraphics[width=\linewidth]{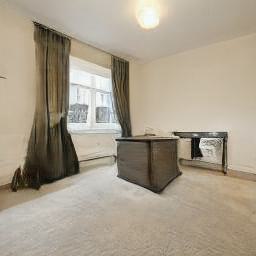}
\end{subfigure}
  \begin{subfigure}[t]{0.11\linewidth}
    \includegraphics[width=\linewidth]{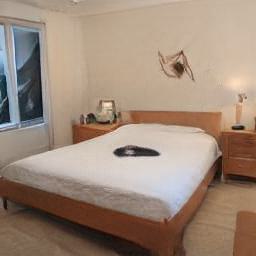}
\end{subfigure}
  \begin{subfigure}[t]{0.11\linewidth}    \includegraphics[width=\linewidth,height=\linewidth]{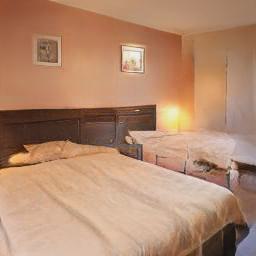}
\end{subfigure}
  \begin{subfigure}[t]{0.11\linewidth}
    \includegraphics[width=\linewidth,height=\linewidth]{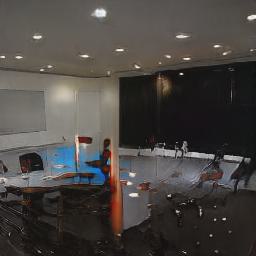}
\end{subfigure}
  \begin{subfigure}[t]{0.11\linewidth}
    \includegraphics[width=\linewidth,height=\linewidth]{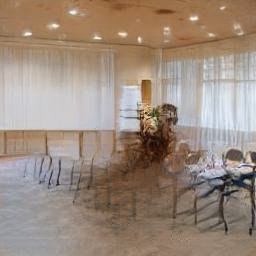}
\end{subfigure}
  \begin{subfigure}[t]{0.11\linewidth}
    \includegraphics[width=\linewidth]{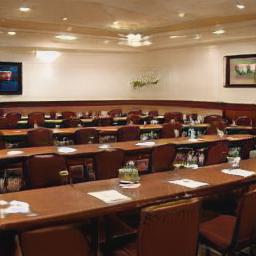}
\end{subfigure}
  \begin{subfigure}[t]{0.11\linewidth}
    \includegraphics[width=\linewidth]{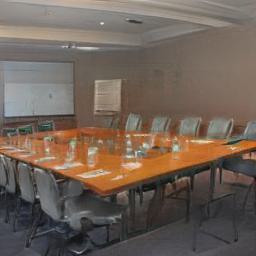}
\end{subfigure}
\caption{\textbf{Original BlobGAN: Removing an object in the scene.} The first row shows the originally generated images; the second row shows the corresponding blob layout maps with marks demonstrating the removed blob. The last row shows the synthesized images after removing one blob. We show these as a comparison to BlobGAN-3D.}
\label{fig:resizing_2d}
\end{figure*}
\begin{figure*}[t]
  \centering
  \begin{subfigure}[t]{0.16\linewidth}
    \includegraphics[width=\linewidth]{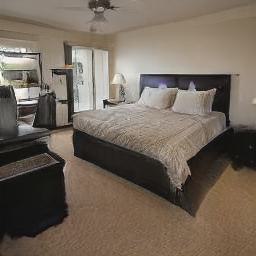}
\end{subfigure}
  \begin{subfigure}[t]{0.16\linewidth}
    \includegraphics[width=\linewidth]{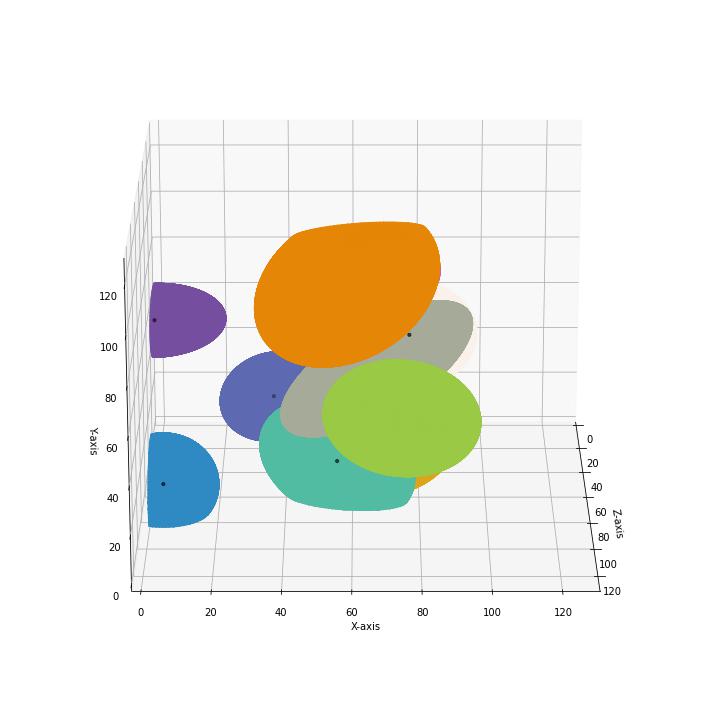}
\end{subfigure}
  \begin{subfigure}[t]{0.16\linewidth}
    \includegraphics[width=\linewidth]{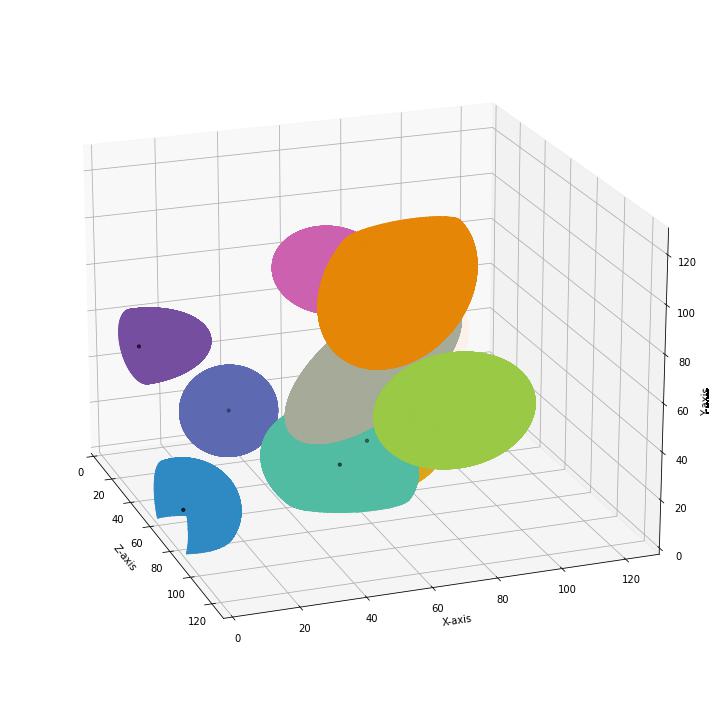}
\end{subfigure}
  \begin{subfigure}[t]{0.16\linewidth}
    \includegraphics[width=\linewidth]{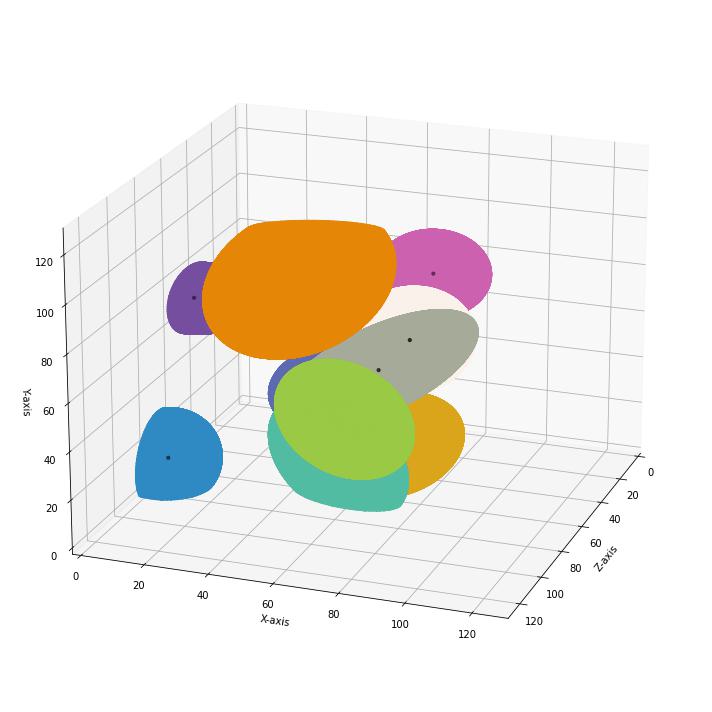}
\end{subfigure}
  \begin{subfigure}[t]{0.16\linewidth}
    \includegraphics[width=\linewidth]{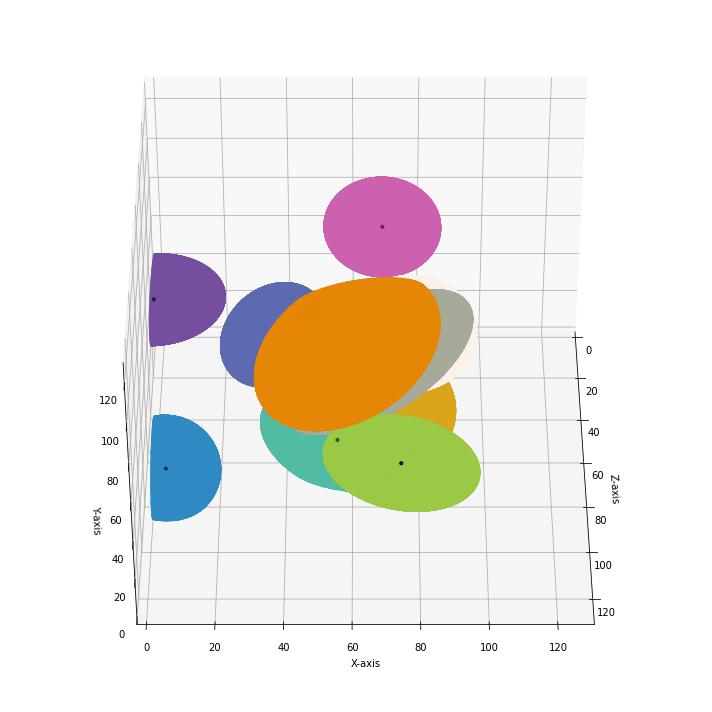}
\end{subfigure}
  \begin{subfigure}[t]{0.16\linewidth}
    \includegraphics[width=\linewidth]{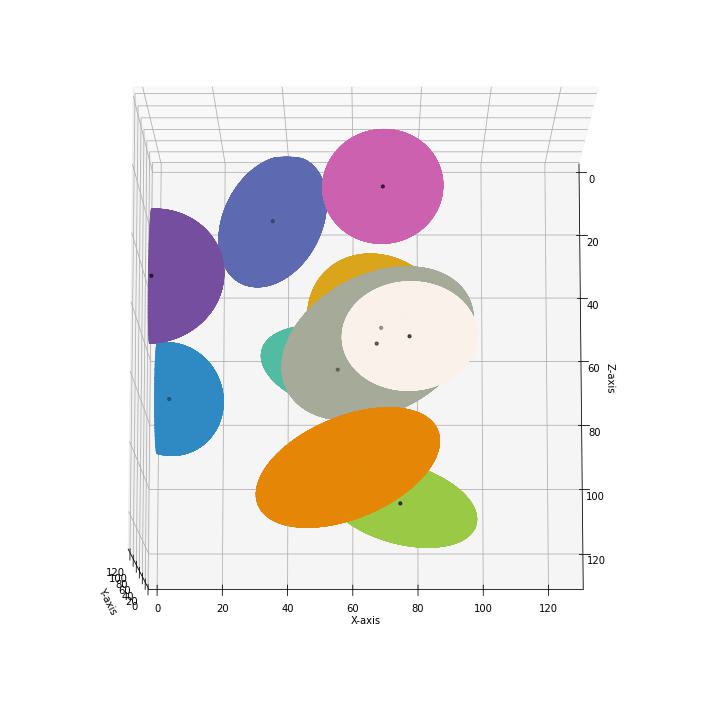}
\end{subfigure}
\quad
  \begin{subfigure}[t]{0.16\linewidth}
    \includegraphics[width=\linewidth]{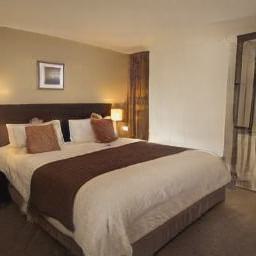}
\end{subfigure}
  \begin{subfigure}[t]{0.16\linewidth}
    \includegraphics[width=\linewidth]{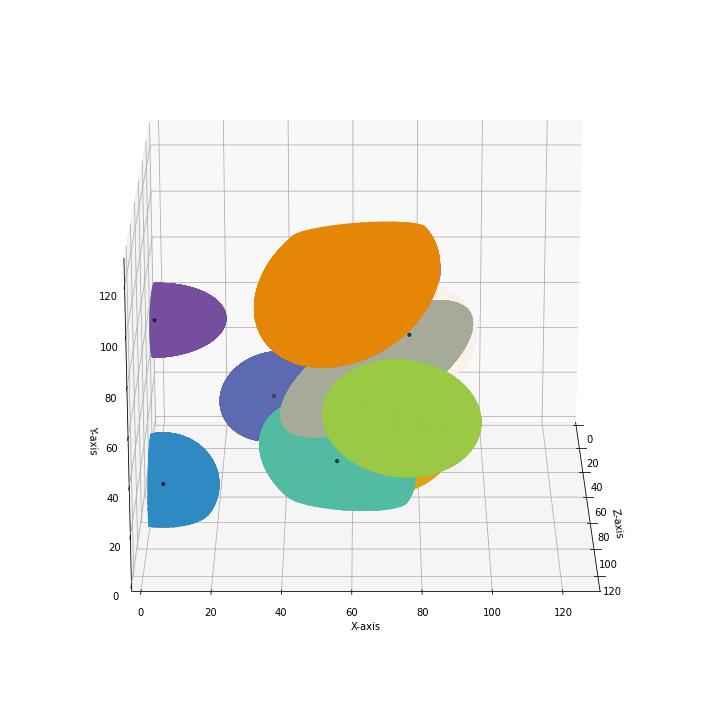}
\end{subfigure}
  \begin{subfigure}[t]{0.16\linewidth}
    \includegraphics[width=\linewidth]{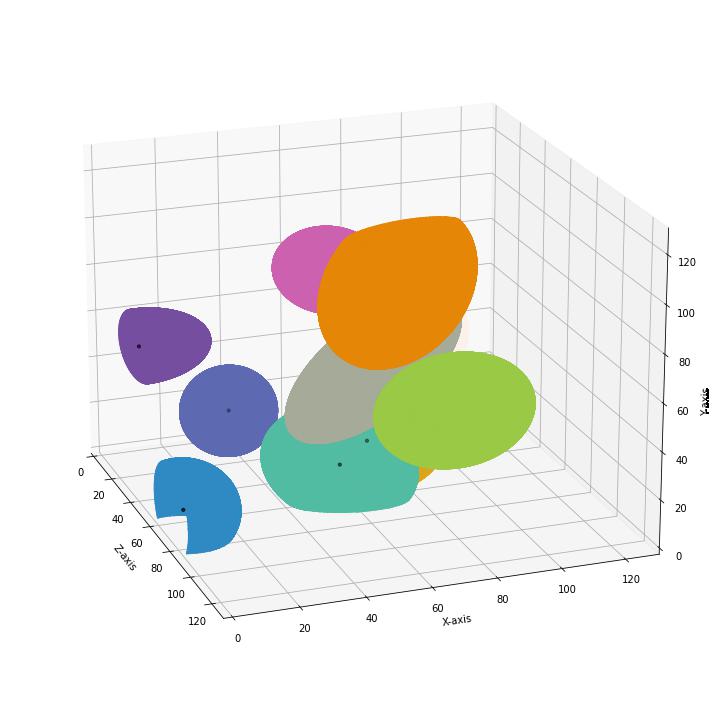}
\end{subfigure}
  \begin{subfigure}[t]{0.16\linewidth}
    \includegraphics[width=\linewidth]{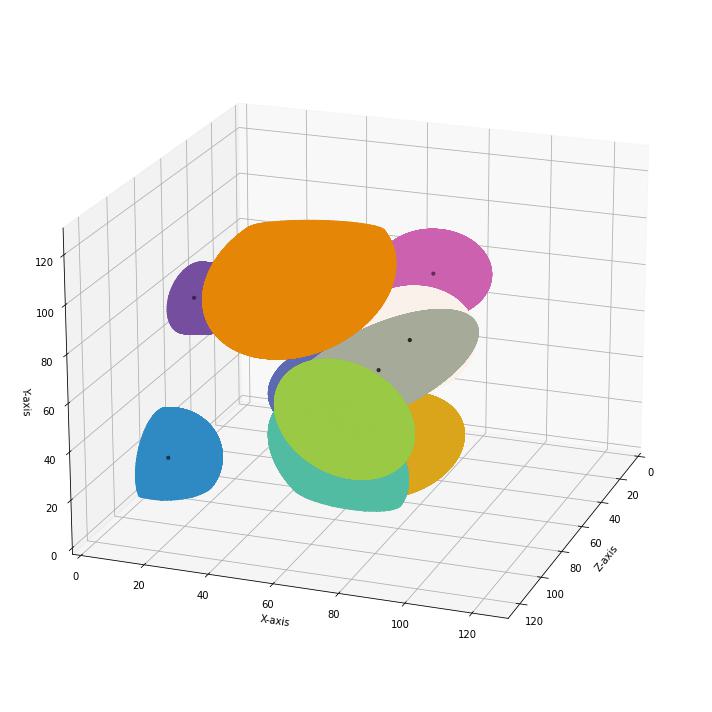}
\end{subfigure}
  \begin{subfigure}[t]{0.16\linewidth}
    \includegraphics[width=\linewidth]{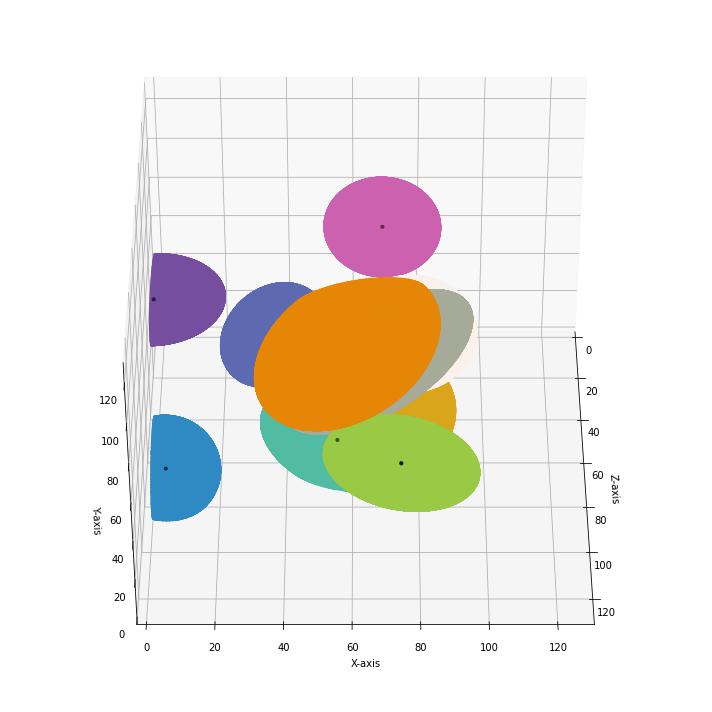}
\end{subfigure}
  \begin{subfigure}[t]{0.16\linewidth}
    \includegraphics[width=\linewidth]{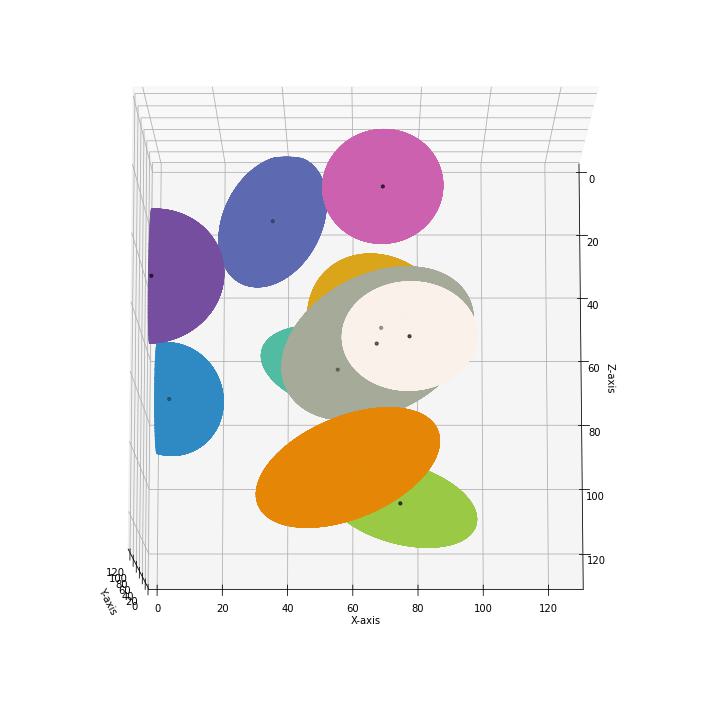}
\end{subfigure}
\caption{Blobs visualization in 3D space. The first column shows the synthetic image, the second to sixth column shows the corresponding blobs in 3D space which are visualized under different camera viewpoints.}
\label{fig:3d_blobs}
\end{figure*}
\begin{figure*}[htbp]
  \centering
  \begin{subfigure}[t]{0.15\linewidth}
    \includegraphics[width=\linewidth]{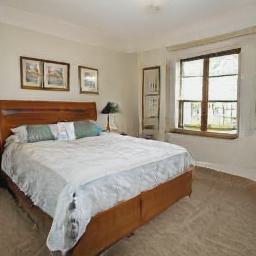}
\end{subfigure}
  \begin{subfigure}[t]{0.15\linewidth}
    \includegraphics[width=\linewidth]{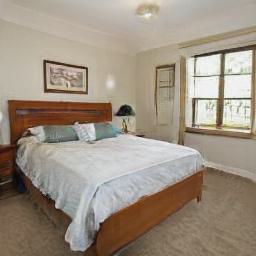}
\end{subfigure}
  \begin{subfigure}[t]{0.15\linewidth}
    \includegraphics[width=\linewidth]{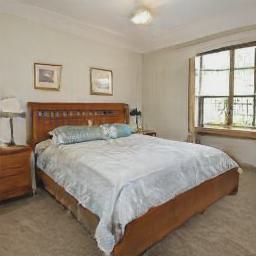}
\end{subfigure}\hspace{1mm}
  \begin{subfigure}[t]{0.15\linewidth}    \includegraphics[width=\linewidth,height=\linewidth]{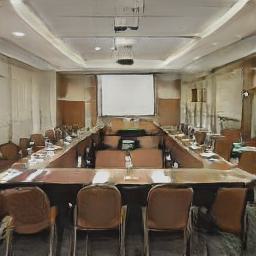}
\end{subfigure}
  \begin{subfigure}[t]{0.15\linewidth}    \includegraphics[width=\linewidth,height=\linewidth]{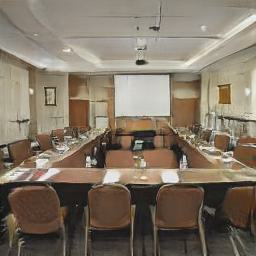}
\end{subfigure}
  \begin{subfigure}[t]{0.15\linewidth}
    \includegraphics[width=\linewidth]{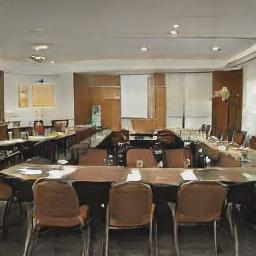}
\end{subfigure}
\quad
  \begin{subfigure}[t]{0.15\linewidth}
    \includegraphics[width=\linewidth]{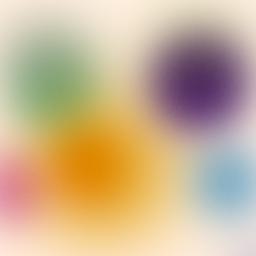}
\end{subfigure}
  \begin{subfigure}[t]{0.15\linewidth}
    \includegraphics[width=\linewidth]{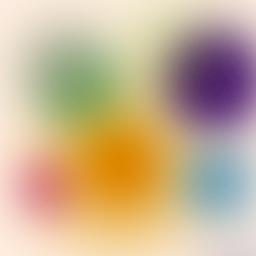}
\end{subfigure}
  \begin{subfigure}[t]{0.15\linewidth}
    \includegraphics[width=\linewidth]{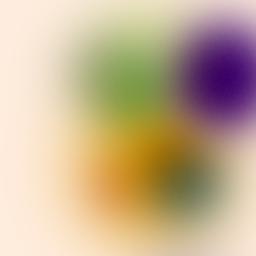}
\end{subfigure}\hspace{1mm}
  \begin{subfigure}[t]{0.15\linewidth}    \includegraphics[width=\linewidth,height=\linewidth]{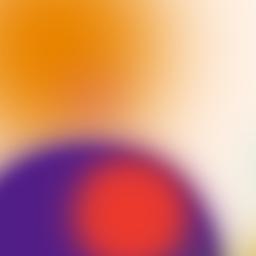}
\end{subfigure}
  \begin{subfigure}[t]{0.15\linewidth}    \includegraphics[width=\linewidth,height=\linewidth]{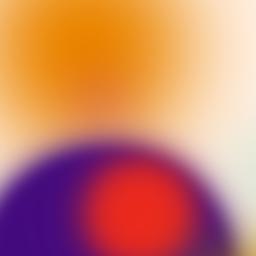}
\end{subfigure}
  \begin{subfigure}[t]{0.15\linewidth}
    \includegraphics[width=\linewidth]{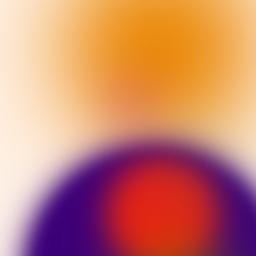}
\end{subfigure}
\quad
  \begin{subfigure}[t]{0.15\linewidth}
    \includegraphics[width=\linewidth]{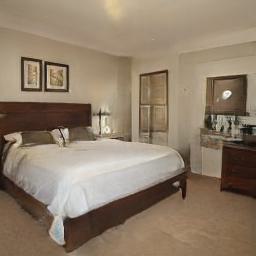}
\end{subfigure}
  \begin{subfigure}[t]{0.15\linewidth}
    \includegraphics[width=\linewidth]{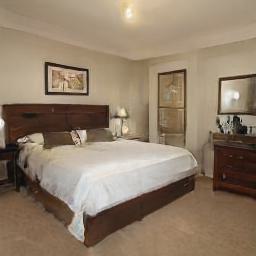}
\end{subfigure}
  \begin{subfigure}[t]{0.15\linewidth}
    \includegraphics[width=\linewidth]{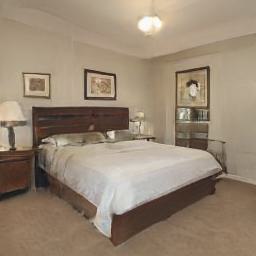}
\end{subfigure}\hspace{1mm}
  \begin{subfigure}[t]{0.15\linewidth}    \includegraphics[width=\linewidth,height=\linewidth]{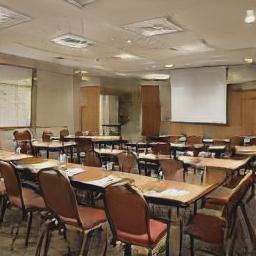}
\end{subfigure}
  \begin{subfigure}[t]{0.15\linewidth}    \includegraphics[width=\linewidth,height=\linewidth]{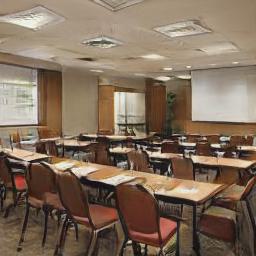}
\end{subfigure}
  \begin{subfigure}[t]{0.15\linewidth}
    \includegraphics[width=\linewidth]{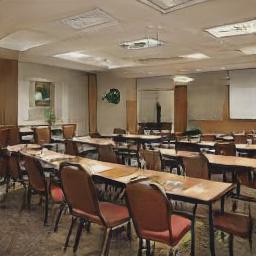}
\end{subfigure}
\quad
  \begin{subfigure}[t]{0.15\linewidth}
    \includegraphics[width=\linewidth]{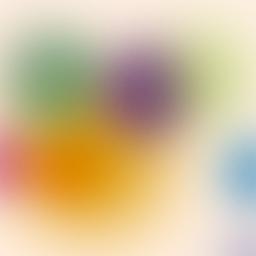}
\end{subfigure}
  \begin{subfigure}[t]{0.15\linewidth}
    \includegraphics[width=\linewidth]{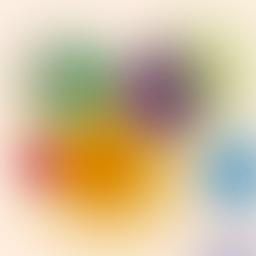}
\end{subfigure}
  \begin{subfigure}[t]{0.15\linewidth}
    \includegraphics[width=\linewidth]{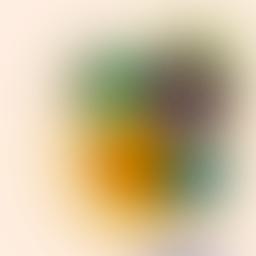}
\end{subfigure}\hspace{1mm}
  \begin{subfigure}[t]{0.15\linewidth}    \includegraphics[width=\linewidth,height=\linewidth]{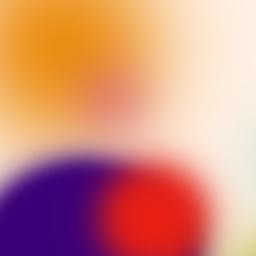}
\end{subfigure}
  \begin{subfigure}[t]{0.15\linewidth}    \includegraphics[width=\linewidth,height=\linewidth]{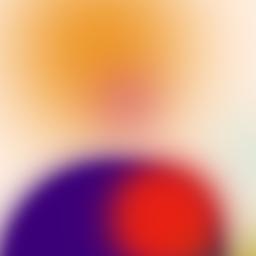}
\end{subfigure}
  \begin{subfigure}[t]{0.15\linewidth}
    \includegraphics[width=\linewidth]{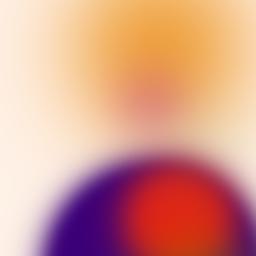}
\end{subfigure}
\quad
  \begin{subfigure}[t]{0.15\linewidth}
    \includegraphics[width=\linewidth]{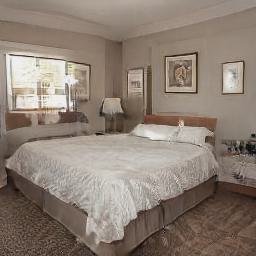}
\end{subfigure}
  \begin{subfigure}[t]{0.15\linewidth}
    \includegraphics[width=\linewidth]{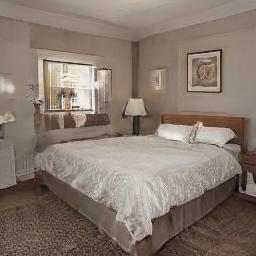}
\end{subfigure}
  \begin{subfigure}[t]{0.15\linewidth}
    \includegraphics[width=\linewidth]{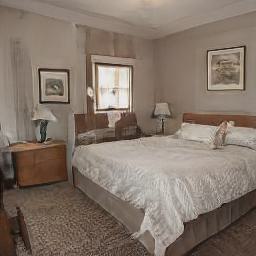}
\end{subfigure}\hspace{1mm}
  \begin{subfigure}[t]{0.15\linewidth}    \includegraphics[width=\linewidth,height=\linewidth]{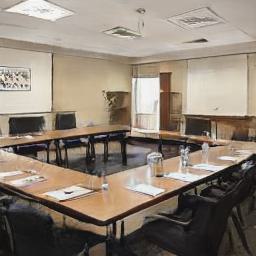}
\end{subfigure}
  \begin{subfigure}[t]{0.15\linewidth}    \includegraphics[width=\linewidth,height=\linewidth]{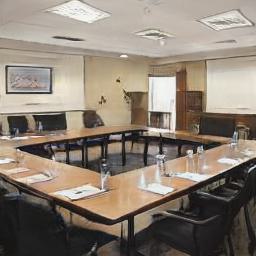}
\end{subfigure}
  \begin{subfigure}[t]{0.15\linewidth}
    \includegraphics[width=\linewidth]{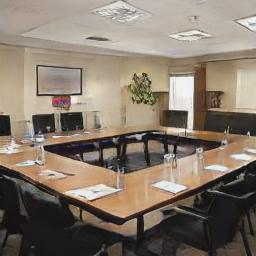}
\end{subfigure}
\quad
  \begin{subfigure}[t]{0.15\linewidth}
    \includegraphics[width=\linewidth]{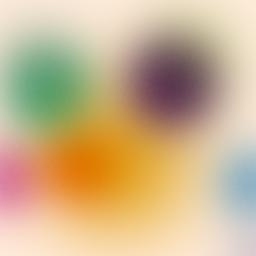}
\end{subfigure}
  \begin{subfigure}[t]{0.15\linewidth}
    \includegraphics[width=\linewidth]{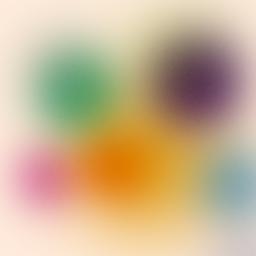}
\end{subfigure}
  \begin{subfigure}[t]{0.15\linewidth}
    \includegraphics[width=\linewidth]{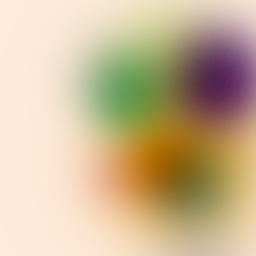}
\end{subfigure}\hspace{1mm}
  \begin{subfigure}[t]{0.15\linewidth}    \includegraphics[width=\linewidth,height=\linewidth]{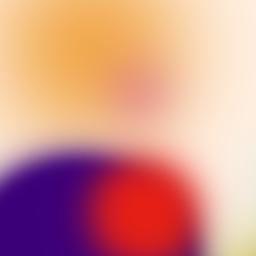}
\end{subfigure}
  \begin{subfigure}[t]{0.15\linewidth}    \includegraphics[width=\linewidth,height=\linewidth]{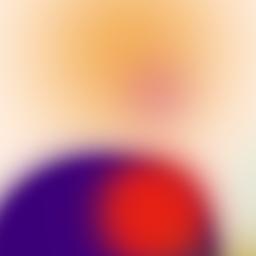}
\end{subfigure}
  \begin{subfigure}[t]{0.15\linewidth}
    \includegraphics[width=\linewidth]{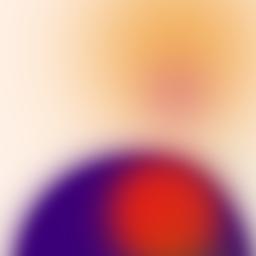}
\end{subfigure}
\quad
  \begin{subfigure}[t]{0.15\linewidth}
    \includegraphics[width=\linewidth]{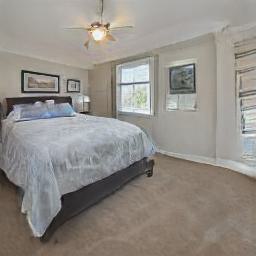}
\end{subfigure}
  \begin{subfigure}[t]{0.15\linewidth}
    \includegraphics[width=\linewidth]{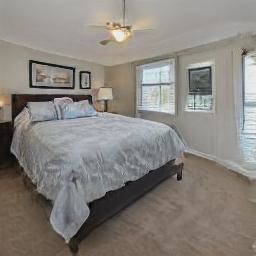}
\end{subfigure}
  \begin{subfigure}[t]{0.15\linewidth}
    \includegraphics[width=\linewidth]{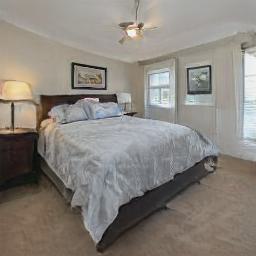}
\end{subfigure}\hspace{1mm}
  \begin{subfigure}[t]{0.15\linewidth}    \includegraphics[width=\linewidth,height=\linewidth]{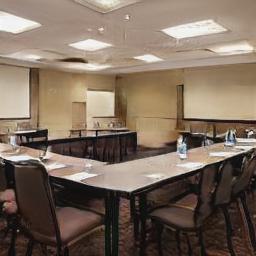}
\end{subfigure}
  \begin{subfigure}[t]{0.15\linewidth}    \includegraphics[width=\linewidth,height=\linewidth]{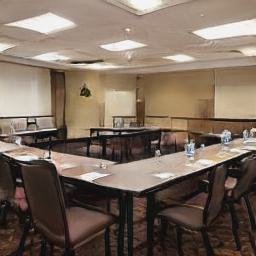}
\end{subfigure}
  \begin{subfigure}[t]{0.15\linewidth}
    \includegraphics[width=\linewidth]{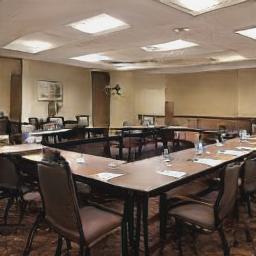}
\end{subfigure}
\quad
  \begin{subfigure}[t]{0.15\linewidth}
    \includegraphics[width=\linewidth]{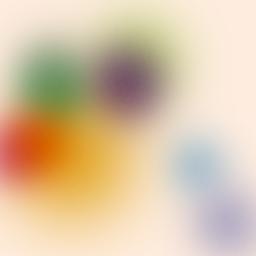}
\end{subfigure}
  \begin{subfigure}[t]{0.15\linewidth}
    \includegraphics[width=\linewidth]{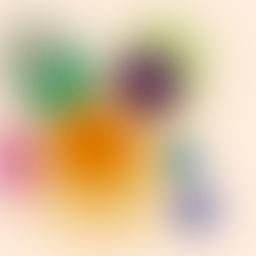}
\end{subfigure}
  \begin{subfigure}[t]{0.15\linewidth}
    \includegraphics[width=\linewidth]{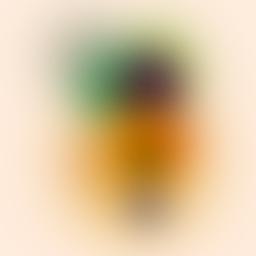}
\end{subfigure}\hspace{1mm}
  \begin{subfigure}[t]{0.15\linewidth}    \includegraphics[width=\linewidth,height=\linewidth]{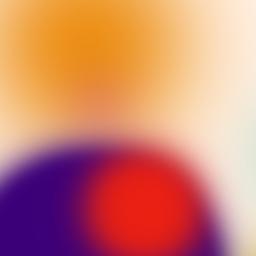}
\end{subfigure}
  \begin{subfigure}[t]{0.15\linewidth}    \includegraphics[width=\linewidth,height=\linewidth]{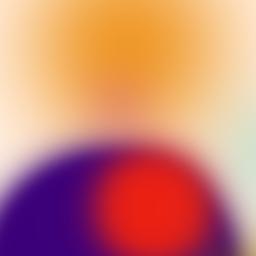}
\end{subfigure}
  \begin{subfigure}[t]{0.15\linewidth}
    \includegraphics[width=\linewidth]{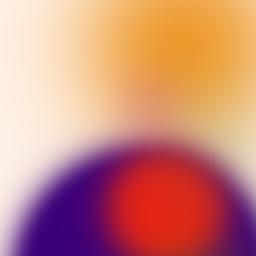}
\end{subfigure}
\caption{\textbf{Moving the camera in the horizontal direction.} For every two rows, the first row shows the generated images, while the second row shows the corresponding blobs layout map.}
\label{fig:moving_yaw}
\vspace{-5mm}
\end{figure*}
\begin{figure*}[htbp]
  \centering
  \begin{subfigure}[t]{0.15\linewidth}
    \includegraphics[width=\linewidth]{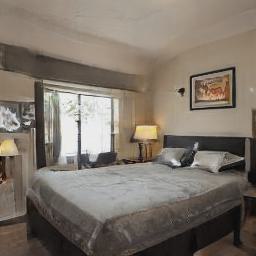}
\end{subfigure}
  \begin{subfigure}[t]{0.15\linewidth}
    \includegraphics[width=\linewidth]{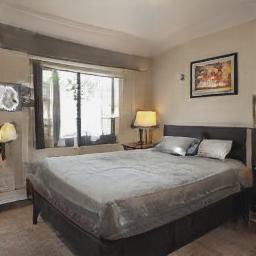}
\end{subfigure}
  \begin{subfigure}[t]{0.15\linewidth}
    \includegraphics[width=\linewidth]{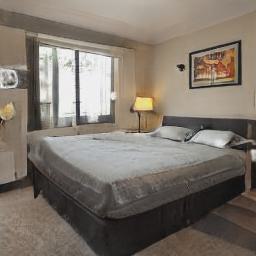}
\end{subfigure}\hspace{1mm}
  \begin{subfigure}[t]{0.15\linewidth}    \includegraphics[width=\linewidth,height=\linewidth]{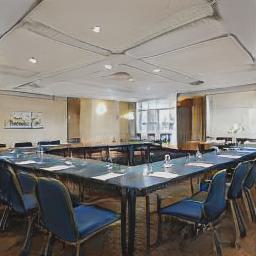}
\end{subfigure}
  \begin{subfigure}[t]{0.15\linewidth}    \includegraphics[width=\linewidth,height=\linewidth]{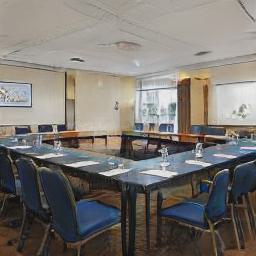}
\end{subfigure}
  \begin{subfigure}[t]{0.15\linewidth}
    \includegraphics[width=\linewidth]{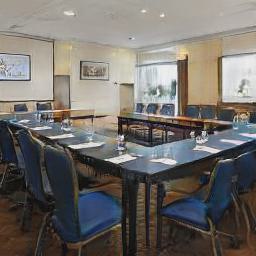}
\end{subfigure}
\quad
  \begin{subfigure}[t]{0.15\linewidth}
    \includegraphics[width=\linewidth]{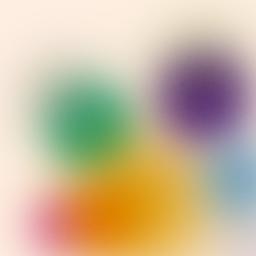}
\end{subfigure}
  \begin{subfigure}[t]{0.15\linewidth}
    \includegraphics[width=\linewidth]{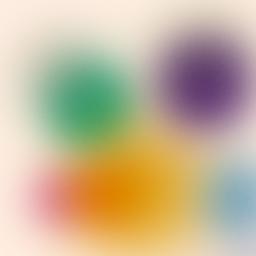}
\end{subfigure}
  \begin{subfigure}[t]{0.15\linewidth}
    \includegraphics[width=\linewidth]{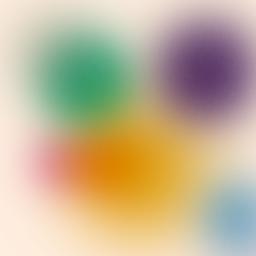}
\end{subfigure}\hspace{1mm}
  \begin{subfigure}[t]{0.15\linewidth}    \includegraphics[width=\linewidth,height=\linewidth]{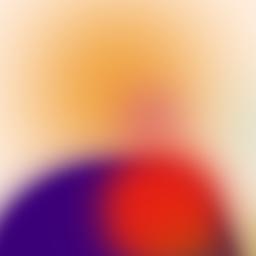}
\end{subfigure}
  \begin{subfigure}[t]{0.15\linewidth}    \includegraphics[width=\linewidth,height=\linewidth]{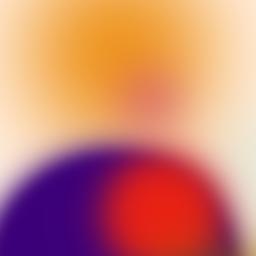}
\end{subfigure}
  \begin{subfigure}[t]{0.15\linewidth}
    \includegraphics[width=\linewidth]{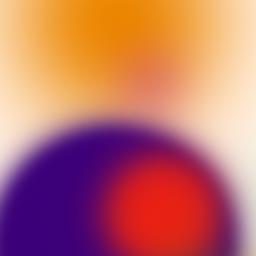}
\end{subfigure}
\quad
  \begin{subfigure}[t]{0.15\linewidth}
    \includegraphics[width=\linewidth]{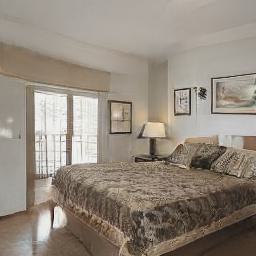}
\end{subfigure}
  \begin{subfigure}[t]{0.15\linewidth}
    \includegraphics[width=\linewidth]{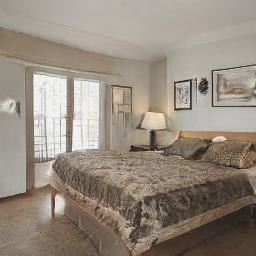}
\end{subfigure}
  \begin{subfigure}[t]{0.15\linewidth}
    \includegraphics[width=\linewidth]{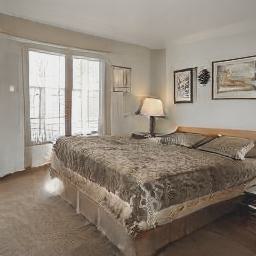}
\end{subfigure}\hspace{1mm}
  \begin{subfigure}[t]{0.15\linewidth}    \includegraphics[width=\linewidth,height=\linewidth]{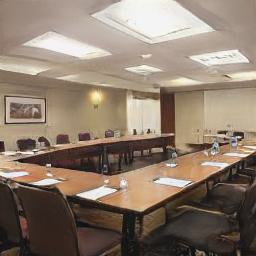}
\end{subfigure}
  \begin{subfigure}[t]{0.15\linewidth}    \includegraphics[width=\linewidth,height=\linewidth]{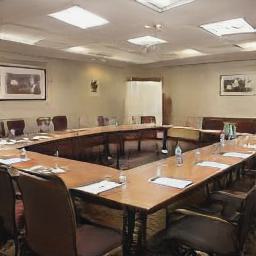}
\end{subfigure}
  \begin{subfigure}[t]{0.15\linewidth}
    \includegraphics[width=\linewidth]{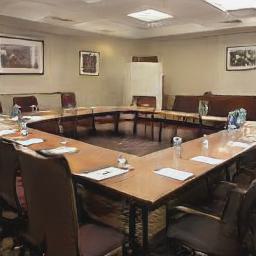}
\end{subfigure}
\quad
  \begin{subfigure}[t]{0.15\linewidth}
    \includegraphics[width=\linewidth]{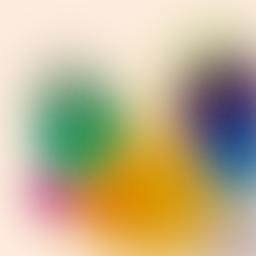}
\end{subfigure}
  \begin{subfigure}[t]{0.15\linewidth}
    \includegraphics[width=\linewidth]{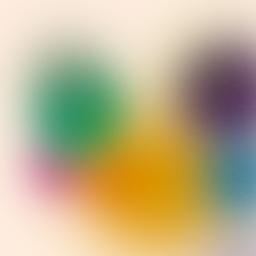}
\end{subfigure}
  \begin{subfigure}[t]{0.15\linewidth}
    \includegraphics[width=\linewidth]{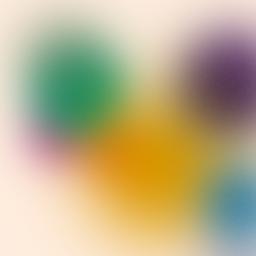}
\end{subfigure}\hspace{1mm}
  \begin{subfigure}[t]{0.15\linewidth}    \includegraphics[width=\linewidth,height=\linewidth]{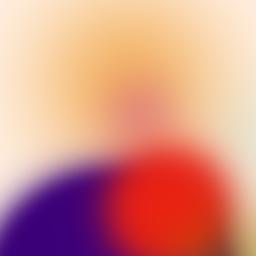}
\end{subfigure}
  \begin{subfigure}[t]{0.15\linewidth}    \includegraphics[width=\linewidth,height=\linewidth]{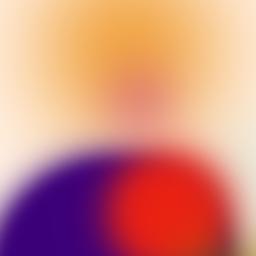}
\end{subfigure}
  \begin{subfigure}[t]{0.15\linewidth}
    \includegraphics[width=\linewidth]{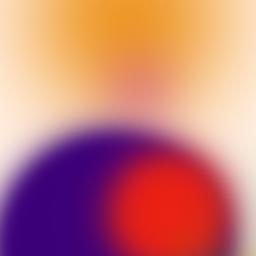}
\end{subfigure}
\quad
  \begin{subfigure}[t]{0.15\linewidth}
    \includegraphics[width=\linewidth]{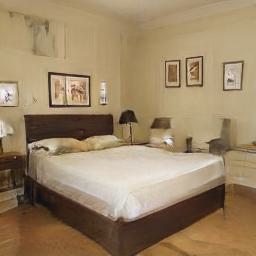}
\end{subfigure}
  \begin{subfigure}[t]{0.15\linewidth}
    \includegraphics[width=\linewidth]{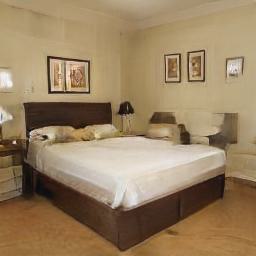}
\end{subfigure}
  \begin{subfigure}[t]{0.15\linewidth}
    \includegraphics[width=\linewidth]{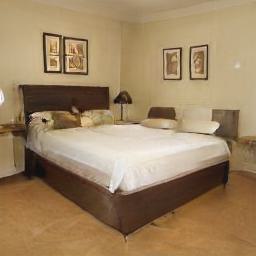}
\end{subfigure}\hspace{1mm}
  \begin{subfigure}[t]{0.15\linewidth}    \includegraphics[width=\linewidth,height=\linewidth]{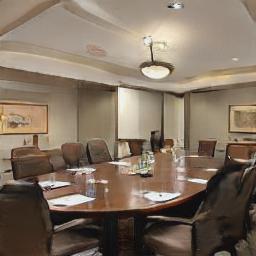}
\end{subfigure}
  \begin{subfigure}[t]{0.15\linewidth}    \includegraphics[width=\linewidth,height=\linewidth]{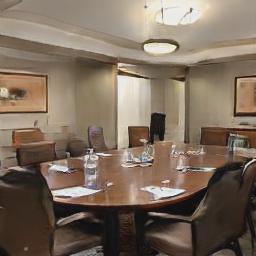}
\end{subfigure}
  \begin{subfigure}[t]{0.15\linewidth}
    \includegraphics[width=\linewidth]{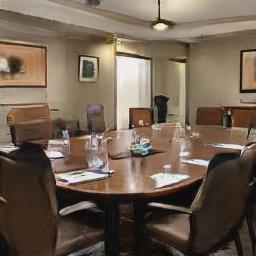}
\end{subfigure}
\quad
  \begin{subfigure}[t]{0.15\linewidth}
    \includegraphics[width=\linewidth]{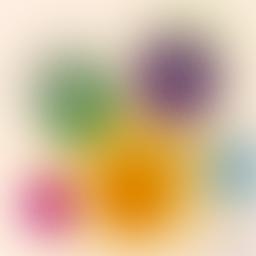}
\end{subfigure}
  \begin{subfigure}[t]{0.15\linewidth}
    \includegraphics[width=\linewidth]{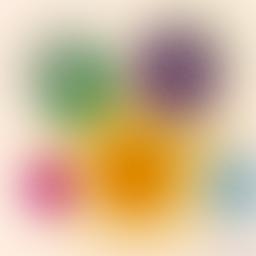}
\end{subfigure}
  \begin{subfigure}[t]{0.15\linewidth}
    \includegraphics[width=\linewidth]{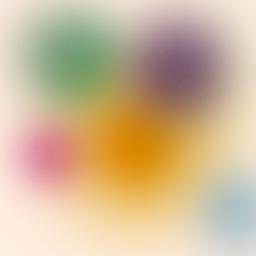}
\end{subfigure}\hspace{1mm}
  \begin{subfigure}[t]{0.15\linewidth}    \includegraphics[width=\linewidth,height=\linewidth]{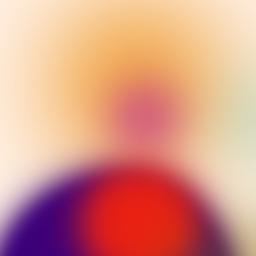}
\end{subfigure}
  \begin{subfigure}[t]{0.15\linewidth}    \includegraphics[width=\linewidth,height=\linewidth]{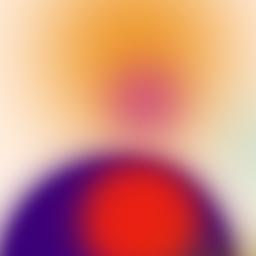}
\end{subfigure}
  \begin{subfigure}[t]{0.15\linewidth}
    \includegraphics[width=\linewidth]{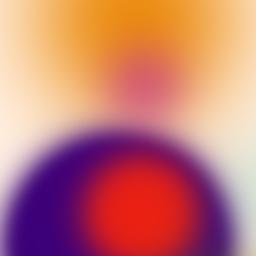}
\end{subfigure}
\quad
  \begin{subfigure}[t]{0.15\linewidth}
    \includegraphics[width=\linewidth]{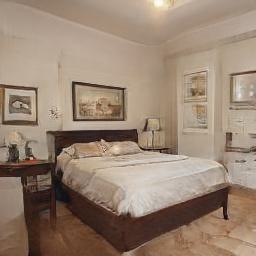}
\end{subfigure}
  \begin{subfigure}[t]{0.15\linewidth}
    \includegraphics[width=\linewidth]{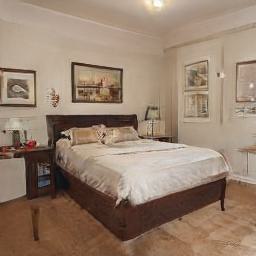}
\end{subfigure}
  \begin{subfigure}[t]{0.15\linewidth}
    \includegraphics[width=\linewidth]{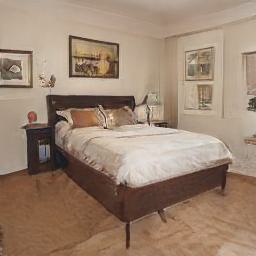}
\end{subfigure}\hspace{1mm}
  \begin{subfigure}[t]{0.15\linewidth}    \includegraphics[width=\linewidth,height=\linewidth]{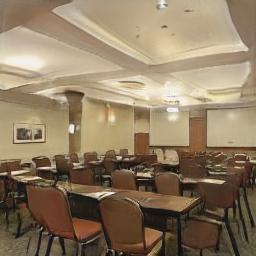}
\end{subfigure}
  \begin{subfigure}[t]{0.15\linewidth}    \includegraphics[width=\linewidth,height=\linewidth]{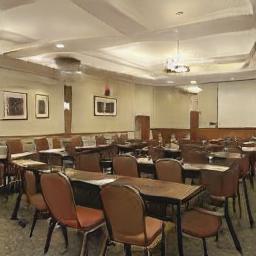}
\end{subfigure}
  \begin{subfigure}[t]{0.15\linewidth}
    \includegraphics[width=\linewidth]{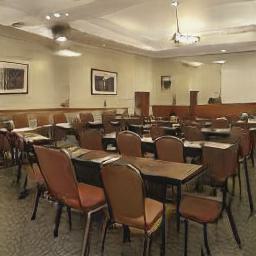}
\end{subfigure}
\quad
  \begin{subfigure}[t]{0.15\linewidth}
    \includegraphics[width=\linewidth]{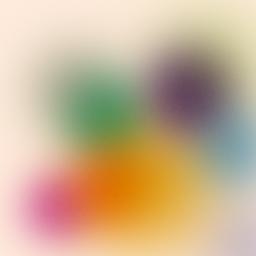}
\end{subfigure}
  \begin{subfigure}[t]{0.15\linewidth}
    \includegraphics[width=\linewidth]{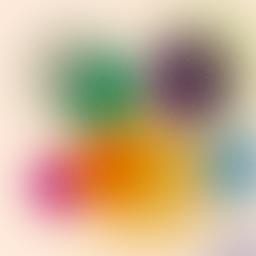}
\end{subfigure}
  \begin{subfigure}[t]{0.15\linewidth}
    \includegraphics[width=\linewidth]{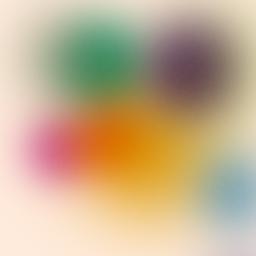}
\end{subfigure}\hspace{1mm}
  \begin{subfigure}[t]{0.15\linewidth}    \includegraphics[width=\linewidth,height=\linewidth]{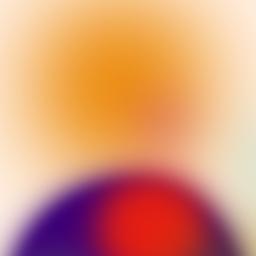}
\end{subfigure}
  \begin{subfigure}[t]{0.15\linewidth}    \includegraphics[width=\linewidth,height=\linewidth]{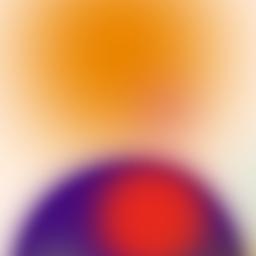}
\end{subfigure}
  \begin{subfigure}[t]{0.15\linewidth}
    \includegraphics[width=\linewidth]{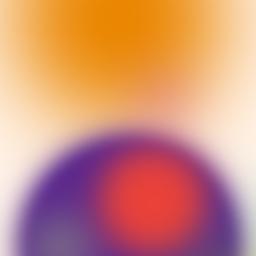}
\end{subfigure}
\caption{\textbf{Moving the camera in the vertical direction.} For every two rows, the first row shows the generated images, while the second row shows the corresponding blobs layout map.}
\label{fig:moving_pitch}
\vspace{-5mm}
\end{figure*}
\begin{figure*}[htbp]
  \centering
  \begin{subfigure}[t]{0.15\linewidth}
    \includegraphics[width=\linewidth]{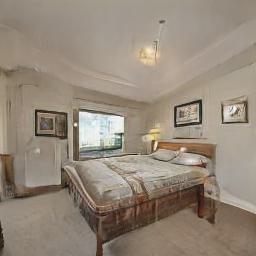}
\end{subfigure}
  \begin{subfigure}[t]{0.15\linewidth}
    \includegraphics[width=\linewidth]{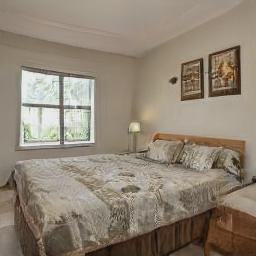}
\end{subfigure}
  \begin{subfigure}[t]{0.15\linewidth}
    \includegraphics[width=\linewidth]{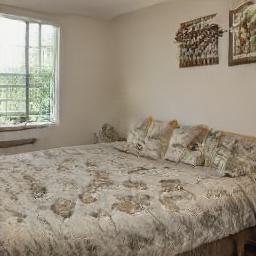}
\end{subfigure}\hspace{1mm}
  \begin{subfigure}[t]{0.15\linewidth}    \includegraphics[width=\linewidth,height=\linewidth]{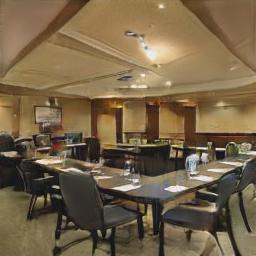}
\end{subfigure}
  \begin{subfigure}[t]{0.15\linewidth}    \includegraphics[width=\linewidth,height=\linewidth]{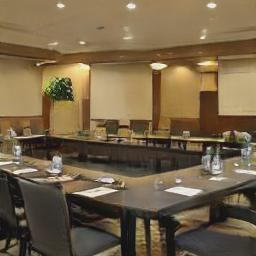}
\end{subfigure}
  \begin{subfigure}[t]{0.15\linewidth}
    \includegraphics[width=\linewidth]{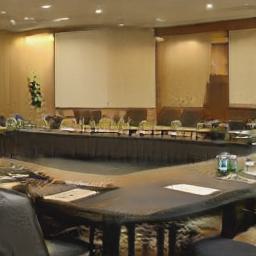}
\end{subfigure}
\quad
  \begin{subfigure}[t]{0.15\linewidth}
    \includegraphics[width=\linewidth]{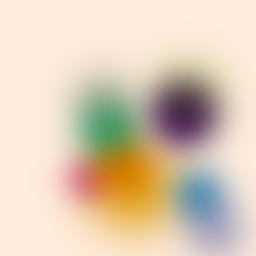}
\end{subfigure}
  \begin{subfigure}[t]{0.15\linewidth}
    \includegraphics[width=\linewidth]{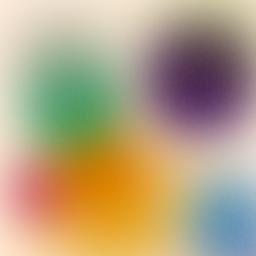}
\end{subfigure}
  \begin{subfigure}[t]{0.15\linewidth}
    \includegraphics[width=\linewidth]{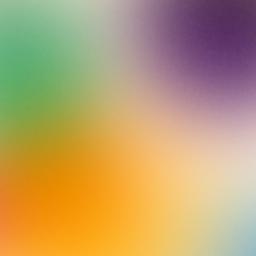}
\end{subfigure}\hspace{1mm}
  \begin{subfigure}[t]{0.15\linewidth}    \includegraphics[width=\linewidth,height=\linewidth]{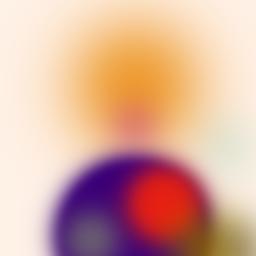}
\end{subfigure}
  \begin{subfigure}[t]{0.15\linewidth}    \includegraphics[width=\linewidth,height=\linewidth]{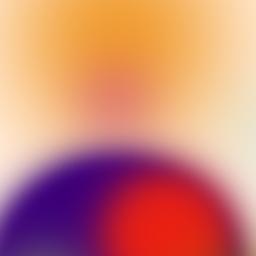}
\end{subfigure}
  \begin{subfigure}[t]{0.15\linewidth}
    \includegraphics[width=\linewidth]{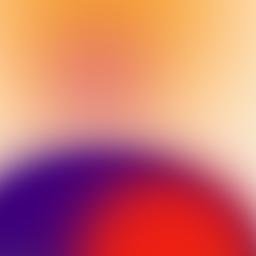}
\end{subfigure}
\quad
  \begin{subfigure}[t]{0.15\linewidth}
    \includegraphics[width=\linewidth]{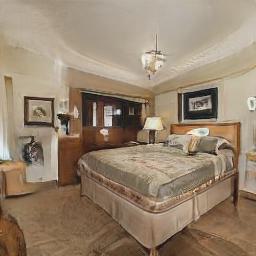}
\end{subfigure}
  \begin{subfigure}[t]{0.15\linewidth}
    \includegraphics[width=\linewidth]{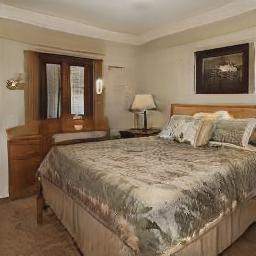}
\end{subfigure}
  \begin{subfigure}[t]{0.15\linewidth}
    \includegraphics[width=\linewidth]{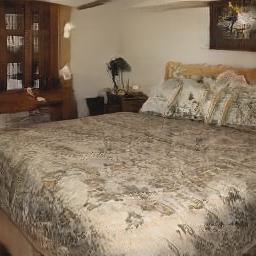}
\end{subfigure}\hspace{1mm}
  \begin{subfigure}[t]{0.15\linewidth}    \includegraphics[width=\linewidth,height=\linewidth]{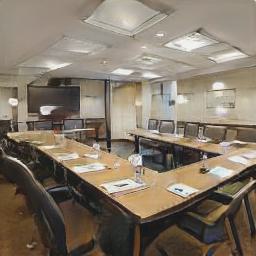}
\end{subfigure}
  \begin{subfigure}[t]{0.15\linewidth}    \includegraphics[width=\linewidth,height=\linewidth]{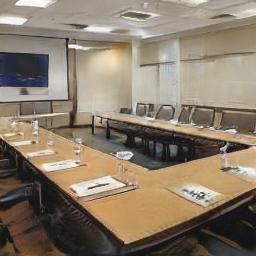}
\end{subfigure}
  \begin{subfigure}[t]{0.15\linewidth}
    \includegraphics[width=\linewidth]{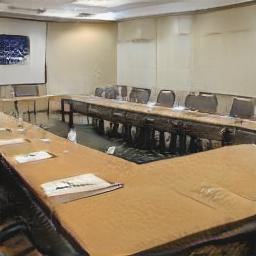}
\end{subfigure}
\quad
  \begin{subfigure}[t]{0.15\linewidth}
    \includegraphics[width=\linewidth]{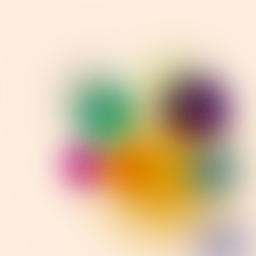}
\end{subfigure}
  \begin{subfigure}[t]{0.15\linewidth}
    \includegraphics[width=\linewidth]{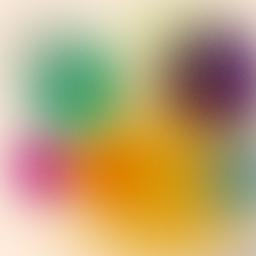}
\end{subfigure}
  \begin{subfigure}[t]{0.15\linewidth}
    \includegraphics[width=\linewidth]{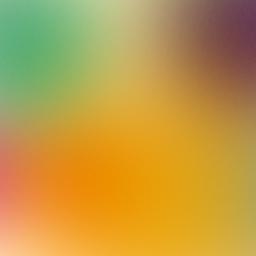}
\end{subfigure}\hspace{1mm}
  \begin{subfigure}[t]{0.15\linewidth}    \includegraphics[width=\linewidth,height=\linewidth]{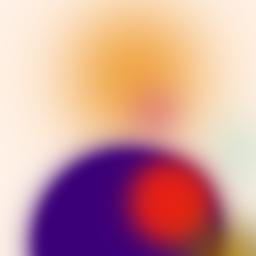}
\end{subfigure}
  \begin{subfigure}[t]{0.15\linewidth}    \includegraphics[width=\linewidth,height=\linewidth]{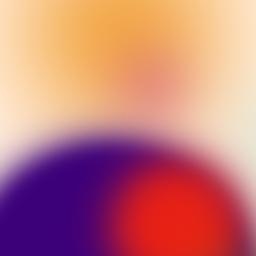}
\end{subfigure}
  \begin{subfigure}[t]{0.15\linewidth}
    \includegraphics[width=\linewidth]{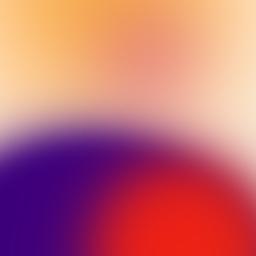}
\end{subfigure}
\quad
  \begin{subfigure}[t]{0.15\linewidth}
    \includegraphics[width=\linewidth]{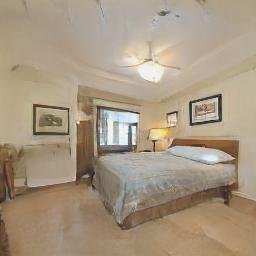}
\end{subfigure}
  \begin{subfigure}[t]{0.15\linewidth}
    \includegraphics[width=\linewidth]{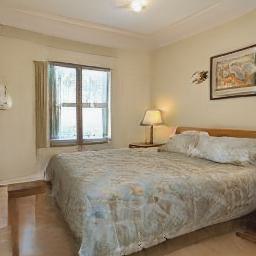}
\end{subfigure}
  \begin{subfigure}[t]{0.15\linewidth}
    \includegraphics[width=\linewidth]{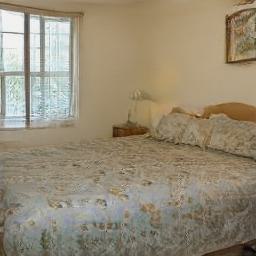}
\end{subfigure}\hspace{1mm}
  \begin{subfigure}[t]{0.15\linewidth}    \includegraphics[width=\linewidth,height=\linewidth]{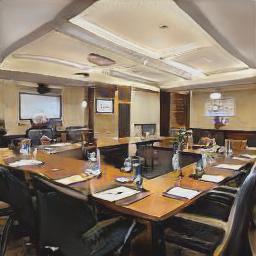}
\end{subfigure}
  \begin{subfigure}[t]{0.15\linewidth}    \includegraphics[width=\linewidth,height=\linewidth]{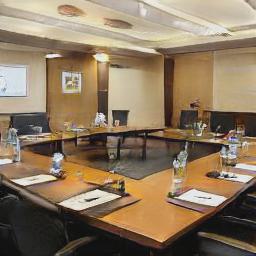}
\end{subfigure}
  \begin{subfigure}[t]{0.15\linewidth}
    \includegraphics[width=\linewidth]{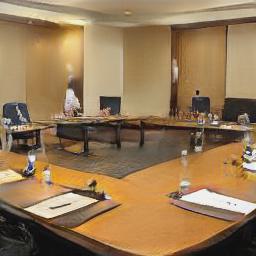}
\end{subfigure}
\quad
  \begin{subfigure}[t]{0.15\linewidth}
    \includegraphics[width=\linewidth]{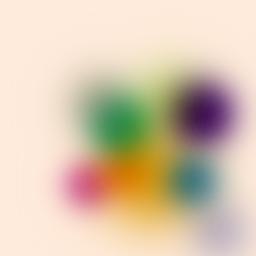}
\end{subfigure}
  \begin{subfigure}[t]{0.15\linewidth}
    \includegraphics[width=\linewidth]{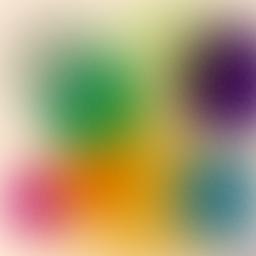}
\end{subfigure}
  \begin{subfigure}[t]{0.15\linewidth}
    \includegraphics[width=\linewidth]{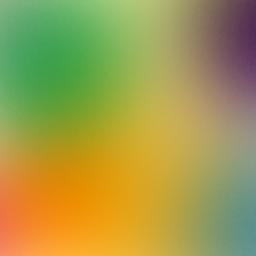}
\end{subfigure}\hspace{1mm}
  \begin{subfigure}[t]{0.15\linewidth}    \includegraphics[width=\linewidth,height=\linewidth]{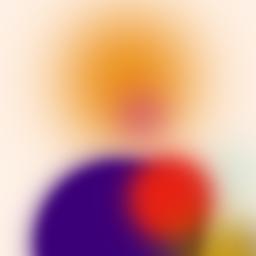}
\end{subfigure}
  \begin{subfigure}[t]{0.15\linewidth}    \includegraphics[width=\linewidth,height=\linewidth]{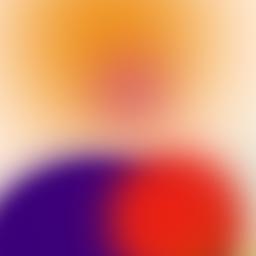}
\end{subfigure}
  \begin{subfigure}[t]{0.15\linewidth}
    \includegraphics[width=\linewidth]{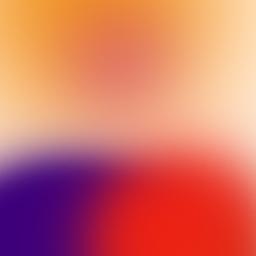}
\end{subfigure}
\quad
  \begin{subfigure}[t]{0.15\linewidth}
    \includegraphics[width=\linewidth]{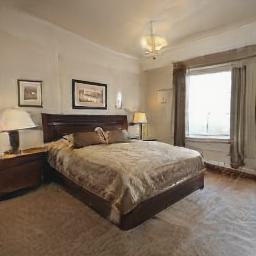}
\end{subfigure}
  \begin{subfigure}[t]{0.15\linewidth}
    \includegraphics[width=\linewidth]{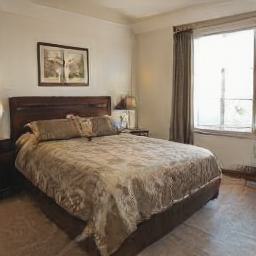}
\end{subfigure}
  \begin{subfigure}[t]{0.15\linewidth}
    \includegraphics[width=\linewidth]{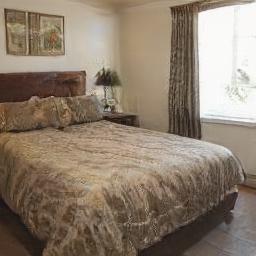}
\end{subfigure}\hspace{1mm}
  \begin{subfigure}[t]{0.15\linewidth}    \includegraphics[width=\linewidth,height=\linewidth]{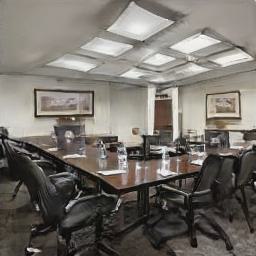}
\end{subfigure}
  \begin{subfigure}[t]{0.15\linewidth}    \includegraphics[width=\linewidth,height=\linewidth]{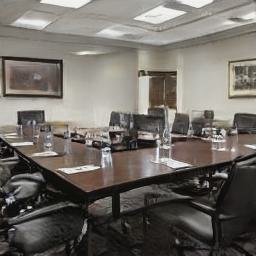}
\end{subfigure}
  \begin{subfigure}[t]{0.15\linewidth}
    \includegraphics[width=\linewidth]{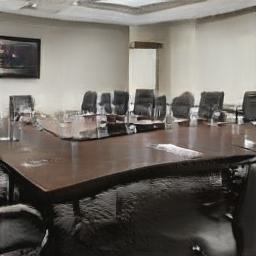}
\end{subfigure}
\quad
  \begin{subfigure}[t]{0.15\linewidth}
    \includegraphics[width=\linewidth]{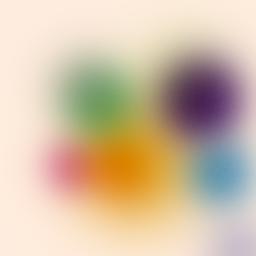}
\end{subfigure}
  \begin{subfigure}[t]{0.15\linewidth}
    \includegraphics[width=\linewidth]{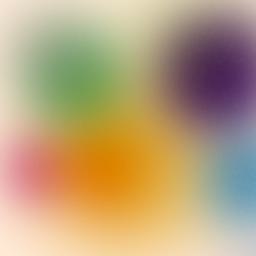}
\end{subfigure}
  \begin{subfigure}[t]{0.15\linewidth}
    \includegraphics[width=\linewidth]{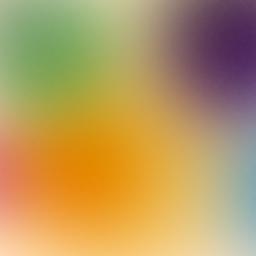}
\end{subfigure}\hspace{1mm}
  \begin{subfigure}[t]{0.15\linewidth}    \includegraphics[width=\linewidth,height=\linewidth]{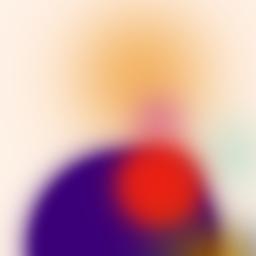}
\end{subfigure}
  \begin{subfigure}[t]{0.15\linewidth}    \includegraphics[width=\linewidth,height=\linewidth]{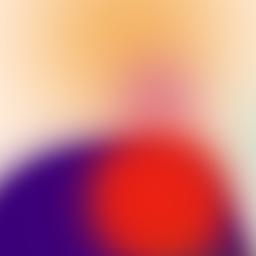}
\end{subfigure}
  \begin{subfigure}[t]{0.15\linewidth}
    \includegraphics[width=\linewidth]{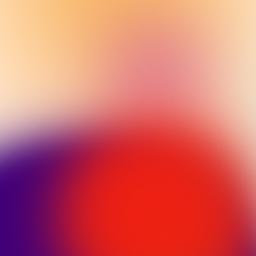}
\end{subfigure}
\caption{\textbf{Moving the camera to simulate zoom in and zoom out effects.} For every two rows, the first row shows the generated images, while the second row shows the corresponding blobs layout map.}
\label{fig:moving_radius}
\end{figure*}

We provide more editing results of moving objects in \cref{fig:moving} and resizing objects in \cref{fig:resizing}. For comparison, we show the editing results of original BlobGAN in \cref{fig:moving_2d} and \cref{fig:resizing_2d}. We visualize the blobs in 3D space in \cref{fig:3d_blobs}. In \cref{fig:foreshortening}, we show more examples of the foreshortening effect along with the blob visualization in 3D space. We also show more qualitative results of multiview rendering by moving the camera in the horizontal direction (\cref{fig:moving_yaw}), vertical direction (\cref{fig:moving_pitch}) and closer to or away from the viewpoint (\cref{fig:moving_radius}). Despite some minor changes due to the entanglement, our method achieves good consistency while enabling realistic multi-object editing on real-world indoor datasets.

\section{Ablation study}
\begin{figure*}[h]
\centering
\begin{subfigure}[t]{0.23\linewidth}
    \includegraphics[width=\linewidth]{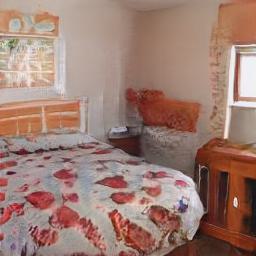}
\end{subfigure}
\begin{subfigure}[t]{0.23\linewidth}
    \includegraphics[width=\linewidth]{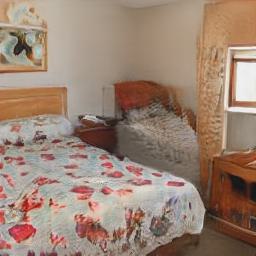}
\end{subfigure}
\begin{subfigure}[t]{0.23\linewidth}
    \includegraphics[width=\linewidth]{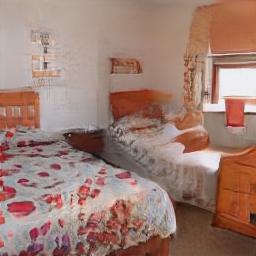}
\end{subfigure}
\begin{subfigure}[t]{0.23\linewidth}
    \includegraphics[width=\linewidth]{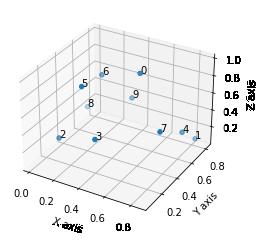}
\end{subfigure}
\quad
\begin{subfigure}[t]{0.23\linewidth}
    \includegraphics[width=\linewidth]{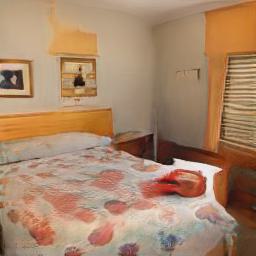}
\end{subfigure}
\begin{subfigure}[t]{0.23\linewidth}
    \includegraphics[width=\linewidth]{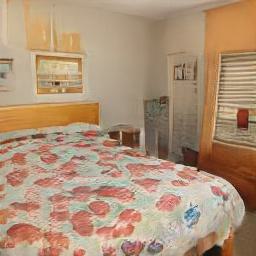}
\end{subfigure}
\begin{subfigure}[t]{0.23\linewidth}
    \includegraphics[width=\linewidth]{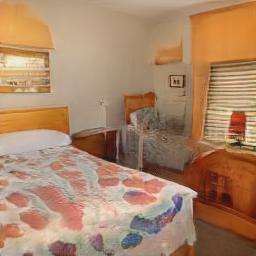}
\end{subfigure}
\begin{subfigure}[t]{0.23\linewidth}
    \includegraphics[width=\linewidth]{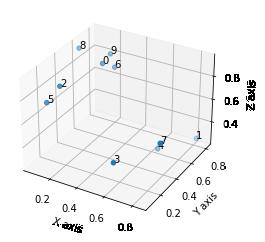}
\end{subfigure}
\caption{Depth supervision ablation study. The first row shows the results without depth supervision, while the second shows the results with depth supervision. The first three columns shows the results of multiview rendering. The fourth column shows the coordinates of all the blobs' centers. With depth supervision, the image quality and the multiview rendering quality are better than without depth supervision. } 
\label{fig:depth_ablation}
\end{figure*}

We provide the results of an ablation study on depth supervision in \cref{fig:depth_ablation}. Because of limited resources, both models with and without depth supervision are not trained till convergence but for roughly the same number of iterations. We can also compute the same depth estimation loss as we defined before:
\begin{equation}
l_{\text{depth}} = ||z_s - D(u, v)||^2,
\label{eq:depth_loss}
\end{equation}
where $(x_s, y_s, z_s)$ are the world coordinates of the blob center, $(u, v)$ are the coordinates of the projection in image space, and $D$ is the depth estimator. We randomly generate 5000 images under the same camera sampling strategy for each model and compute the corresponding depth estimation loss. Without depth supervision, the average depth estimation loss is 0.197. With depth supervision, this number drops around 36\% to 0.116.

\clearpage
\section{Limitations}
\begin{figure*}[h]
\centering   
\begin{subfigure}[t]{0.22\linewidth}
\includegraphics[width=\linewidth]{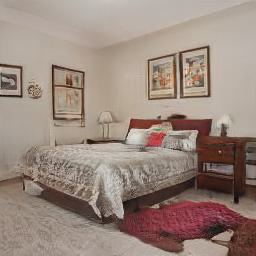}
\end{subfigure}
\begin{subfigure}[t]{0.22\linewidth}
\includegraphics[width=\linewidth]{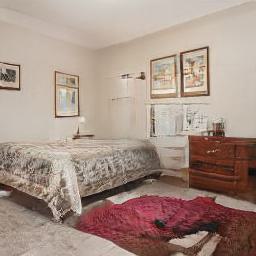}
\end{subfigure}
\hspace{3mm}
\begin{subfigure}[t]{0.22\linewidth}
\includegraphics[width=\linewidth]{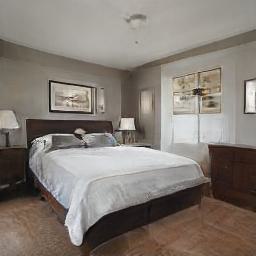}
\end{subfigure}
\begin{subfigure}[t]{0.22\linewidth}
\includegraphics[width=\linewidth]{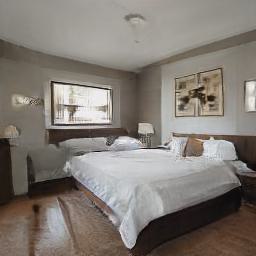}
\end{subfigure}
   \caption{\textbf{Typical failure cases in our method}. When moving the bed in the scene, sometimes the original bedside will disappear, either because of being occluded by the painting blob above the bed, or the bedside is actually part of the painting blob above. A new bedside may appear on the other side of the bed, which is actually due to the failed disentanglement of the other painting blob.}
   \label{fig:failure}
   \end{figure*}

Our work also has several limitations. 
Both BlobGAN and our work can obtain good disentanglement of individual objects in an unsupervised manner on a complex dataset. However, in both methods, the disentanglement is not perfect. Changing one blob may still influence the appearance or location of other objects in the scene. We show some typical failure cases in our method in \cref{fig:failure}. The depth estimator can help to correct the depth of the centroid of the blob. However, as the shape and size of the blobs are not perfectly matched to the shape and size of the corresponding object, the estimated depth from image space may not reveal the depth of the object in 3D scene. Despite good image quality, the time BlobGAN and BlobGAN-3D take to converge is very long compared to other GANs. We explain this long training time by the fact that the scene is modeled as a collection of blobs, which can be considered as an intermediate representation. Rendering the blobs first and mapping them to an image of the scene takes longer than just rendering the scene itself directly. 
\par


\begin{figure*}[t]
  \centering
  \begin{subfigure}[t]{0.11\linewidth}
    \includegraphics[width=\linewidth]{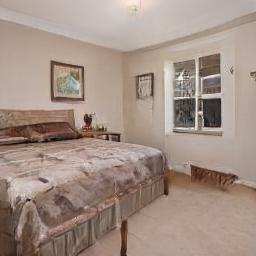}
\end{subfigure}
\begin{subfigure}[t]{0.11\linewidth}
    \includegraphics[width=\linewidth]{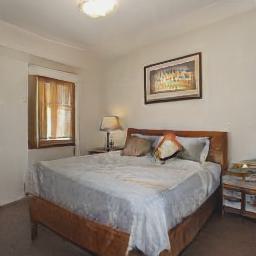}
\end{subfigure}
  \begin{subfigure}[t]{0.11\linewidth}
    \includegraphics[width=\linewidth]{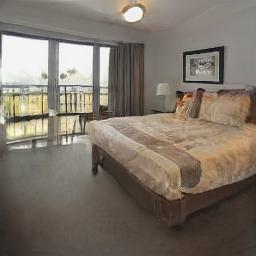}
\end{subfigure}
  \begin{subfigure}[t]{0.11\linewidth}
    \includegraphics[width=\linewidth,height=\linewidth]{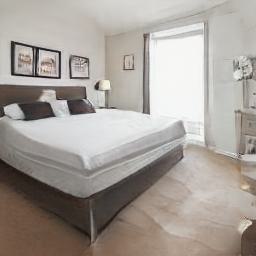}
\end{subfigure}
  \begin{subfigure}[t]{0.11\linewidth}
    \includegraphics[width=\linewidth,height=\linewidth]{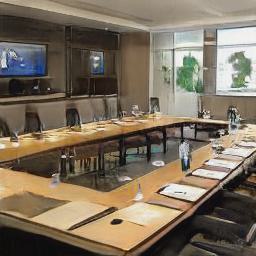}
\end{subfigure}
  \begin{subfigure}[t]{0.11\linewidth}
    \includegraphics[width=\linewidth,height=\linewidth]{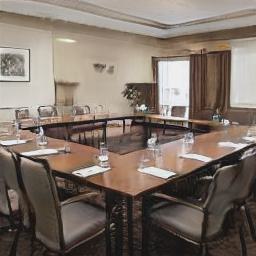}
\end{subfigure}
  \begin{subfigure}[t]{0.11\linewidth}
    \includegraphics[width=\linewidth]{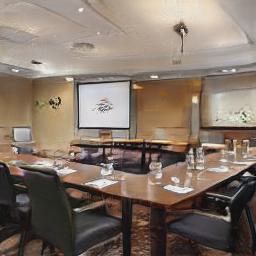}
\end{subfigure}
  \begin{subfigure}[t]{0.11\linewidth}
    \includegraphics[width=\linewidth]{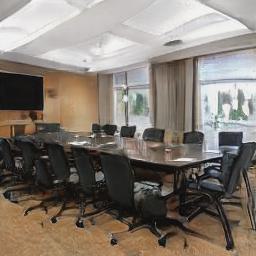}
\end{subfigure}
\quad
  \begin{subfigure}[t]{0.11\linewidth}
       \begin{overpic}[width=\textwidth]{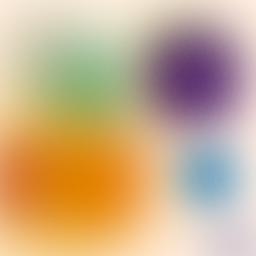}
\linethickness{1.5pt}
\put(35,30){\color[RGB]{255,255,255}\vector(1,0){40}}
\end{overpic}
\end{subfigure}
  \begin{subfigure}[t]{0.11\linewidth}
   \begin{overpic}[width=\textwidth]{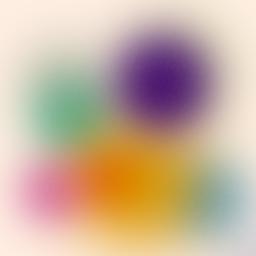}
\linethickness{1.5pt}
\put(55,30){\color[RGB]{255,255,255}\vector(1,0){40}}
\end{overpic}
\end{subfigure}
  \begin{subfigure}[t]{0.11\linewidth}
  \begin{overpic}[width=\textwidth]{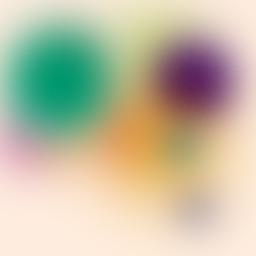}
\linethickness{1.5pt}
\put(77,68){\color[RGB]{255,255,255}\vector(1,0){25}}
\end{overpic}  
\end{subfigure}
  \begin{subfigure}[t]{0.11\linewidth}
   \begin{overpic}[width=\textwidth]{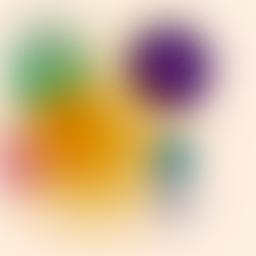}
\linethickness{1.5pt}
\put(67,72){\color[RGB]{255,255,255}\vector(1,0){30}}
\end{overpic}  
\end{subfigure}
  \begin{subfigure}[t]{0.11\linewidth}
  \begin{overpic}[width=\textwidth]{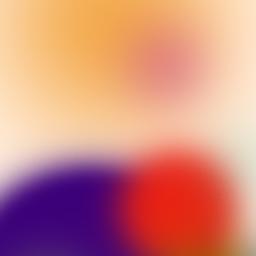}
\linethickness{1.5pt}
\put(71,21){\color[RGB]{255,255,255}\vector(1,0){30}}
\end{overpic}  
\end{subfigure}
  \begin{subfigure}[t]{0.11\linewidth}
  \begin{overpic}[width=\textwidth]{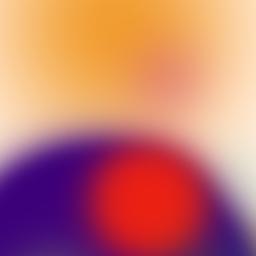}
\linethickness{1.5pt}
\put(58,21){\color[RGB]{255,255,255}\vector(1,0){30}}
\end{overpic}  
\end{subfigure}
  \begin{subfigure}[t]{0.11\linewidth}
  \begin{overpic}[width=\textwidth]{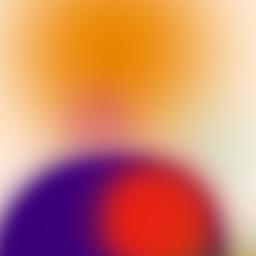}
\linethickness{1.5pt}
\put(40,58){\color[RGB]{255,255,255}\vector(1,0){30}}
\end{overpic}  
\end{subfigure}
  \begin{subfigure}[t]{0.11\linewidth}
  \begin{overpic}[width=\textwidth]{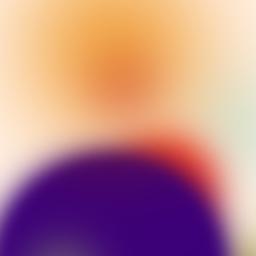}
\linethickness{1.5pt}
\put(48,70){\color[RGB]{255,255,255}\vector(1,0){30}}
\end{overpic}  
\end{subfigure}
\quad
  \begin{subfigure}[t]{0.11\linewidth}
    \includegraphics[width=\linewidth]{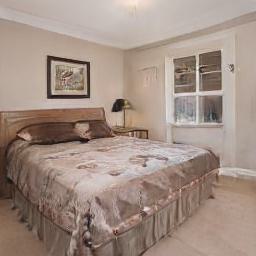}
\end{subfigure}
  \begin{subfigure}[t]{0.11\linewidth}
    \includegraphics[width=\linewidth]{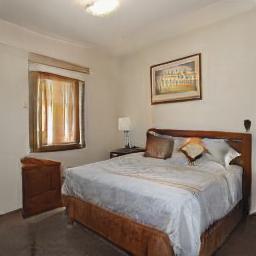}
\end{subfigure}
  \begin{subfigure}[t]{0.11\linewidth}
    \includegraphics[width=\linewidth]{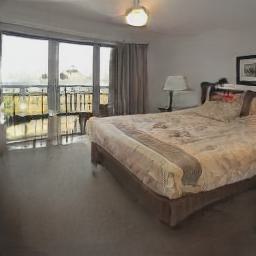}
\end{subfigure}
  \begin{subfigure}[t]{0.11\linewidth}    \includegraphics[width=\linewidth,height=\linewidth]{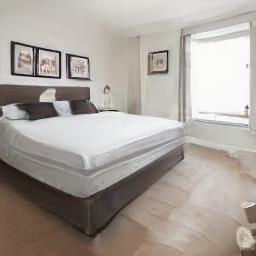}
\end{subfigure}
  \begin{subfigure}[t]{0.11\linewidth}
    \includegraphics[width=\linewidth,height=\linewidth]{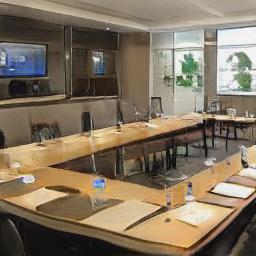}
\end{subfigure}
  \begin{subfigure}[t]{0.11\linewidth}
    \includegraphics[width=\linewidth,height=\linewidth]{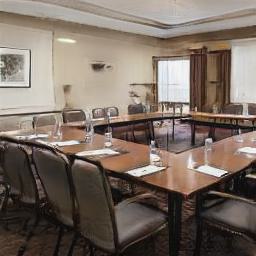}
\end{subfigure}
  \begin{subfigure}[t]{0.11\linewidth}
    \includegraphics[width=\linewidth]{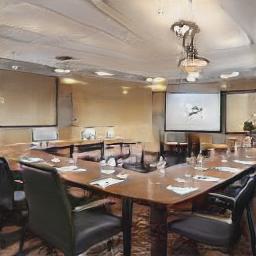}
\end{subfigure}
  \begin{subfigure}[t]{0.11\linewidth}
    \includegraphics[width=\linewidth]{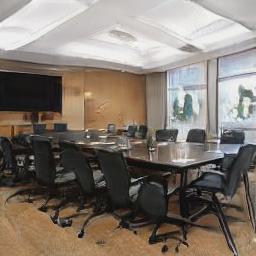}
\end{subfigure}
\caption{\textbf{BlobGAN-3D: Moving an object in the scene.} The first row shows the original generated images; the second row shows the corresponding blob layout maps with marks demonstrating the moving direction. The last row shows the synthesized images after moving one blob.}
\label{fig:moving}
\end{figure*}

\clearpage
\begin{figure*}[t]
  \centering
  \begin{subfigure}[t]{0.11\linewidth}
    \includegraphics[width=\linewidth]{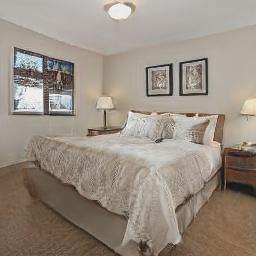}
\end{subfigure}
  \begin{subfigure}[t]{0.11\linewidth}
    \includegraphics[width=\linewidth]{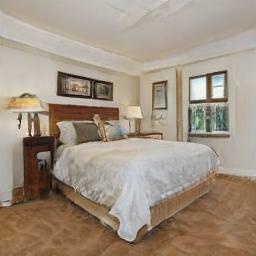}
\end{subfigure}
  \begin{subfigure}[t]{0.11\linewidth}
    \includegraphics[width=\linewidth]{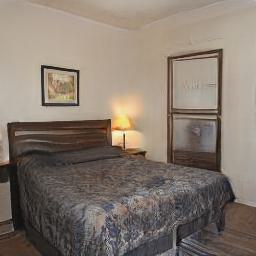}
\end{subfigure}
  \begin{subfigure}[t]{0.11\linewidth}
    \includegraphics[width=\linewidth,height=\linewidth]{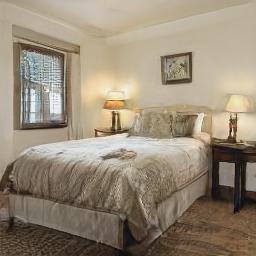}
\end{subfigure}
  \begin{subfigure}[t]{0.11\linewidth}
    \includegraphics[width=\linewidth,height=\linewidth]{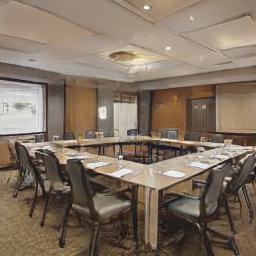}
\end{subfigure}
  \begin{subfigure}[t]{0.11\linewidth}
    \includegraphics[width=\linewidth,height=\linewidth]{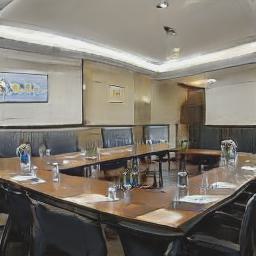}
\end{subfigure}
  \begin{subfigure}[t]{0.11\linewidth}
    \includegraphics[width=\linewidth]{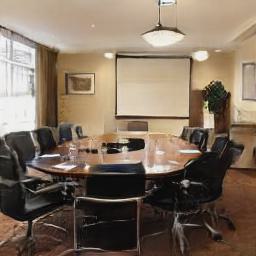}
\end{subfigure}
  \begin{subfigure}[t]{0.11\linewidth}
    \includegraphics[width=\linewidth]{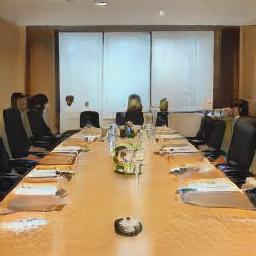}
\end{subfigure}
\quad
\\
  \begin{subfigure}[t]{0.11\linewidth}
  \begin{overpic}[width=\textwidth]{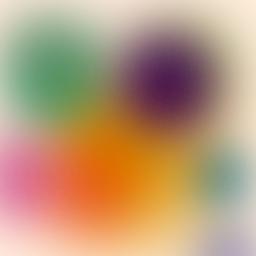}
\linethickness{1.5pt}
\put(38,40){\color[RGB]{255,255,255}\line(1,-1){20}}
\put(58,40){\color[RGB]{255,255,255}\line(-1,-1){20}}
\end{overpic} 
\end{subfigure}
  \begin{subfigure}[t]{0.11\linewidth}
  \begin{overpic}[width=\textwidth]{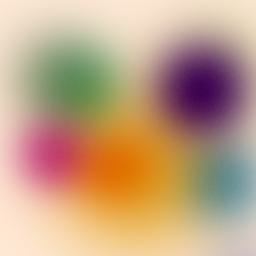}
\linethickness{1.5pt}
\put(43,40){\color[RGB]{255,255,255}\line(1,-1){20}}
\put(63,40){\color[RGB]{255,255,255}\line(-1,-1){20}}
\end{overpic} 
\end{subfigure}
  \begin{subfigure}[t]{0.11\linewidth}
  \begin{overpic}[width=\textwidth]{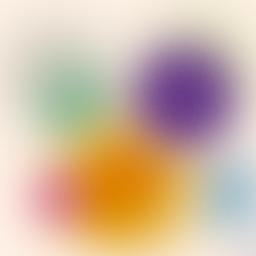}
\linethickness{1.5pt}
\put(66,70){\color[RGB]{255,255,255}\line(1,-1){15}}
\put(81,70){\color[RGB]{255,255,255}\line(-1,-1){15}}
\end{overpic} 
\end{subfigure}
  \begin{subfigure}[t]{0.11\linewidth}
  \begin{overpic}[width=\textwidth]{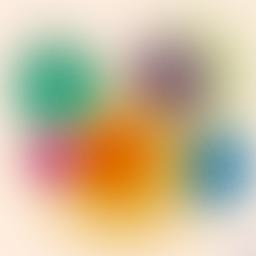}
\linethickness{1.5pt}
\put(62,77){\color[RGB]{255,255,255}\line(1,-1){15}}
\put(77,77){\color[RGB]{255,255,255}\line(-1,-1){15}}
\end{overpic} 
\end{subfigure}
  \begin{subfigure}[t]{0.11\linewidth}
  \begin{overpic}[width=\textwidth]{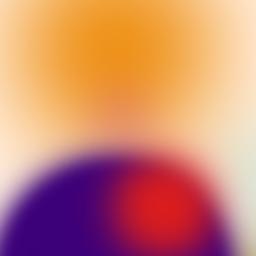}
\linethickness{1.5pt}
\put(35,40){\color[RGB]{255,255,255}\line(1,-1){30}}
\put(63,40){\color[RGB]{255,255,255}\line(-1,-1){30}}
\end{overpic} 
\end{subfigure}
  \begin{subfigure}[t]{0.11\linewidth}
  \begin{overpic}[width=\textwidth]{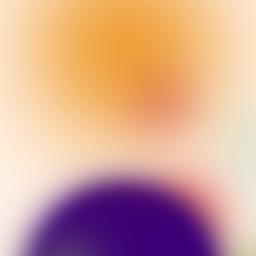}
\linethickness{1.5pt}
\put(38,32){\color[RGB]{255,255,255}\line(1,-1){28}}
\put(64,32){\color[RGB]{255,255,255}\line(-1,-1){28}}
\end{overpic} 
\end{subfigure}
  \begin{subfigure}[t]{0.11\linewidth}
  \begin{overpic}[width=\textwidth]{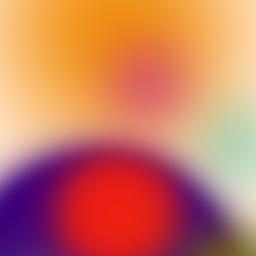}
\linethickness{1.5pt}
\put(48,78){\color[RGB]{255,255,255}\line(1,-1){20}}
\put(68,78){\color[RGB]{255,255,255}\line(-1,-1){20}}
\end{overpic} 
\end{subfigure}
  \begin{subfigure}[t]{0.11\linewidth}
  \begin{overpic}[width=\textwidth]{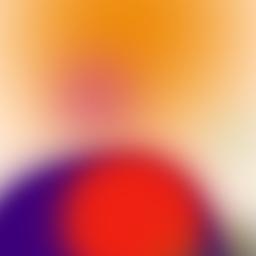}
\linethickness{1.5pt}
\put(28,72){\color[RGB]{255,255,255}\line(1,-1){20}}
\put(48,72){\color[RGB]{255,255,255}\line(-1,-1){20}}
\end{overpic} 
\end{subfigure}
\quad
  \begin{subfigure}[t]{0.11\linewidth}
    \includegraphics[width=\linewidth]{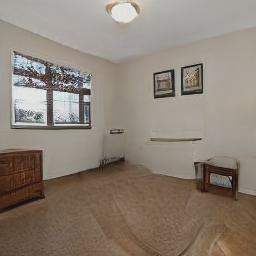}
\end{subfigure}
  \begin{subfigure}[t]{0.11\linewidth}
    \includegraphics[width=\linewidth]{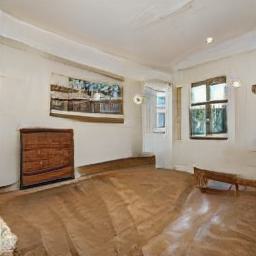}
\end{subfigure}
  \begin{subfigure}[t]{0.11\linewidth}
    \includegraphics[width=\linewidth]{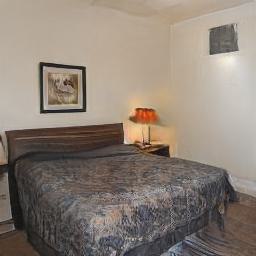}
\end{subfigure}
  \begin{subfigure}[t]{0.11\linewidth}    \includegraphics[width=\linewidth,height=\linewidth]{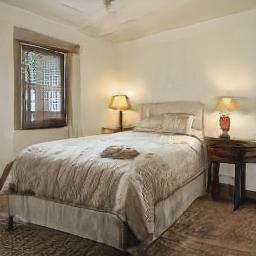}
\end{subfigure}
  \begin{subfigure}[t]{0.11\linewidth}
    \includegraphics[width=\linewidth,height=\linewidth]{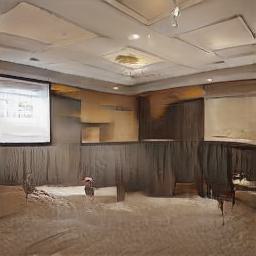}
\end{subfigure}
  \begin{subfigure}[t]{0.11\linewidth}
    \includegraphics[width=\linewidth,height=\linewidth]{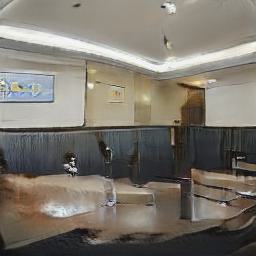}
\end{subfigure}
  \begin{subfigure}[t]{0.11\linewidth}
    \includegraphics[width=\linewidth]{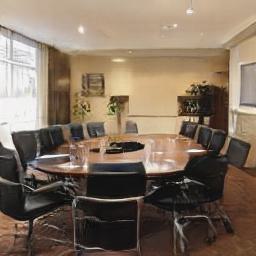}
\end{subfigure}
  \begin{subfigure}[t]{0.11\linewidth}
    \includegraphics[width=\linewidth]{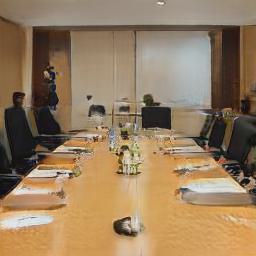}
\end{subfigure}
\caption{\textbf{BlobGAN-3D: Removing an object in the scene.} The first row shows the originally generated images; the second row shows the corresponding blob layout maps with marks demonstrating the removed blob. The last row shows the synthesized images after removing one blob.}
\label{fig:resizing}
\end{figure*}

\clearpage

\begin{figure*}[t]
  \centering
  \begin{subfigure}[t]{0.16\linewidth}
    \includegraphics[width=\linewidth]{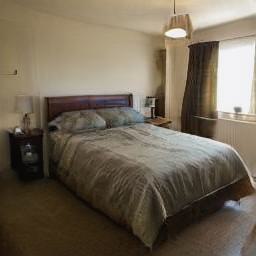}
\end{subfigure}
  \begin{subfigure}[t]{0.16\linewidth}
    \includegraphics[width=\linewidth]{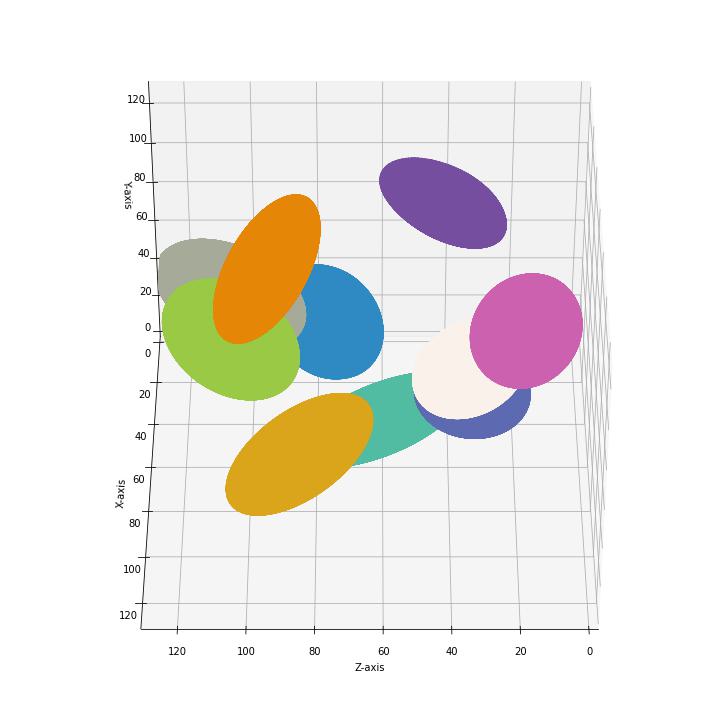}
\end{subfigure}
  \begin{subfigure}[t]{0.16\linewidth}
    \includegraphics[width=\linewidth]{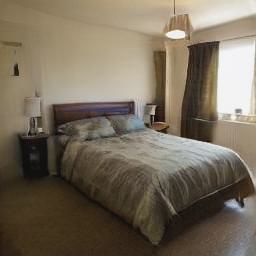}
\end{subfigure}
  \begin{subfigure}[t]{0.16\linewidth}
    \includegraphics[width=\linewidth]{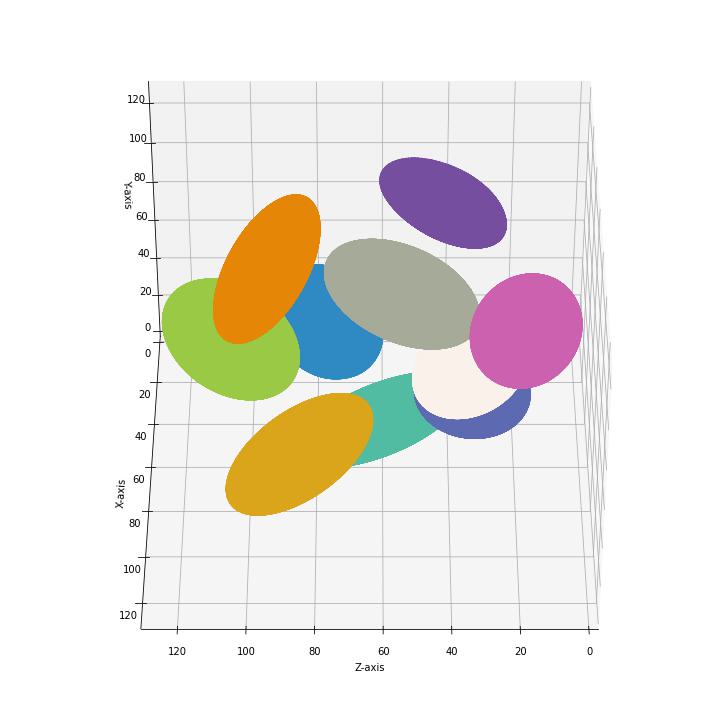}
\end{subfigure}
  \begin{subfigure}[t]{0.16\linewidth}
    \includegraphics[width=\linewidth]{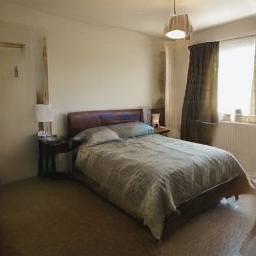}
\end{subfigure}
  \begin{subfigure}[t]{0.16\linewidth}
    \includegraphics[width=\linewidth]{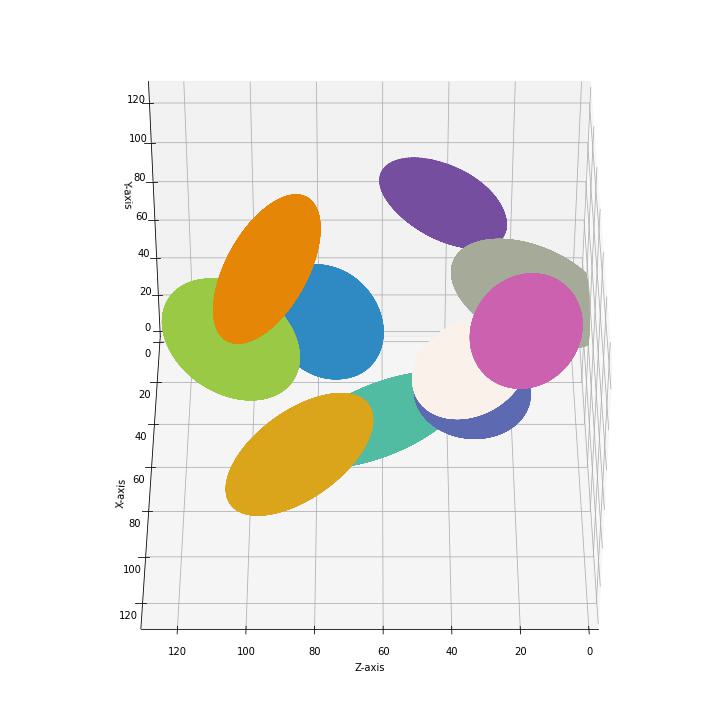}
\end{subfigure}
\quad
  \begin{subfigure}[t]{0.16\linewidth}
    \includegraphics[width=\linewidth]{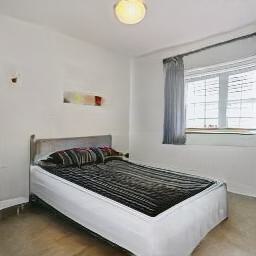}
\end{subfigure}
  \begin{subfigure}[t]{0.16\linewidth}
    \includegraphics[width=\linewidth]{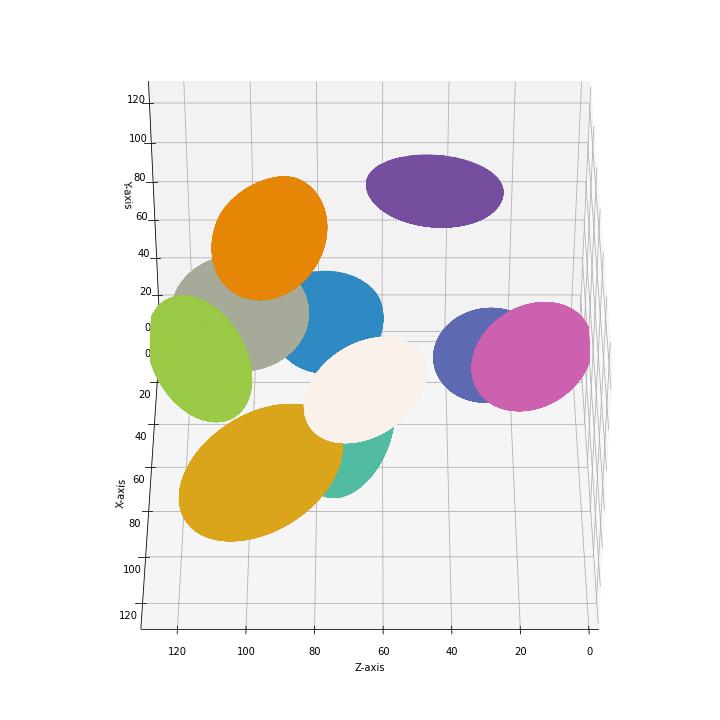}
\end{subfigure}
  \begin{subfigure}[t]{0.16\linewidth}
    \includegraphics[width=\linewidth]{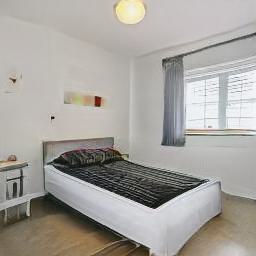}
\end{subfigure}
  \begin{subfigure}[t]{0.16\linewidth}
    \includegraphics[width=\linewidth]{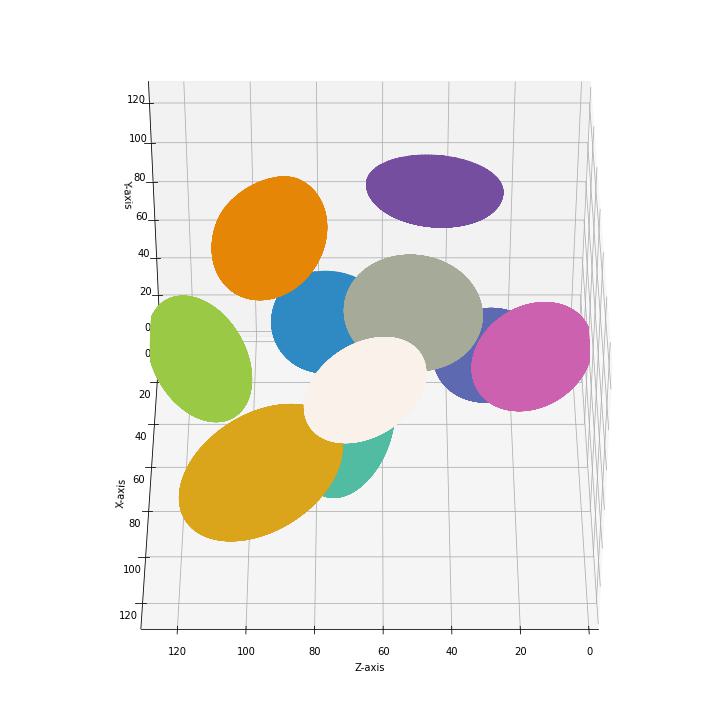}
\end{subfigure}
  \begin{subfigure}[t]{0.16\linewidth}
    \includegraphics[width=\linewidth]{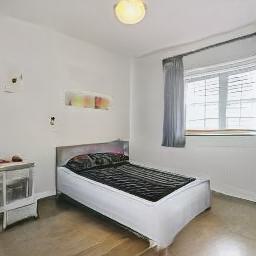}
\end{subfigure}
  \begin{subfigure}[t]{0.16\linewidth}
    \includegraphics[width=\linewidth]{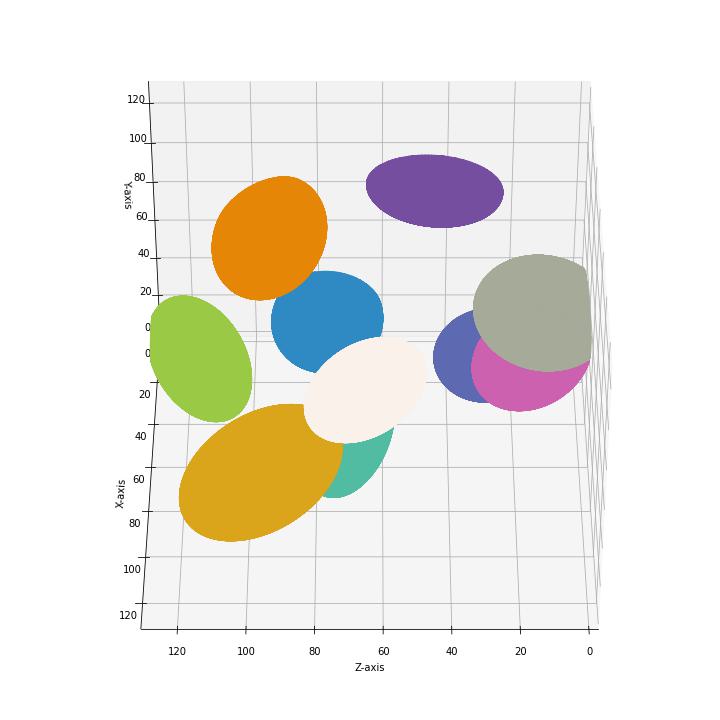}
\end{subfigure}
\caption{\textbf{Foreshortening effect.} When moving the gray blob which represents the bed along the depth dimension (Z-axis) away from the viewpoint, the size of the bed becomes smaller. } 
\label{fig:foreshortening}
\end{figure*}


\end{document}